\documentclass{article}
\usepackage{arxiv}

\usepackage{tabularx}
\usepackage[utf8]{inputenc}
\usepackage[titletoc,title]{appendix}
\usepackage{xcolor}
\usepackage{algorithm, algpseudocode}
\usepackage{url}
\def\code#1{\texttt{#1}}
\def\comment#1{\Comment{\textit{#1}}}
\usepackage{epigraph}
\usepackage{gensymb}
\algrenewcommand\textproc{}

\usepackage{etoolbox}
\makeatletter
\newlength\epitextskip
\pretocmd{\@epitext}{\em}{}{}
\apptocmd{\@epitext}{\em}{}{}
\patchcmd{\epigraph}{\@epitext{#1}\\}{\@epitext{#1}\\[\epitextskip]}{}{}
\makeatother
\usepackage{amsmath,amsfonts,amssymb,mathtools}

\usepackage{graphicx,float}
\usepackage{caption}
\usepackage{subcaption}

\usepackage{authblk}

\usepackage[square,numbers]{natbib}
\title{Discovering interpretable models of scientific image data with deep learning}
\author[1]{Christopher J. Soelistyo}
\author[1,2,3]{Alan R. Lowe}
\affil[1]{Alan Turing Institute}
\affil[2]{University College London}
\affil[3]{Institute for the Physics of Living Systems}

\affil[ ]{Correspondence to: \texttt{\{csoelistyo,alowe\}@turing.ac.uk}}
\date{}

\begin{document}
\maketitle

\begin{abstract}
    How can we find interpretable, domain-appropriate models of natural phenomena given some complex, raw data such as images? Can we use such models to derive scientific insight from the data? In this paper, we propose some methods for achieving this. In particular, we implement disentangled representation learning, sparse deep neural network training and symbolic regression, and assess their usefulness in forming interpretable models of complex image data. We demonstrate their relevance to the field of bioimaging using a well-studied test problem of classifying cell states in microscopy data. We find that such methods can produce highly parsimonious models that achieve $\sim98\%$ of the accuracy of black-box benchmark models, with a tiny fraction of the complexity. We explore the utility of such interpretable models in producing scientific explanations of the underlying biological phenomenon.
\end{abstract}
\epigraph{All models are wrong but some are useful.}{George Box \cite{box_robustness_1979}}

\section{Introduction}

Advances in artificial intelligence have raised its potential for use in the sciences. This is particularly true in fields such as bioimaging, where the abundance of raw and complex data has placed a premium on algorithms capable of efficiency extracting high-level regularities. However, the impressive capabilities of deep neural networks are accompanied by some of their less desirable properties. In particular, these models are notoriously un-interpretable, and they are susceptible to forming representations that are inappropriate to the domain of application. In this study, we explore some methods for capitalizing on the strengths of deep learning while mitigating its weaknesses. The broad goal is the construction of a discovery system that can induce, given some data, a set of laws that human scientists may consider acceptable. We assess the potential for such a system in the field of bioimaging.

\section{Background}
\subsection{The promise of deep learning in scientific discovery}

Science consists of the construction of theories and the assessment of their ability to accurately predict real-world phenomena. Observation and experiment are the ultimate criteria of theory; however, they generally do not \emph{determine} the theories themselves, whose construction has typically involved an element of intuition \cite{popper_logic_2002}, or the application of reliable heuristics \cite{langley_scientific_1992}.

Nevertheless, developments in artificial intelligence have facilitated the use of \emph{data-driven} modes of scientific discovery, where observations play a key role in an inductive process of theory construction. An inductive method essentially consists of two elements: a "space" of possible solutions, and a rule, or "heuristic", that defines how this space is searched, typically based on observed data \cite{langley_scientific_1992}. In the context of scientific discovery, this is the space of all possible theories, or models.

An ideal inductive system has a broad theory space and a heuristic that is generally applicable. Moreover, for purposes of speed and precision, it should be capable of instantiation in a computer program. All three criteria are satisfied by deep neural networks (DNNs). Unlike "traditional" machine-learning (ML) models, such as linear regression models or support vector machines, DNNs are unconstrained in the mathematical form that their final model configurations can approximate \cite{hornik_multilayer_1989}. Furthermore, gradient descent via backpropagation serves as a general-purpose heuristic that can be successfully applied in a multitude of contexts. Therefore, by training DNNs on well-defined problems, we may instill in them some knowledge of how some system operates. For example, a DNN trained to predict planetary motions may gain some knowledge of classical mechanics \cite{lemos_rediscovering_2022}.

Deep learning therefore exhibits unique strengths in domains where the correct mathematical form for a model might not be obvious. This is particularly true for cases involving raw data, such as images or audio, where the individual input features are not meaningful in and of themselves. While traditional ML models achieve competitive performance for many problems involving "tabular" data, which \emph{does} involve meaningful features, DNNs still typically dominate in cases involving raw data \cite{semenova_existence_2022}. They therefore hold significant potential for fields such as astronomy \cite{smith_astronomia_2023} and especially biology, where an "imaging tsunami" has produced enormous amounts of raw data whose analysis requires computational aid \cite{ouyang_imaging_2017}.

\subsection{The dangers of deep learning in scientific discovery}
\label{section:dangers_deep_learning}

\begin{figure}
 \centering
    \begin{subfigure}[b]{0.49\textwidth}
      \includegraphics[width=\textwidth]{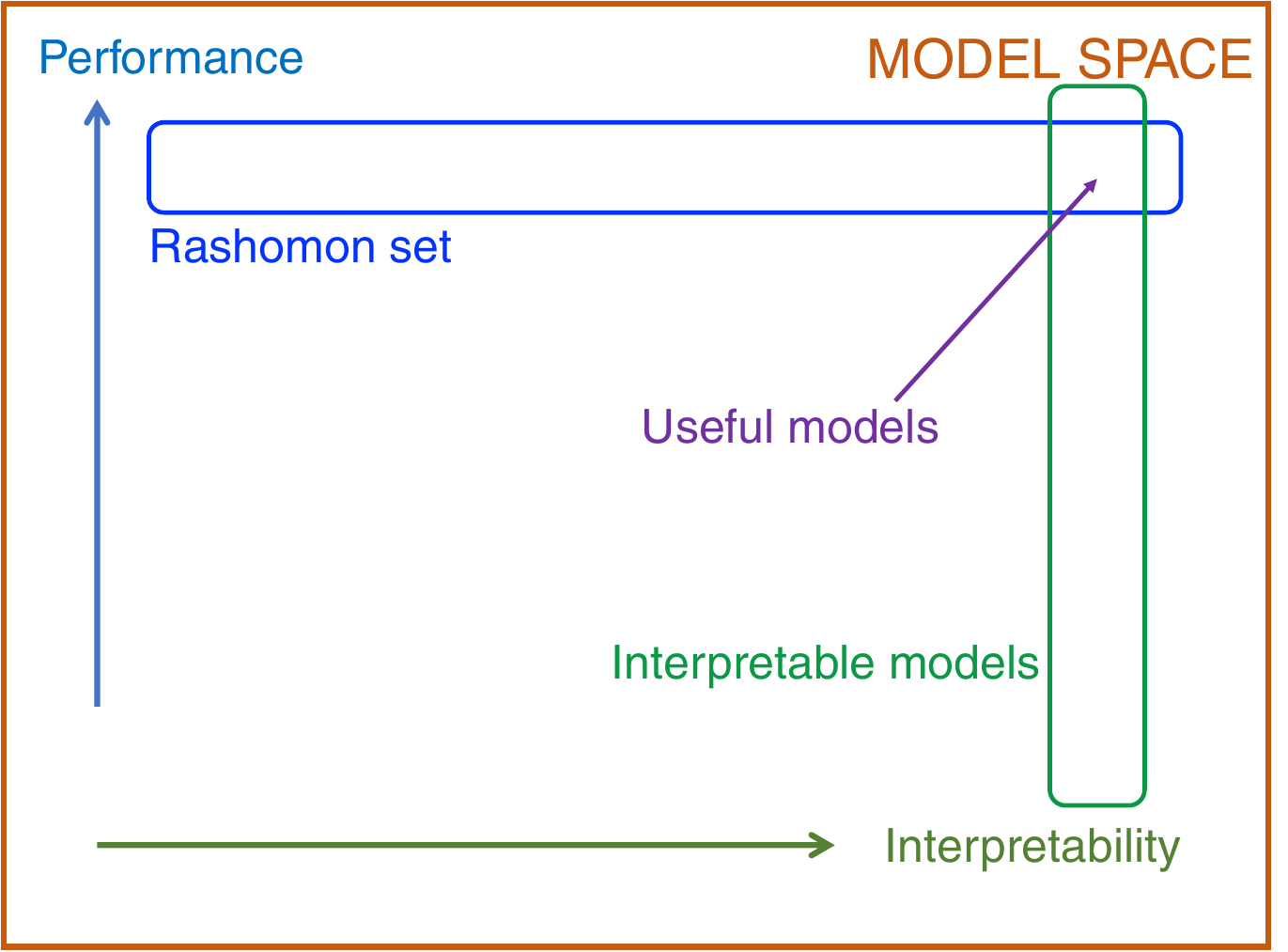}
      \caption{}
    \end{subfigure}
    \begin{subfigure}[b]{0.49\textwidth}
      \includegraphics[width=\textwidth]{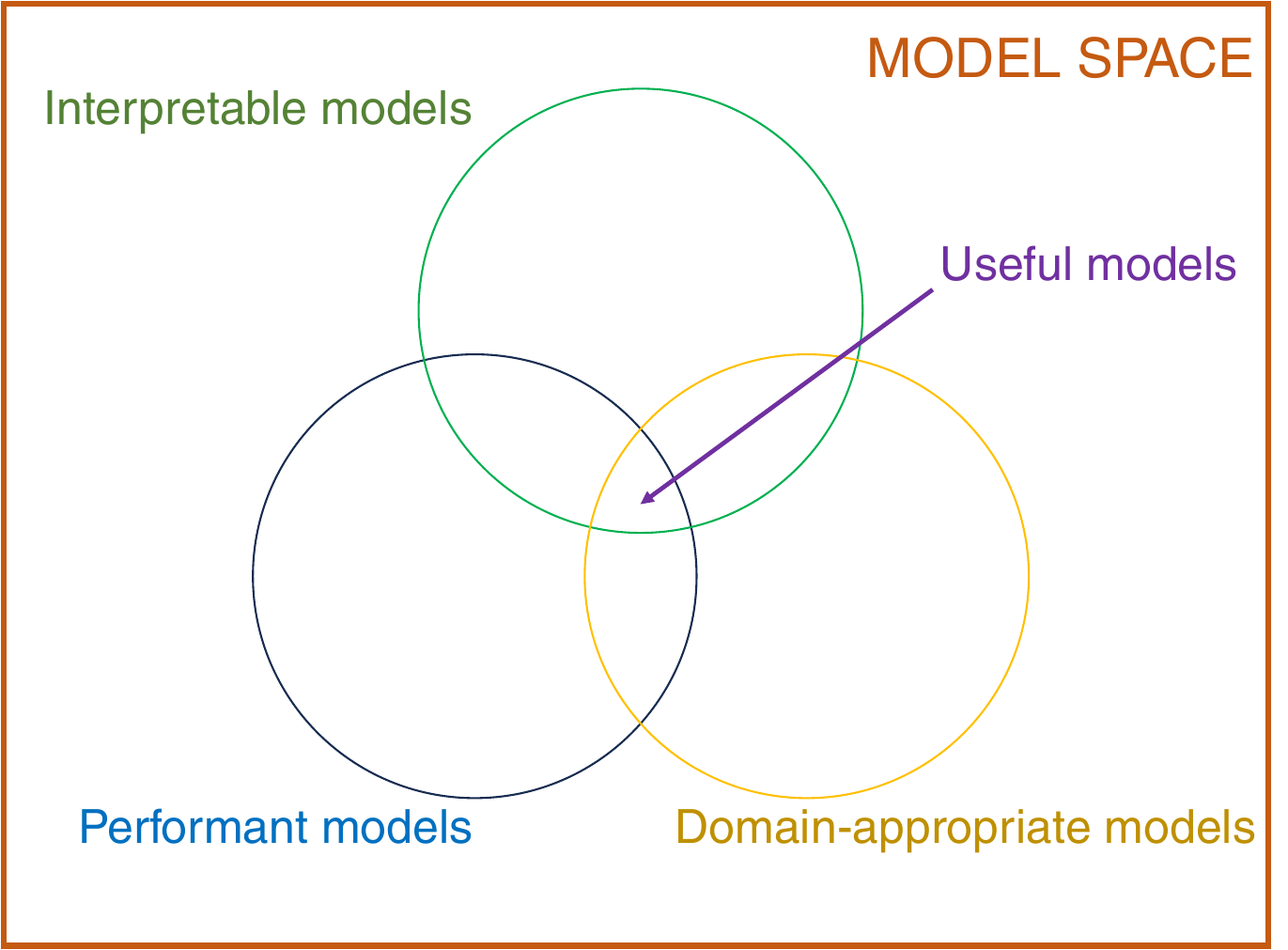}
      \caption{}
    \end{subfigure}
    \caption{Fundamental concepts of the study. \textbf{(a)} "Rashomon set" concept. \textbf{(b)} The "sweet spot" of useful scientific models, at the intersection of the performant, interpretable and domain-appropriate model subspaces within the overall model space.}
    \label{fig:model_concept}
\end{figure}

Despite their impressive capabilities, deep neural networks display some characteristics that may cause detriment to their use in scientific discovery. The most obvious is their inherent complexity, which typically renders them un-interpretable. Furthermore, the complexity of a DNN trained on a particular task will almost always exceed the minimum model complexity actually required to conduct the task. Indeed, for any task, there typically exists a broad set of minimum error models - called a "Rashomon set" \cite{breiman_statistical_2001, semenova_existence_2022} - whose members may range from complex, un-interpretable models, to interpretable ones (Fig. \ref{fig:model_concept}a). DNNs typically represent the former. The challenge, then, is to find the latter.

Secondly, the ability of DNNs to leverage all statistical regularities present in the data exposes them to the risk of "shortcut learning". This is where a model learns some strategy for solving a problem that can be considered inappropriate to the problem domain. \citet{geirhos_shortcut_2020} give the example that "if cows happen to be on grassland for most of the training data, detecting grass instead of cows becomes a successful strategy for solving a classification problem [of cows] in an unintended way". They therefore warn that "we must not confuse performance on a \emph{dataset} with the acquisition of an \emph{underlying ability}". 

Hence, it may be necessary, when working in some domain, to specify those constraints that would determine the "domain-appropriateness" of a model. This may include some distinction between "admissible" and "in-admissible features". For example, when predicting planetary motions, admissible features for a model might be the position and mass of planets, while in-admissible features may include the zodiac, or the color of the planets. The identification of features used by a model may, in turn, require that it be interpretable.

Considering these risks, it is clear that to be "useful" as a scientific model, it is not sufficient that a model be "performant" in the sense of attaining high accuracy with respect to observations. The model should additionally be interpretable, such that human scientists can learn from it, and it should be appropriate within its domain of application (Figure \ref{fig:model_concept}b). We propose these three criteria of usefulness to further develop George Box's famous dictum that "all models are wrong but some are useful" \cite{box_robustness_1979}\footnote{Assessments of the utility of AI in science have been the subject of much discussion. For example, the U.S. Department of Energy defines "domain-awareness", "interpretability" and "robustness" as three key requirements for usefulness \cite{baker_workshop_2019}, while a \emph{Nature} survey of researchers concluded that the main negative scientific impact of AI was "more reliance on pattern recognition without understanding" \cite{van_noorden_ai_2023}. Related is the suitability or otherwise of AI models for use in high-stakes applications, where lack of interpretability can be dangerous \cite{rudin_stop_2019}.}.

\subsection{The third way: an ideal discovery system}
\label{section:third_way}

The basic premise of this study is that it is possible to leverage the benefits of deep learning while mitigating its drawbacks. The resulting method would be "free-form" in the sense of containing a broad, unconstrained model space, but it would also be minimally complex, and by that virtue, be interpretable, and provide maximum safeguards against shortcut learning.

This work is oriented toward the construction of such a method. The eventual goal is to the build a discovery system that can leverage the strengths of free-form techniques such as deep learning to construct models and theories that human scientists would find appealing, both in their elegance and their predictive power.

\subsection{Bioimaging as an ideal test domain}

Several factors conspire to make bioimaging an ideal domain of application for deep-learning-based scientific discovery. The introduction of tools such as high-throughput microscopy has led to an explosion in the availability of bioimaging data \cite{ouyang_imaging_2017, wollman_high_2007}. Furthermore, these data take the form of images, which typically include complex biological structures and a great deal of noise. They are therefore typically difficult to analyze without the aid of computational techniques \cite{driscoll_data_2021}. Moreover, the sheer complexity of biological systems such as living cells often necessitates the use of tools that can extract high-level patterns and regularities that may difficult to express in the kind of pristine mathematics found in the physical sciences. Again, deep learning is uniquely suited for the task.

\section{Prior work}
\label{section:prior_work}

Early work in computational discovery systems focused on the discovery of symbolic physical laws, with the \textsc{Bacon} and \textsc{Dalton} programs as prominent examples \cite{langley_scientific_1992, langley_bacon_1977, langley_data-driven_1981}. While these programs used pre-determined sets of production rules, later symbolic approaches focused on genetic evolution algorithms \cite{koza_genetic_1994, schmidt_distilling_2009, guimera_bayesian_2020, udrescu_ai_2020, cranmer_discovering_2020}. 

Symbolic methods typically operate on tabular data, but deep learning has facilitated the extraction of useful representations from raw data as well. This has involved the discovery of relevant "state variables" of an evolving physical system \cite{chen_automated_2022} and the use of representation learning to explore problems in molecular biology \cite{zaritsky_interpretable_2021, soelistyo_learning_2022, DBLP:conf/iccvw/SoelistyoCL23}. In the cell biology field, deep learning has also been used gain scientific insight into relationships between cell cycle phase and subcellular structure \cite{nagao_robust_2020}, cell morphology and motility \cite{nishimoto_predicting_2019}, cell motility and phenotype \cite{kimmel_deep_2021}, and more.

Many approaches enforce domain-appropriateness by adding explicit constraints embodying prior knowledge. In physics, this has involving the enforcement of known physical laws \cite{pun_physically_2019, chan_real-time_2020, karniadakis_physics-informed_2021} or the use of graph neural networks to explicitly model interactions between atoms \cite{xie_crystal_2018} or celestial objects \cite{lemos_rediscovering_2022}.

\section{Goal and strategy}
\label{section:strategy}

The aim of this study is to demonstrate and assess some methods for building an ideal discovery system, as described in \S\ref{section:third_way}. These methods will be assessed on a test problem, taken from the field of bioimaging. In doing so, we aim not only to assess the feasibility of such methods, but also to demonstrate that for this particular test problem, a broad Rashomon set indeed exists. This would in turn raise the potential that interpretable, high-performing models can be found for other applications in the field.

\subsection{The test problem: classifying chromatin morphology in live-cell microscopy data}
\label{section:test_problem}

The scientific question we aim to answer is, "what distinguishes a cell in interphase from one in metaphase?", where these are distinct stages of the cell cycle. For our dataset, we collected thousands of microscopy images of cells both in interphase and metaphase. Hence, the problem presented to our models was to place these images into the correct category, with the eventual goal of being able to inform us what criteria separate the two.

Prior knowledge informs us that the distinguishing characteristic relates to the organization of chromatin in the cells. When a cell is in interphase, the chromatin is distributed very diffusely around the cell. However, when it is in metaphase, the chromatin is aligned very sharply along an axis\footnote{When live cells undergo cell division, or mitosis, they pass through a series of stages in which their genetic material, or chromatin, adopts different forms of organization. In interphase, when the cell is not undergoing mitosis, the chromatin is distributed very diffusely around the cell. However, in the next stage, prometaphase, chromatin condensation begins to occur. When the cell enters metaphase, the chromatin, now organized into chromosomes, is aligned neatly along the cell equator by a structure called the mitotic spindle. Finally, during anaphase, the two chromatids produced by replication of each chromosome drift toward opposite poles of the cell.}.

\begin{figure}
 \centering
       \includegraphics[width=\textwidth]{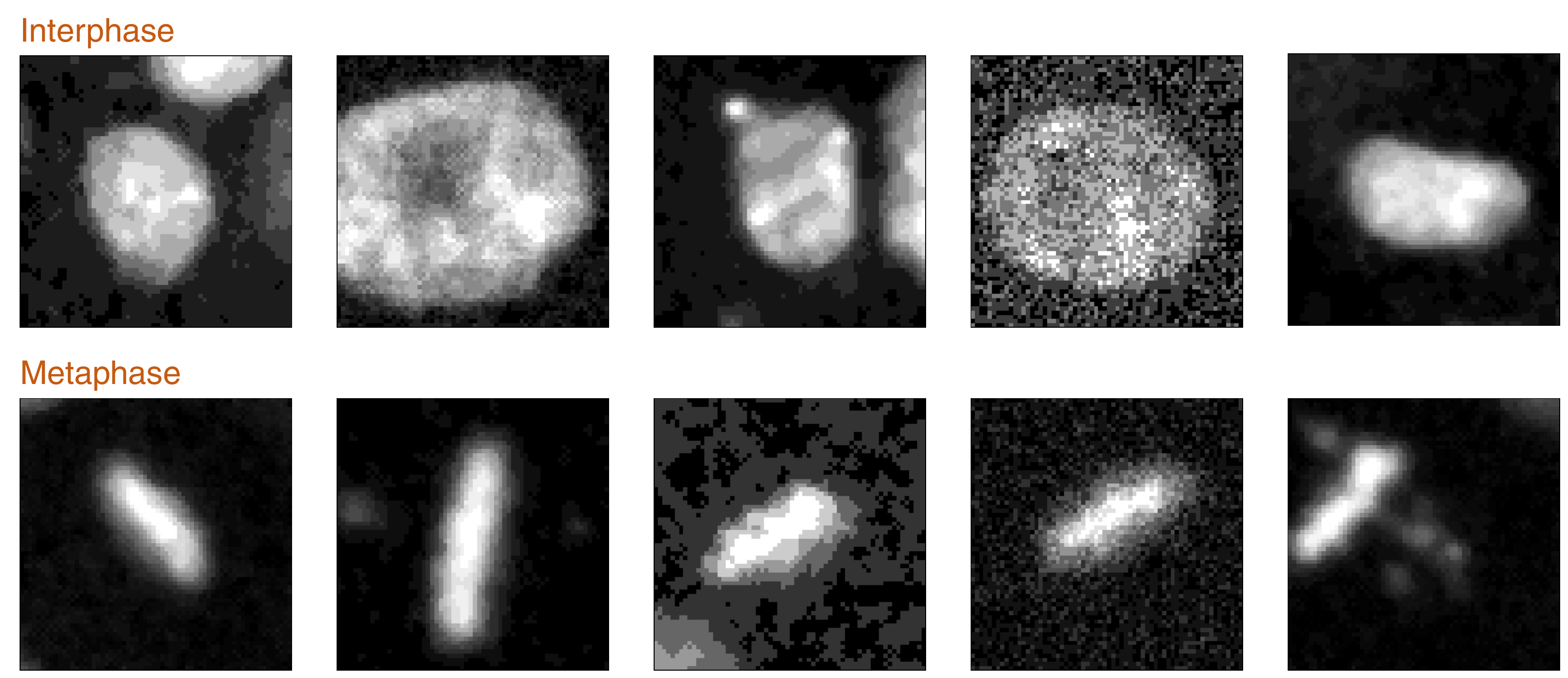}
       \caption{Example images for cells in interphase (top) and metaphase (bottom).}
    \label{fig:interphase_metaphase}
\end{figure}

The image dataset comprising cells in each of these two stages, was acquired using cell culture, high-throughput fluorescence microscopy and automated cell tracking as previously described \cite{bove_local_2017, ulicna_automated_2021, soelistyo_learning_2022, soelistyo_machine_2023}. In these experimental datasets, fluorescently tagged histone markers are used to visualize chromatin organization in cultured MDCK (Madin-Darby Canine Kidney) cells. Pixel intensities in the image dataset correspond to the density of chromatin (Fig. \ref{fig:interphase_metaphase}). Since chromatin is found predominantly in the cell nucleus, these image generally portray the nuclei of the cells.

The problem presented to our models is therefore a straightforward image classification problem; given an image, classify it as either a cell in interphase, or in metaphase. Despite the apparent simplicity of the problem, the dataset itself presents some challenges, including the abundance of noise and the diversity of situations in which we find the cells - for example, the nuclei are not always positioned at the exact center of the image, and there may be neighboring cells visible in the frame (Fig. \ref{fig:interphase_metaphase}). As such, the task is to identify the salient factors from the raw image data, and this is ideally suited to deep neural networks.

Prior knowledge would inform us that classification should be made on the basis of nuclear/chromatin morphology, including nuclear size and eccentricity. It is our objective to test whether this is the case for our models. Some domain constraints we seek to assess are:
\begin{enumerate}
    \item \textbf{Classifications should be based on central cell morphology:} In our images, cell nuclei in the neighborhood of the central cell may come into view. We would prefer our models to classify based on morphology of the central cell alone, and not aspects of this neighborhood.
    \item \textbf{Classifications should not be hyper-sensitive to pixel variations:} In general, models that capture some natural phenomenon should be sensitive to perturbation only at scales at which the underlying phenomenon is sensitive. In our case, it would generally take a significant alteration of chromatin morphology to legitimately transform an interphase image into a metaphase image and vice versa.
    \item \textbf{Classifications should not depend on spatial orientation of the image:} Model output should be invariant to spatial transformations such as image rotation and flipping. This is because these factors in turn depend only on the spatial orientation of the microscope, and not on the underlying biological system itself.
\end{enumerate}

Our dataset consisted of 3929 metaphase images and 4092 interphase images. We used a $90\%:10\%$ split between our training and testing sets.

\subsection{The strategy}

We chose three main methods for increasing the parsimony and interpretability of our deep-learning-based models:
\begin{enumerate}
    \item \textbf{Disentangled representation learning:} The discovery of a semantic latent representation, whose elements correspond to separate concepts. Representation learning models can transform raw data into semantically meaningful data, which can be used for downstream tasks such as classification. For this, we use a $\beta$-TCVAE \cite{chen_isolating_2019}.
    \item \textbf{Sparse neural network training:} Training of minimally connected neural networks that select inputs discriminately and minimize the complexity of the learnt function.
    \item \textbf{Symbolic regression:} Discovery of high-performing symbolic expression models, using the latent features deemed relevant by the sparse training procedure. Expressed in the language of mathematics, these expressions are highly interpretable.
\end{enumerate}

Our general approach is to train multiple models on the test problem, including some that use these methods and some that do not. We then analyze these models to assess the impact of these methods on the criteria of performance, interpretability and domain-appropriateness. For the latter criterion, we employ adversarial attacks to discover those image perturbations that can induce changes in classification, then compare those with our pre-established domain constraints (\S\ref{section:test_problem}).

To this end, we train four different model types, of differing interpretability, on our test problem. In order of increasing interpretability, they are:
\begin{enumerate}
    \item \textbf{Scheme 1: Dense CNN + Dense Head}: Dense convolutional neural network, followed by a dense classification head comprised of fully-connected layers.
    \item \textbf{Scheme 2: }$\mathbf{\beta}$\textbf{-TCVAE + Dense Head}: Convolutional $\beta$-TCVAE followed by a fully-connected dense classification head.
    \item \textbf{Scheme 3: }$\mathbf{\beta}$\textbf{-TCVAE + Sparse Head}: Convolutional $\beta$-TCVAE followed by a sparse classification head.
    \item \textbf{Scheme 4: }$\mathbf{\beta}$\textbf{-TCVAE + Symbolic expression}: Convolutional $\beta$-TCVAE, whose latent variables are related to the model output by a symbolic expression.
\end{enumerate}

Scheme 1 models are completely un-interpretable; they act as a baseline for the interpretability assessment of Scheme 2-4 models. These latter models rely on a semantic latent space. Scheme 2 models use a highly complex classification function while Scheme 3 \& 4 models use simpler functions. Crucially, Scheme 4 models are completely interpretable - they express the classification function as a mathematical expression based on the $\beta$-TCVAE latent representation. Scheme 3 models are interpretable as well, albeit, as we shall see, to a more limited degree (see \S\ref{section:sparse_network_analysis}). 

\section{Methods}
\subsection{Total Correlation VAE}
\label{section:methods_tcvae}

The Total Correlation Variational Autoencoder ($\beta$-TCVAE) \cite{chen_isolating_2019} is a variant of the variational autoencoder (VAE) \cite{kingma_auto-encoding_2022}. The VAE is a latent variable model that consists of an encoder network that compresses the input data into a probabilistic latent representation, and a decoder network that reconstructs the original input from this latent vector. The loss function used to train the VAE is the sum of a reconstruction loss term, which penalizes the difference between the reconstruction and the original input, and a "distribution" term that penalizes the Kullback-Leibler divergence between the latent probability distribution and a unit Gaussian prior. This latter term regularizes the latent space such that smooth traversals in this space correspond to smooth traversals in the input space.

The $\beta$-TCVAE is a variant of the VAE that has been designed specifically to produce disentangled latent spaces, where separate latent variables encode separate concepts. When applied on images, these may correspond to visual concepts such as size and shape. Mathematical details for the $\beta$-TCVAE can be found in Appendix \ref{section:extended_methods_tcvae}.

We trained the model on roughly 2.1 million images randomly sampled from our microscope footage. We extracted $64\times64$ pixel crops around each cell, which corresponds to roughly $21.3{\mu}$m along each side.

\subsection{Sparsity: RigL}
\label{section:methods_rigl}

For sparse neural network training, we use a dynamic pruning algorithm known as \emph{RigL} \cite{evci_rigging_2019}. This algorithm is premised on the "lottery ticket hypothesis", which states that dense neural networks will contain sub-networks that, when trained in isolation, can achieve test performance that matches the original dense network \cite{frankle_lottery_2019}. The general aim of sparse training algorithms is to identify this sub-network\footnote{A broad review of sparsity techniques is provided by \citet{hoefler_sparsity_2021}.}. \emph{RigL} achieves this by dynamically pruning and re-growing connections at fixed intervals during training. When "pruned", a connection weight is set to zero, and it ceases to update during training.

We modify this algorithm to optimize it for our application. Specifically, we introduce a "warm-up" period at the beginning of training where the network is trained densely, and we implement two post-training pruning steps that remove unnecessary connections from the final network. Details of the algorithm can be found in Appendix \ref{section:extended_methods_rigl}.

\subsection{Symbolic regression}
\label{section:methods_symbolic_regression}

Symbolic regression is a method to identify analytic expressions that approximate the output of an arbitrary function or dataset. We used
\textsc{PySR} \cite{cranmer_pysr_2023}, an open-source symbolic regression package that runs a genetic evolutionary algorithm \cite{koza_genetic_1994} to optimize symbolic expressions with respect to some loss function. More precisely, it converges on solutions by producing populations of symbolic expressions - described as expression trees - then alternating between rounds of tournament selection and mutation to obtain gradually better expressions. Here, the criterion for "fitness" is some combination of a pre-specified loss function (e.g., mean squared error) and a complexity metric. Complexity is calculated by summing the node-specific complexity score over all nodes in an expression tree, which depends on the identity of the node (i.e., the specific operator or constant that the node represents). The full algorithm can be found in \citet{cranmer_pysr_2023}.

\subsection{Adversarial attacks}
\label{section:methods_fgsm}

To assess the robustness of our classification networks, we implement a simple and efficient adversarial attack known as the Fast Gradient Sign Method (FGSM) \cite{goodfellow_explaining_2015}. This attack transforms some input data $\mathbf{x}$ with predicted class label $y$, such that the perturbed data $\tilde{\mathbf{x}}$ is classified by the model into another class $\tilde{y}$. FGSM forms $\tilde{\mathbf{x}}$ by adding to $\mathbf{x}$ some perturbation $\boldsymbol{\eta}$; i.e., $\tilde{\mathbf{x}} = \mathbf{x} + \boldsymbol{\eta}$. This perturbation is calculated based on the sign of the gradient of the loss function $L$ with respect to the input $\mathbf{x}$:
\begin{equation}
    \boldsymbol{\eta} = \epsilon \text{ sign}(\nabla_{\mathbf{x}}L(\boldsymbol{\theta}, \mathbf{x}, y)),
\label{eq:fgsm}
\end{equation}
where $\epsilon$ is the pre-specified perturbation magnitude and $\boldsymbol{\theta}$ represents the network parameters.

\section{Results}

\subsection{Disentangling image factors}

\begin{figure}
 \centering
       \includegraphics[width=\textwidth]{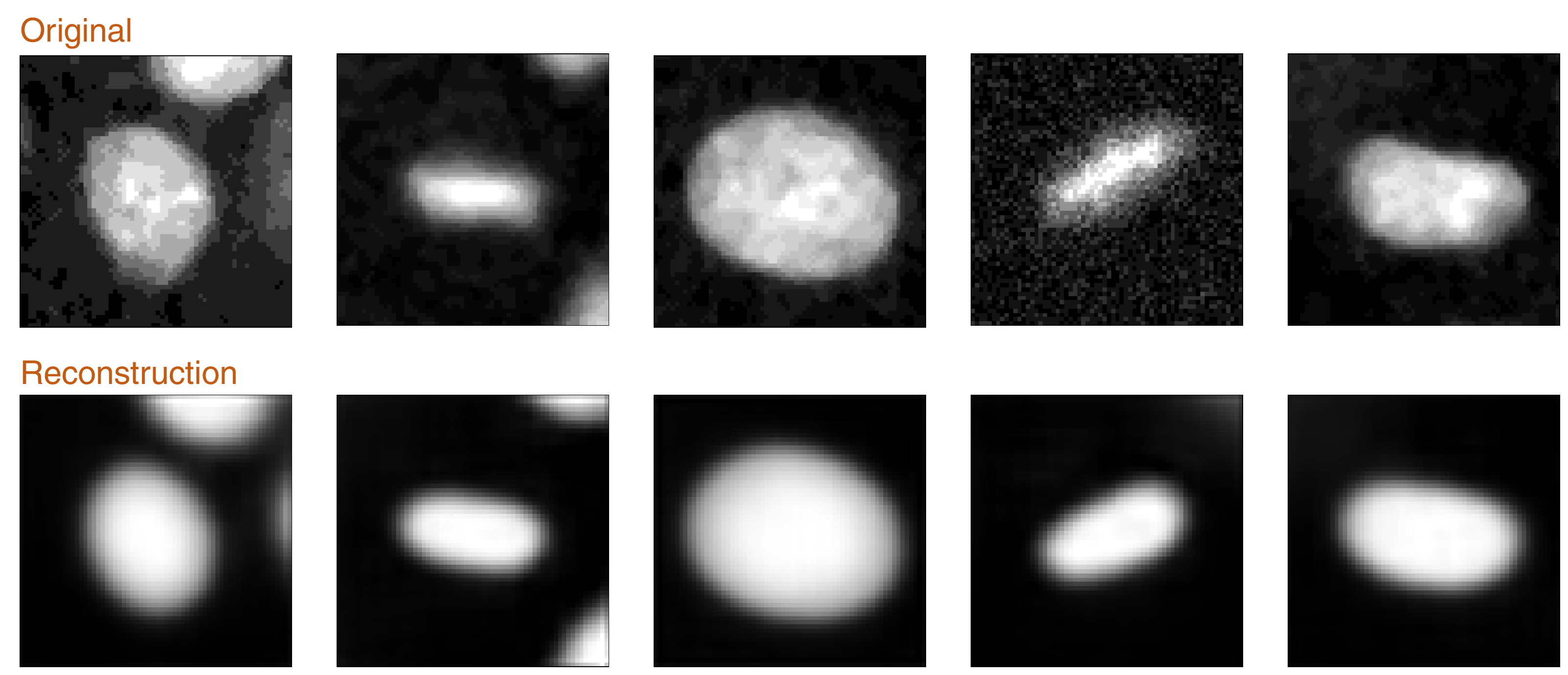}
       \caption{Example reconstructions produced by the $\beta$-TCVAE.}
    \label{fig:reconstructions}
\end{figure}

When trained, the reconstructions produced by the $\beta$-TCVAE sufficiently captured the distinct morphological features of the cells (Fig. \ref{fig:reconstructions}) in the dataset. Moreover, the $\beta$-TCVAE managed to extract disentangled latent features that were, to a large degree, interpretable. These included four features that encode central cell morphology (Fig. \ref{fig:latent_variables_cc}), two that encode central cell position, sixteen that encode neighborhood features, and a further ten whose traversals do not produce perceptible changes in image space (some position and neighborhood features are shown in Appendix \ref{section:extended_traversals}). We interpreted these latent features using traversals in latent space.

\newcommand{\figwidth}{0.75}
\begin{figure}
 \centering
    \includegraphics[width=\figwidth\textwidth]{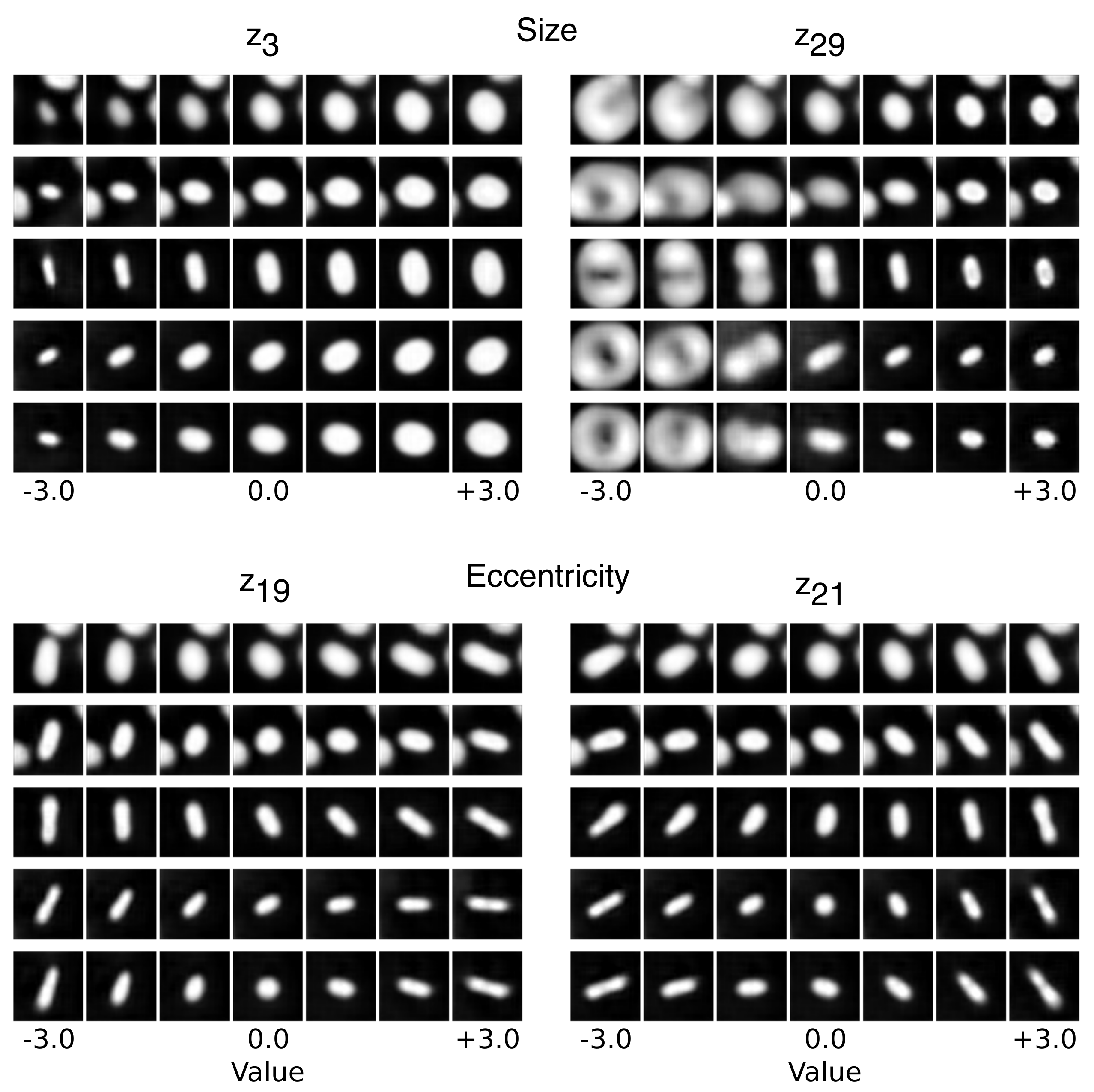}
       \caption{Latent variables that encode central cell morphology.}
    \label{fig:latent_variables_cc}
\end{figure}

Variable $z_{17}$ appears to encode central cell aspect ratio, while $z_{21}$ has an equivalent function, but for the diagonal axes (hence, we term it "diagonality"). Both $z_3$ and $z_{29}$ appear to encode central cell size. Initially, we had expected that cell size would be encoded by only one variable. However, after training several $\beta$-TCVAE models with the same hyperparameters, we observed that this separation persistently appears. For this particular model, $z_3$ appears to capture the specific effect on cell size of neighborhood crowding. Hence, decreases in cell size are accompanied by corresponding increases in local cell density (Fig. \ref{fig:latent_variables_cc}). Moreover, there is an upper limit on cell size that can be captured by $z_3$. Meanwhile, $z_{29}$ appears to capture a neighborhood-independent representation of cell size. This can be seen in Fig. \ref{fig:latent_variables_cc}a, where changes in cell size are mostly unaccompanied by changes in the neighborhood (notwithstanding some small alterations). Additionally, $z_{29}$ is able to capture a larger range of cell size than $z_3$.

Prior knowledge would inform us that only the four central cell morphology features would be relevant to cell state classification. The challenge was to assess whether our classification models conform to this assumption.

\subsection{Cell state classification}
\label{section:classification}

For classification, we investigated the four model schemes introduced in \S\ref{section:strategy}. In Scheme 1, a Dense CNN reduces the input image to a feature vector, which is then processed by the dense fully-connected head. In Schemes 2, 3 \& 4, the input is the latent representation of the image produced by the $\beta$-TCVAE. For implementation details, see Appendix \ref{section:extended_methods_classification}. All models $f$ reduce the input $x$ to a single output scalar $f(x)$, which is used to make the classification $y$ following:
\begin{equation}
y = 
\begin{cases}
    \code{interphase},& \text{if    } f(x) < 0\\
    \code{metaphase},& \text{if } f(x) \geq 0.
\end{cases}
\end{equation}

For training Scheme 3 models, we found the optimal sparsity level using the hyper-parameter search program \textsc{Optuna}, as described in Appendix \ref{section:sparsity_tuning}. For each scheme, we trained ten models (Table \ref{table:test_performance}). More details on the acquisition of Scheme 4 models (involving symbolic expressions) can be found in \S\ref{section:decision_boundary}. Strikingly, the enormous decrease in complexity of the classification head from Scheme 2 to 4 is accompanied only by a relatively minor decrease in performance, as measured by testing accuracy. For example, on average, Scheme 3 models attain 98\% of the accuracy of Scheme 2 models, with only 2.2\% of the parameter count and 2.1\% of the expression size. Meanwhile, Scheme 4 models also attain about 98\% of the accuracy of Scheme 2 models, but with only 0.2\% of the expression size.

\begin{table}[h!]
\begin{center}
\begin{tabularx}{0.9\textwidth}{cccccc}
 \hline
 \hline
 \textbf{Scheme} & \textbf{Encoder} & \textbf{Head} & 
 \textbf{No. of head parameters} &
 \textbf{Head expression size} & \textbf{Accuracy} \\
 \hline
 1  & CNN & Dense & $1040$  & $9697$ & $99.7 \pm 0.1\%$ \\

 2  & VAE & Dense & $1040$  & $8641$ & $99.0 \pm 0.1\%$ \\

 3  & VAE & Sparse & $23 \pm 2$  & $180 \pm 20$ & $97.0 \pm 0.2\%$\\

 4  & VAE & Symbolic & N/A  & $17 \pm 7$ & $97.4 \pm 0.2\%$\\
\hline
\hline
\end{tabularx}
\vspace{0.5cm}
\caption{Testing performance across ten models within each scheme. Errors represent the standard deviation across ten models from each scheme. No. of head parameters is calculated by summing the number of active connection weights. "Head expression size" is the number of nodes in the expression tree equivalent to the model concerned. For Scheme 1-3, head expression size is calculated according to the method outlined in Appendix \ref{appendix:head_size_calculation}.}
\label{table:test_performance}
\end{center}
\end{table}

\subsection{Sparse network analysis}
\label{section:sparse_network_analysis}

\begin{figure}
 \centering
 \includegraphics[width=0.5\textwidth]{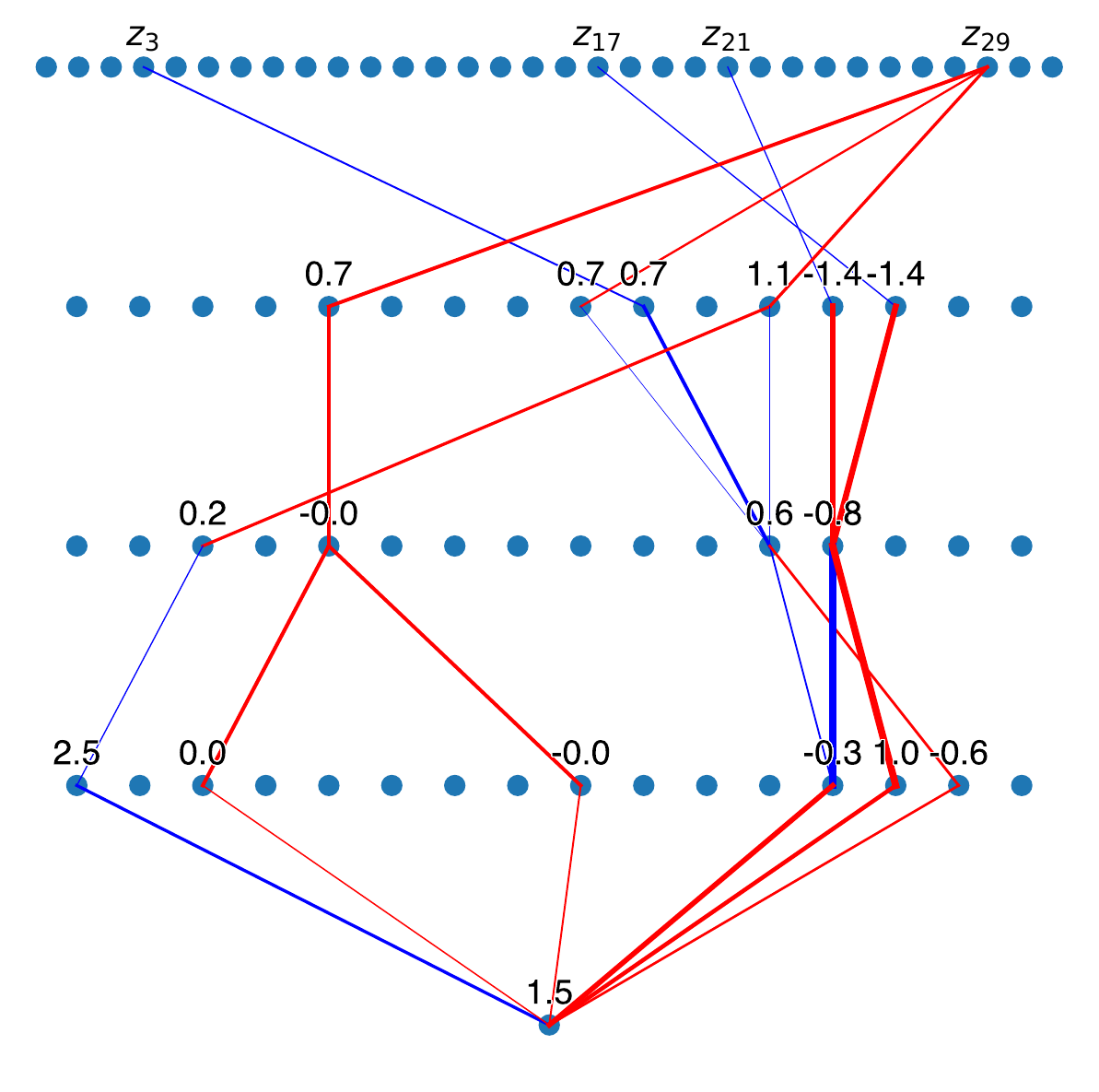}
       \caption{Topology of the highest-performing Scheme 3 model. Blue connections are positively weighted, red connections are negatively weighted. The thickness of the connection line is proportional to its weight magnitude. Bias values are written above their respective neuron. The network flow is up to down, so the top layer is the input layer and the bottom layer is the output layer.}
    \label{fig:scheme3_8_topology}
\end{figure}

Can we inspect the connectivity of our sparse models to interpret their operation? As a test example, we chose to examine our highest-performing Scheme 3 model, which achieved 97.3\% test accuracy. The topology of this model is shown in Fig. \ref{fig:scheme3_8_topology}.

Strikingly, we find that the model has learnt that the minimal set of required input features corresponds exactly to the latent variables that encode central cell morphology (Fig. \ref{fig:latent_variables_cc}) and ignored those describing the neighborhood (Fig. \ref{fig:latent_variables_neighborhood}). In fact, this was true of all ten of our Scheme 3 models (see Appendix \ref{appendix:sparse_model_topologies} for illustrations of the topologies). Thus, our modified \emph{RigL} procedure, paired with hyper-parameter optimization with \textsc{Optuna}, was able to find both the number and selection of input variables that prior knowledge would deem relevant to the task, and with impressive consistency.

From Fig. \ref{fig:scheme3_8_topology}, we can also see that the eccentricity terms $z_{17}$ and $z_{21}$ contribute to the network output through a set of connected paths, or a "stream" (Fig. \ref{fig:streams}a). Meanwhile, the size terms $z_{3}$ and $z_{29}$ also contribute through a stream that in the 3rd layer interferes with the eccentricity stream (Fig. \ref{fig:streams}b). 

\newcommand{\expwidth}{0.43}
\begin{figure}
 \centering
    \begin{subfigure}[b]{\expwidth\textwidth}\includegraphics[width=\textwidth,trim={2.5cm 2.5cm 2.5cm 2.5cm},clip]{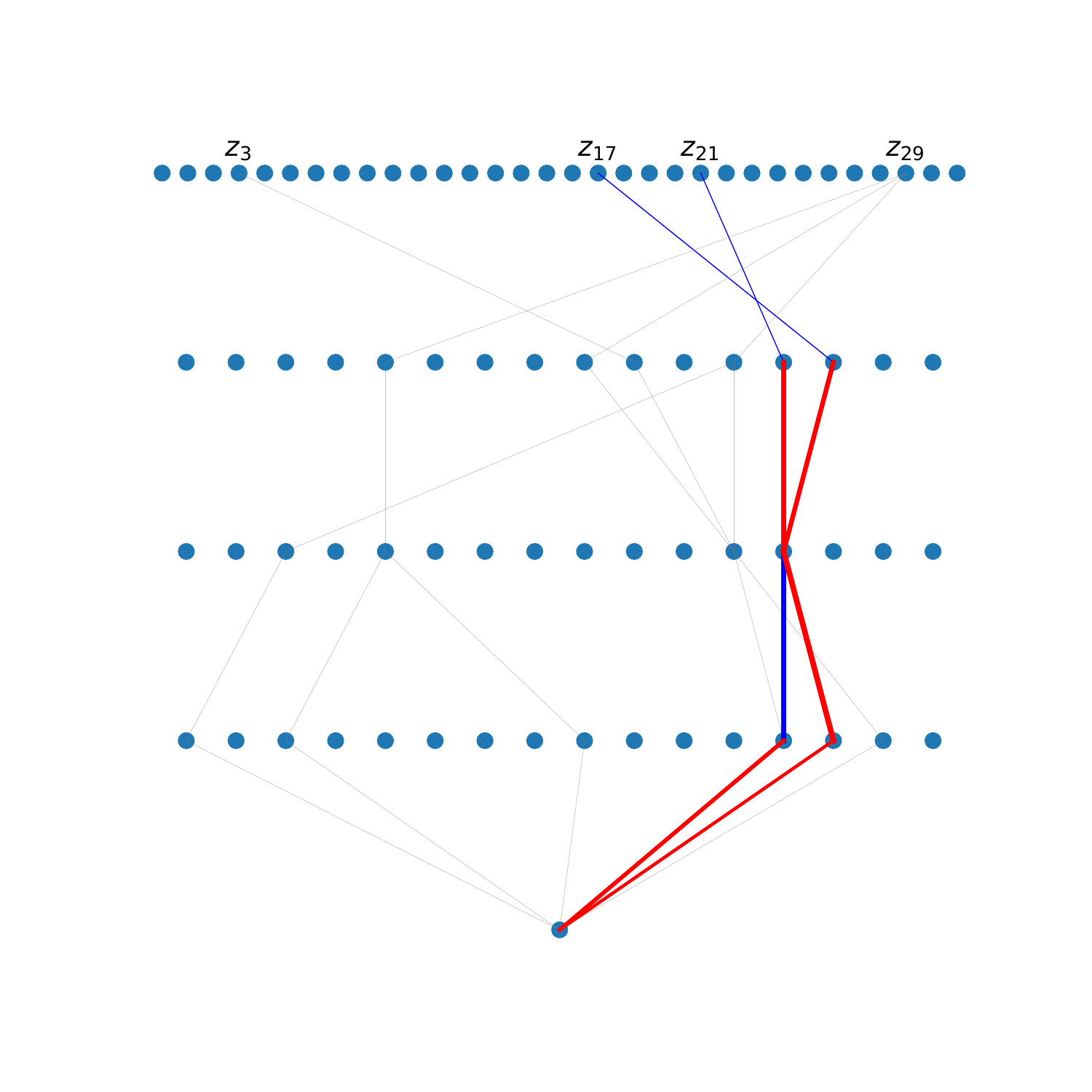}
      \caption{Eccentricity ($z_{17}$ \& $z_{21}$) stream.}
    \end{subfigure}
    \begin{subfigure}[b]{\expwidth\textwidth}\includegraphics[width=\textwidth,trim={2.5cm 2.5cm 2.5cm 2.5cm},clip]{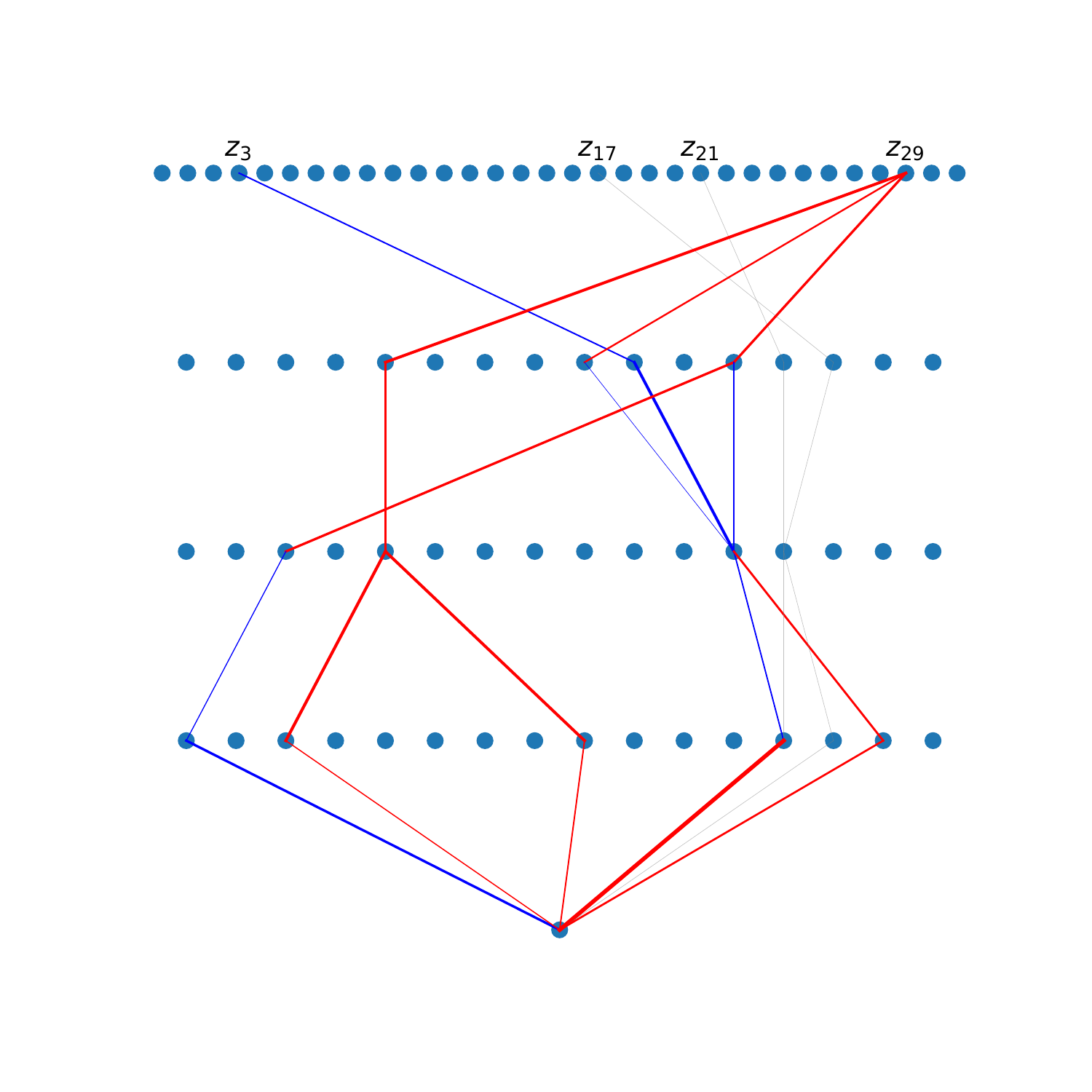}
      \caption{Size ($z_{3}$ \& $z_{29}$) stream.}
    \end{subfigure}
    \caption{Sets of connection paths, or "streams" relating the network inputs to the output neuron. Grayed out connections belong the network but not to the stream.}
    \label{fig:streams}
\end{figure}

To make sense of our sparse network, we first separated it into several sub-networks, then plotted the output of each sub-network - its "response" - as a function of its inputs. The result is a "response map" analogous to the tuning curves employed in neuroscience. This approach is possible due to the sparsity of the network, which enables decomposition into simple, low-dimensional components that may be responsible for different, identifiable functions.

The first sub-network we examined was the eccentricity-independent component of the cell size stream, which happens to include only $z_{29}$ as an input (Fig. \ref{fig:sub_network_size}a). We can see from the response curve (Fig. \ref{fig:sub_network_size}b) that high values of $z_{29}$ - indicating a lower size - generally correspond to higher response values. However, the behavior is non-monotonic, which may suggest some unnecessary complexity in the response function. Finally, we can see that the response of this sub-network is wholly constrained to the positive region, suggesting that the decisive operations must lie in other sub-networks.  

\renewcommand{\expwidth}{0.43}
\begin{figure}
 \centering
    \begin{subfigure}[b]{\expwidth\textwidth}
    \includegraphics[width=\textwidth,trim={2.5cm 2.5cm 2.5cm 2.5cm},clip]{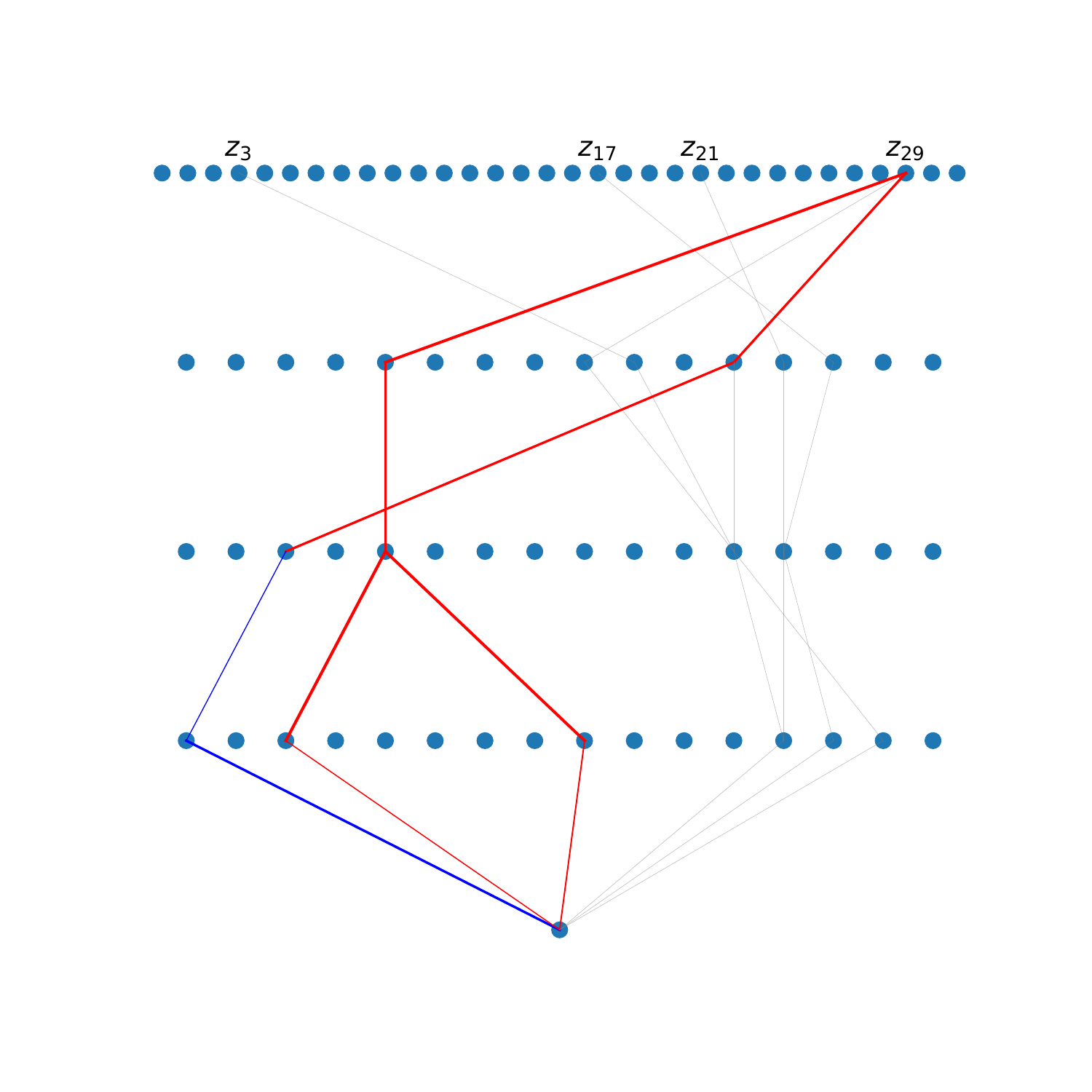}
      \caption{Sub-network topology. Grayed out connections belong the network but not to the sub-network.}
    \end{subfigure}
    \begin{subfigure}[b]{\expwidth\textwidth}
      \includegraphics[width=\textwidth]{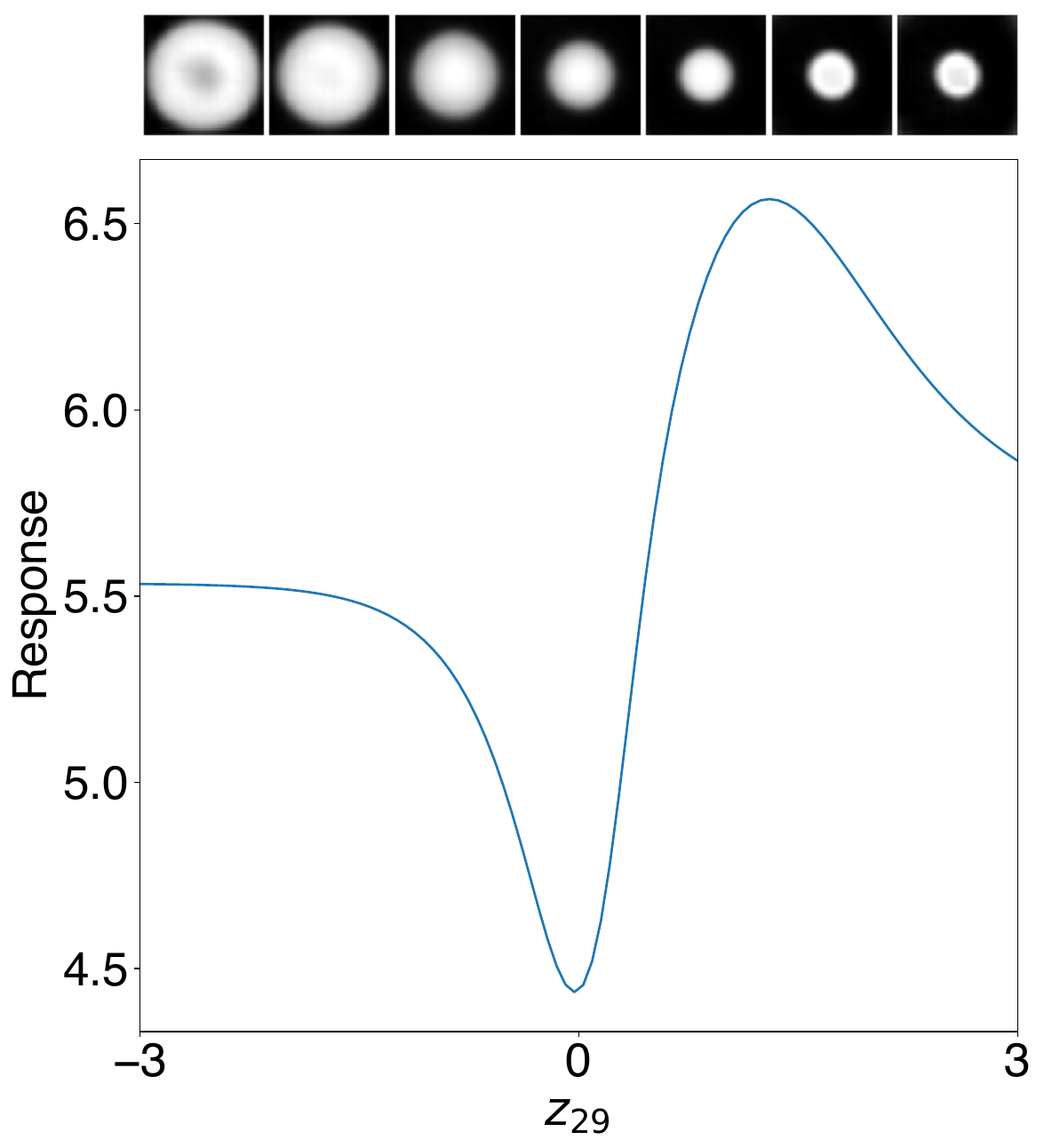}
      \caption{Response curve.}
    \end{subfigure}
    \caption{Sub-network 1. This sub-network responds to nuclear size.}
    \label{fig:sub_network_size}
\end{figure}

Given that there is some interaction between the size and eccentricity streams, we decided to decompose the remainder of the whole network into three sub-networks. The first of these - sub-network 2 - gathers input from the size variables and terminates in the third fully-connected layer (Fig. \ref{fig:sub_network_06}). The response of this sub-network can be interpreted as the size of the cell's chromatin signature - it generally increases with increasing $z_3$ and decreasing $z_{29}$.

\renewcommand{\expwidth}{0.43}
\begin{figure}
 \centering
    \begin{subfigure}[b]{\expwidth\textwidth}
    \includegraphics[width=\textwidth,trim={2.5cm 2.5cm 2.5cm 2.5cm},clip]{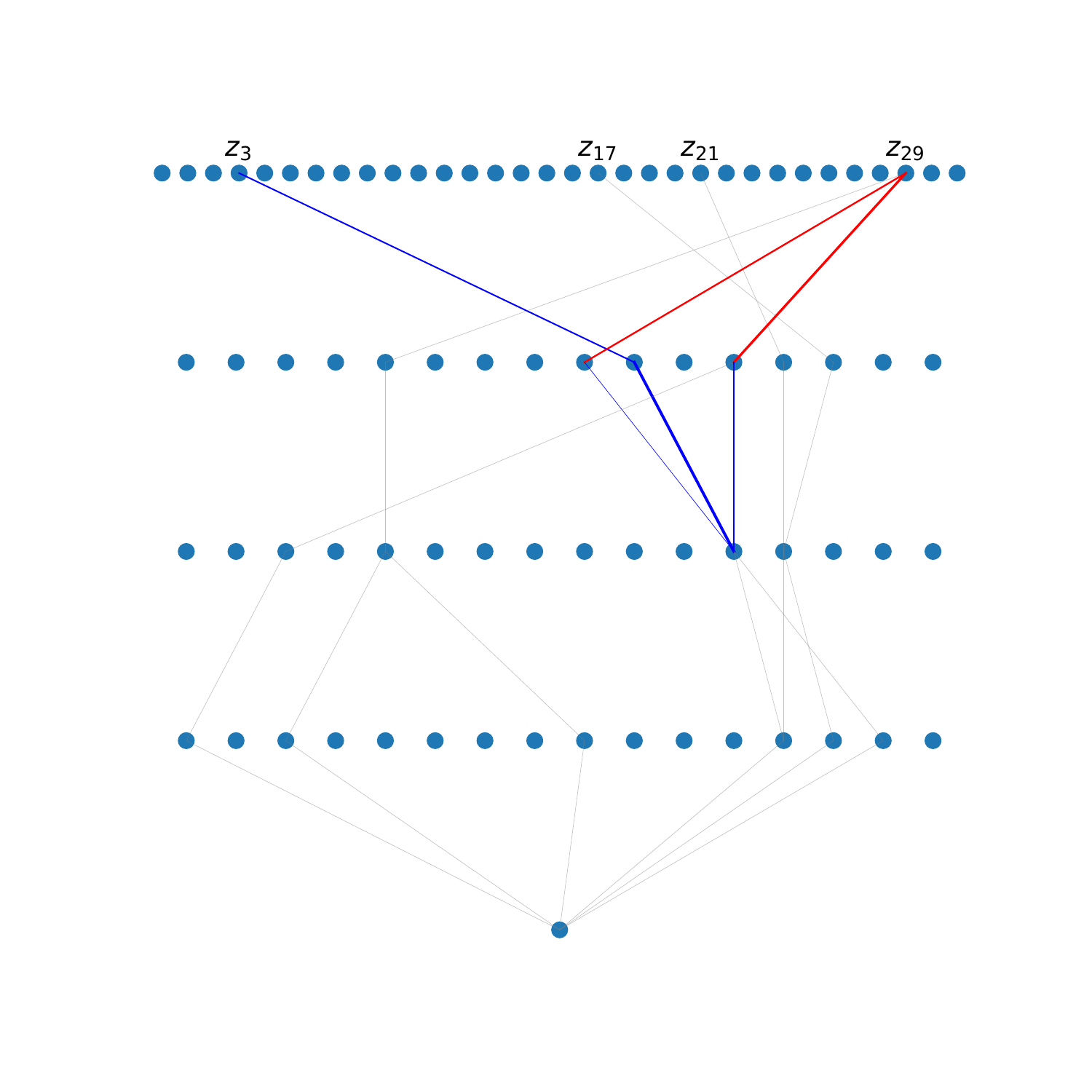}
      \caption{Sub-network topology.}
    \end{subfigure}
    \begin{subfigure}[b]{\expwidth\textwidth}
      \includegraphics[width=\textwidth]{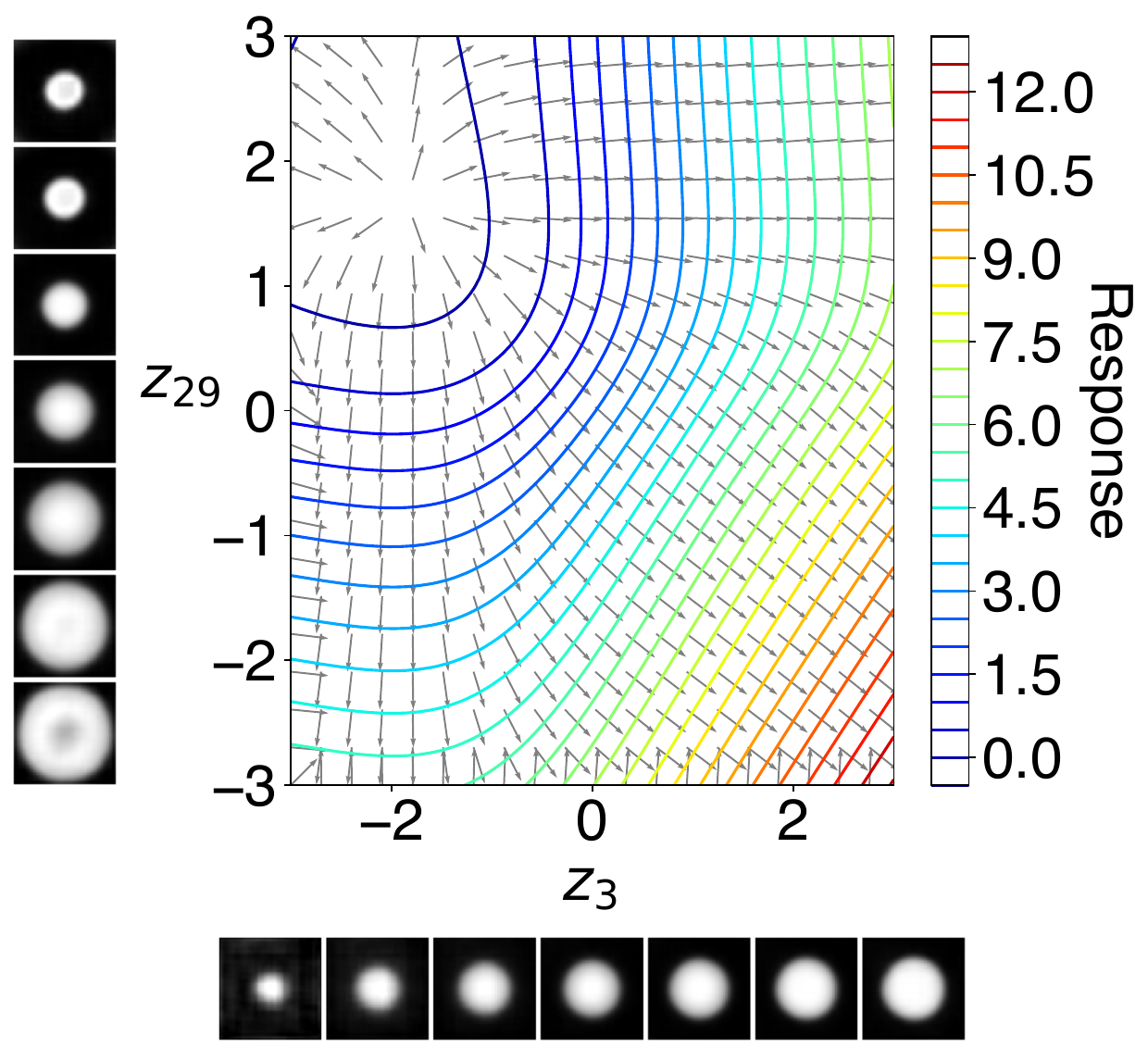}
      \caption{Response contour map.}
    \end{subfigure}
    \caption{Sub-network 2. This sub-network responds to nuclear size. Colored contour lines represent the response value at each point in the latent sub-space. Gray arrows represent the gradient of the response value. Images are decoded traversals in latent space along the axis of each latent variable.}
    \label{fig:sub_network_06}
\end{figure}

For our third sub-network, we chose the corresponding module for eccentricity (Fig. \ref{fig:sub_network_08}). This time, the response can be interpreted as a measure of "roundness"; it is greatest at $z_{17}, z_{21} = 0$ and decreases in radial fashion from the origin.

\renewcommand{\expwidth}{0.43}
\begin{figure}
 \centering
    \begin{subfigure}[b]{\expwidth\textwidth}
    \includegraphics[width=\textwidth,trim={2.5cm 2.5cm 2.5cm 2.5cm},clip]{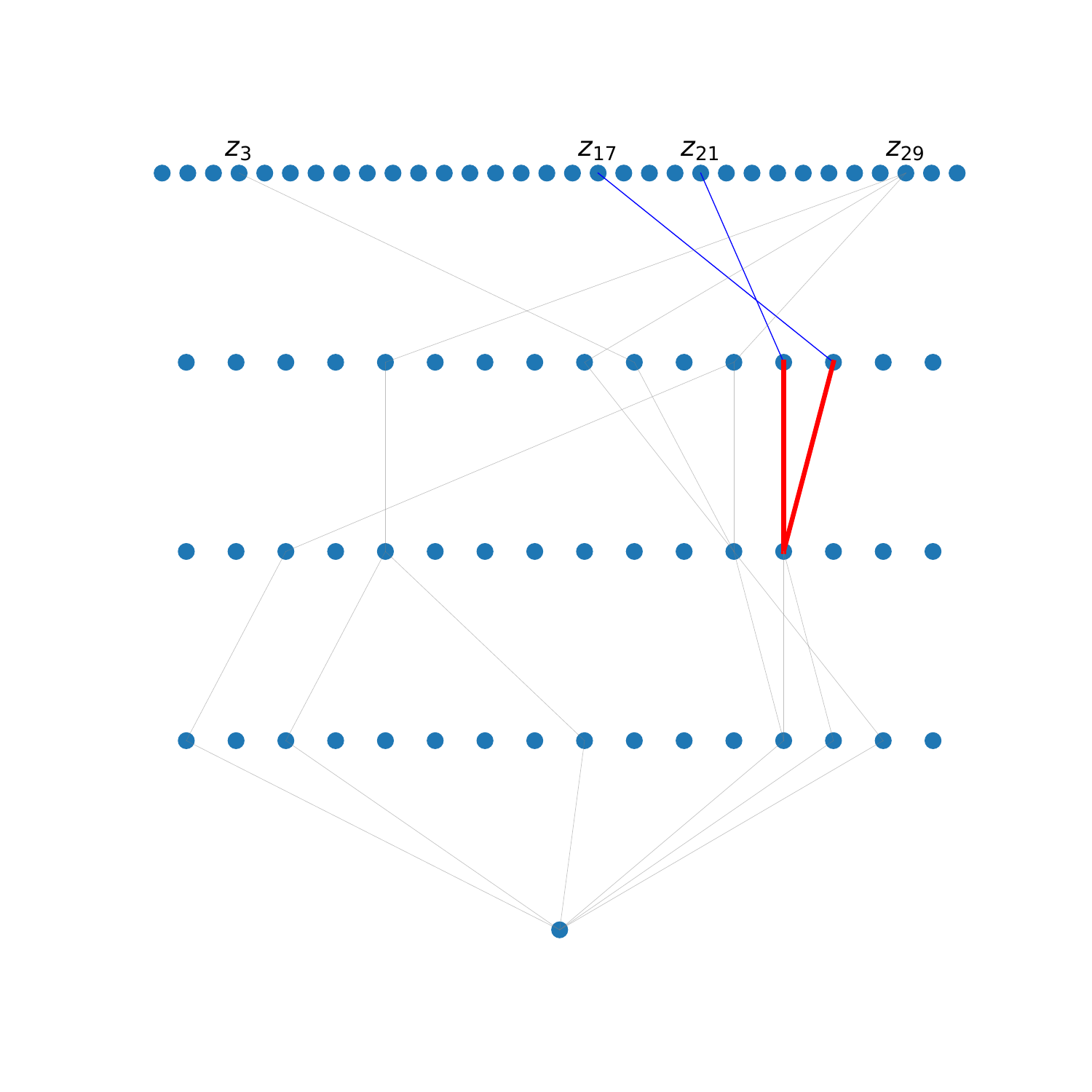}
      \caption{Sub-network topology.}
    \end{subfigure}
    \begin{subfigure}[b]{\expwidth\textwidth}
      \includegraphics[width=\textwidth]{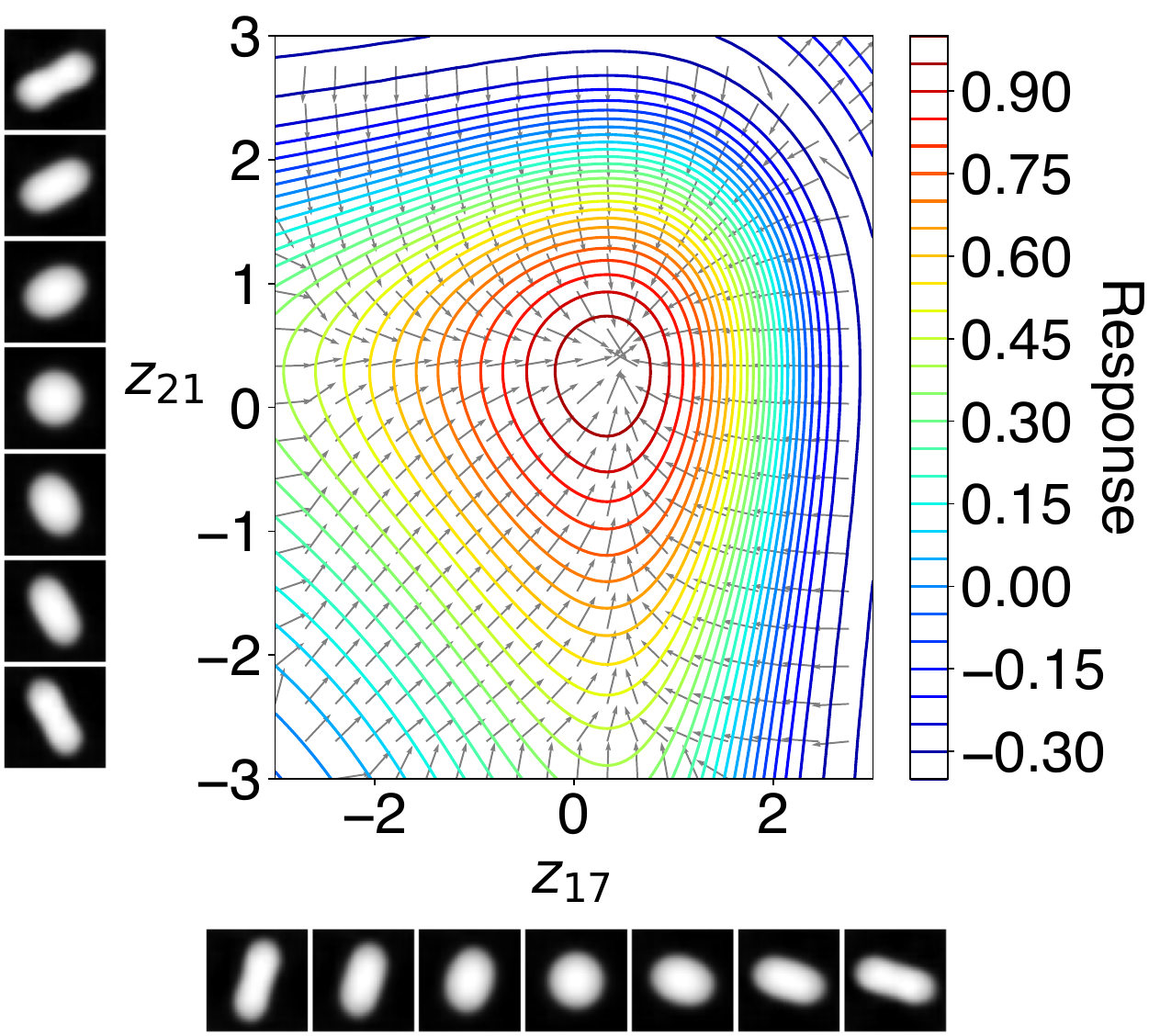}
      \caption{Response contour map.}
    \end{subfigure}
    \caption{Sub-network 3. This sub-network responds to nuclear eccentricity.}
    \label{fig:sub_network_08}
\end{figure}

Finally, our fourth sub-network receives the outputs of the second and third sub-networks (hereafter, "$z_{size}$" and "$z_{round}$" respectively) and processes them into a final contribution to the output of the whole network (Fig. \ref{fig:sub_network_comb}). We can immediately deduce that the the response value is most sensitive to $z_{size}$; for any value of $z_{round}$, changes in $z_{size}$ are sufficient to determine the classification. Sensitivity to $z_{round}$ is comparably smaller. Furthermore, for small values of $z_{size}$, it is non-monotonic, again suggesting some unnecessary complexity. Finally, we observe that the response of sub-network 4 is pre-dominantly negative; hence in most cases, the output of sub-network 1 is required to push the final output into the positive range, which entails classification of metaphase.

\renewcommand{\expwidth}{0.43}
\begin{figure}
 \centering
    \begin{subfigure}[b]{\expwidth\textwidth}
    \includegraphics[width=\textwidth,trim={2.5cm 2.5cm 2.5cm 2.5cm},clip]{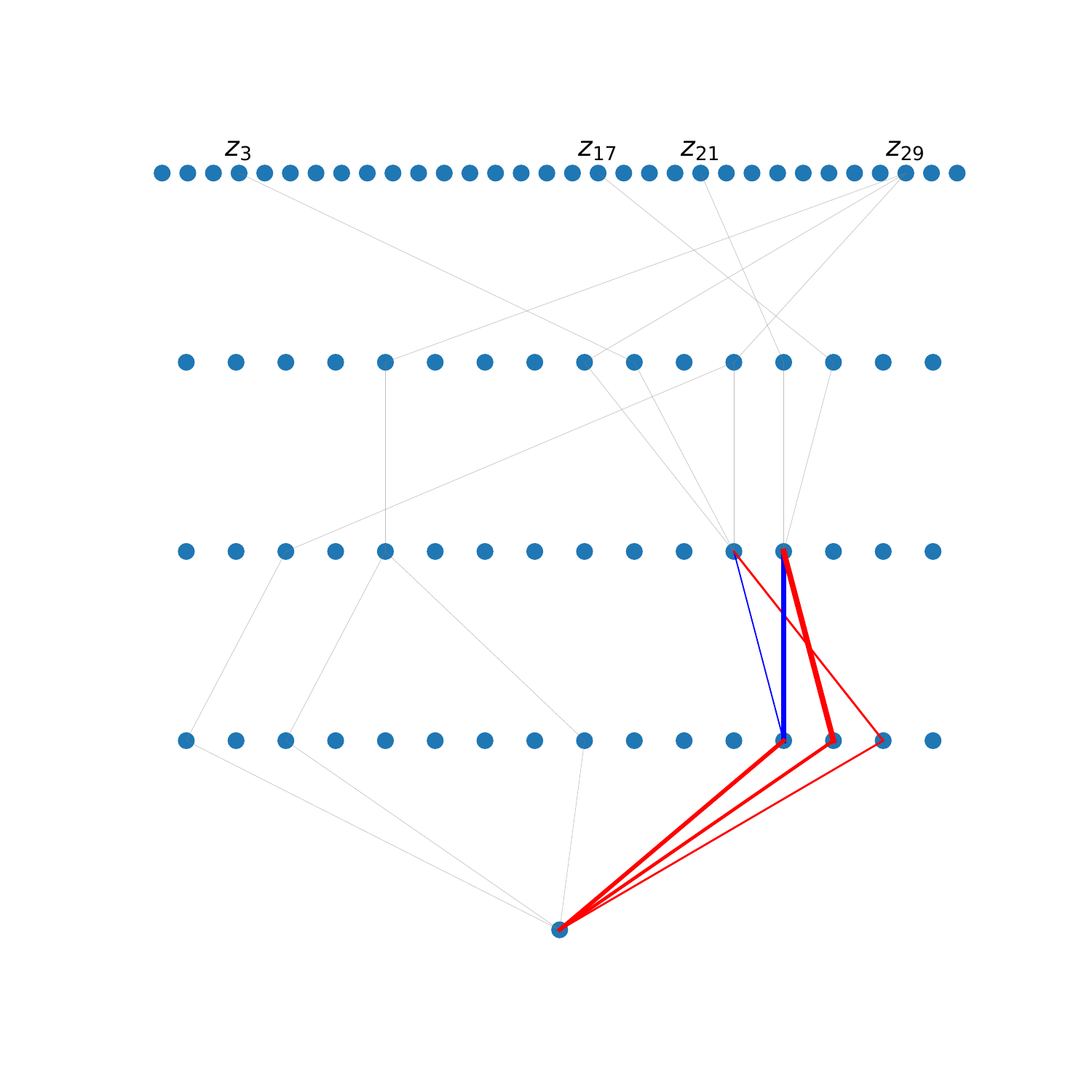}
      \caption{Sub-network topology.}
    \end{subfigure}
    \begin{subfigure}[b]{\expwidth\textwidth}
      \includegraphics[width=\textwidth]{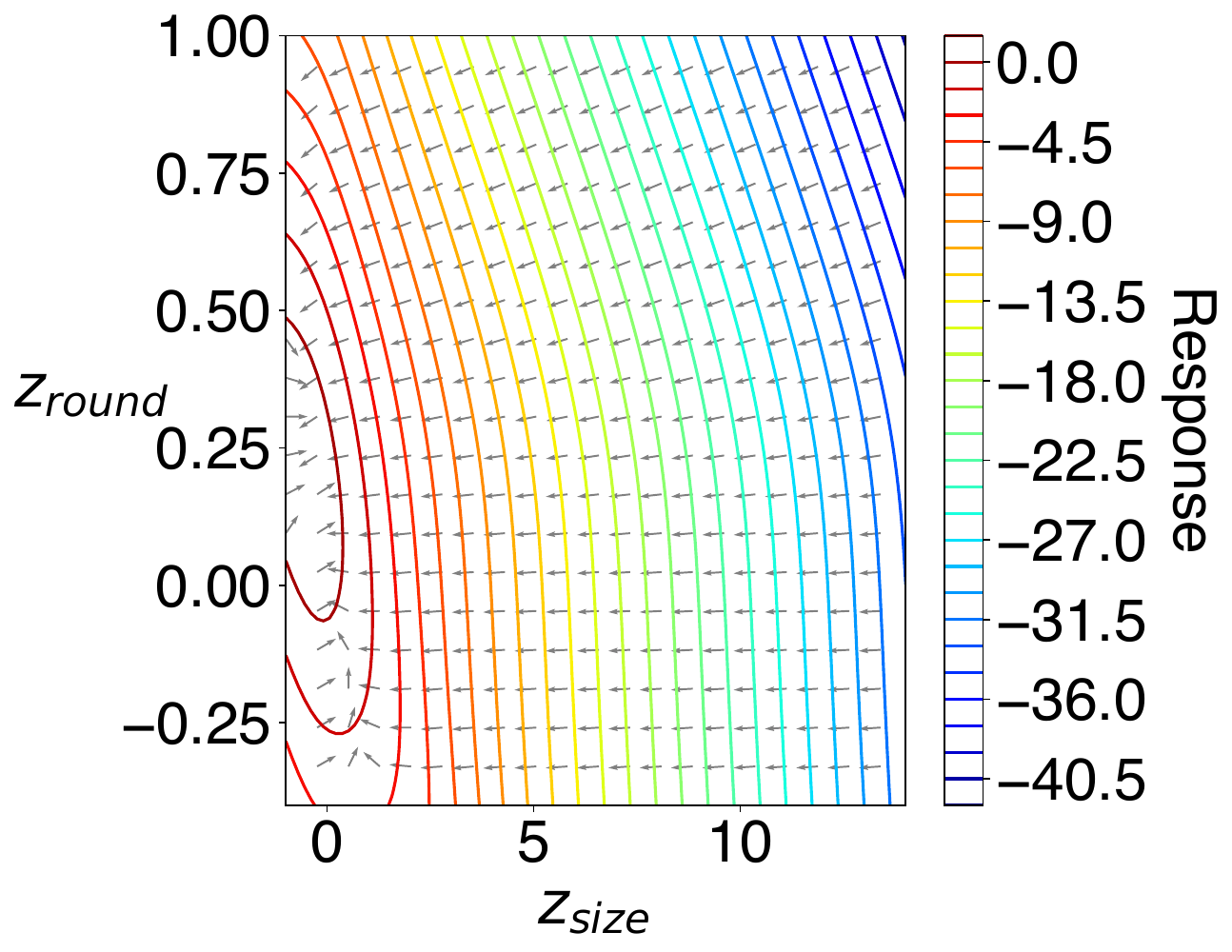}
      \caption{Response contour map. $z_{size}$ and $z_{round}$ refer to the output values of the second and third sub-networks respectively.}
    \end{subfigure}
    \caption{Sub-network 4. This sub-network responds to both size and eccentricity.}
    \label{fig:sub_network_comb}
\end{figure}

From our brief study of this particular sparse network, we can therefore list some of its behaviors:
\begin{enumerate}
    \item $\mathbf{z_{17}}$ \textbf{\&} $\mathbf{z_{21}}$ \textbf{are treated virtually equally.} These variables influence the final network output only through the response of sub-network 3 ("$z_{round}$"), and its response map is symmetric about the $z_{17}=z_{21}$ line (Fig. \ref{fig:sub_network_08}b). This fact is consistent with domain knowledge, as the two latent variables essentially represent the same physical feature, but in different rotational orientations.
    \item $\mathbf{z_{3}}$ \textbf{\&} $\mathbf{z_{29}}$ \textbf{are treated almost equally, though in opposite directions.} They influence the final network output primarily through "$z_{size}$", whose associated response map is roughly symmetric about the $z_{29}=-z_{3}$ line.
    \item $\mathbf{z_{29}}$ \textbf{is more important than }$\mathbf{z_{3}}$. This is due to the influence of $z_{29}$ on the output of sub-network 1 (Fig. \ref{fig:sub_network_size}).
    \item \textbf{Size terms }$\mathbf{z_{29}}$ \textbf{\&} $\mathbf{z_{3}}$ \textbf{are the primary determinants of classification.} Eccentricity terms $z_{17}$ \& $z_{21}$ influence the final output only through $z_{round}$, whose impact is marginal relative to that of $z_{size}$.
    \item \textbf{Metaphase classification is encouraged by high eccentricity.} For any fixed value of $z_{size}$, decreasing $z_{round}$ will increase the response value of sub-network 4 in most cases.
\end{enumerate}

In summary, we were able to gain significant insight into the behavior of our model due to its sparsity, which enabled us to decompose it into several sub-networks whose response functions can be studied in isolation or in tandem. Interpretation of the model is feasible because each sub-network accepts a very small number of input variables, which allows visualization of the response map over input space. Furthermore, the parsimony of the network kept the number of independent sub-networks to a feasible range. These criteria are completely absent in densely connected neural networks.

Nevertheless, analyzing sparse networks can be an onerous task, especially for domains or problems more complex than the one presented to our models here. Moreover, our analysis revealed some features of the model that we suspect may reflect unnecessary complexity, such as the non-monotonic response dependence of sub-network 4 on $z_{round}$ (Fig. \ref{fig:sub_network_comb}b) or that of sub-network 1 on $z_{29}$ (Fig. \ref{fig:sub_network_size}b). Such behaviors may reflect the data, but may also reflect accidental biases introduced by the model architecture.

In \S\ref{section:decision_boundary}, we explore how symbolic regression can address these problems, by further reducing the complexity of the model.

\subsection{Decision boundary discovery}
\label{section:decision_boundary}

Our final strategy to reduce the complexity of the model is to find simple analytic expressions that produce interpretable decision boundaries in input space, using symbolic regression. Apart from satisfying domain-specific constraints, the reduction of the input space from 32 to 4 dimensions significantly aided the symbolic regression procedure. This is because genetic algorithms (GAs) - such as that used by \emph{PySR} - tend to scale poorly with the number of input variables, owing to the fact that linearly increasing the number of input variables would exponentially increase the number of possible expression trees \cite{cranmer_pysr_2023}. Implementation details can be found in Appendix \ref{section:extended_methods_pysr}.

An important consideration is the loss function used by the GA to perform the symbolic regression (see \S\ref{section:methods_symbolic_regression}). We assessed two strategies for choosing the loss function and target output values:
\begin{enumerate}
    \item \textbf{Hinge loss}: Our target values $y'$ are the ground-truth labels of the cell states, expressed as $-1$ (for interphase) or $+1$ (for metaphase). Our loss function is the hinge loss \cite{rosasco_are_2004}, widely used in support vector machines, which is defined as:
    \begin{equation}
        \mathcal{L}(y, y') = \text{max}(0, 1-y'y).
    \end{equation}
    Hence, $y$ is encouraged to share the same sign with $y'$.  In this strategy, the neural network outputs are not used; however, we still use only the four latent variables identified as relevant by our Scheme 3 models.
    \item \textbf{MSE loss}: Our target values $y'$ are the outputs of a neural network - in this case, we use our Scheme 3 models. Our loss function is mean-squared error:
    \begin{equation}
        \mathcal{L}(y, y') = (y' - y)^2.
    \end{equation}
    Hence, our symbolic expressions are encouraged to adhere as closely as possible to the function learnt by the neural network.
\end{enumerate}
Overall, we found slightly greater performance with the hinge loss method than with the MSE loss method; however, this difference is minute (roughly $<$1\% accuracy). A collection of these models is shown in Appendix \ref{section:extended_symbolic}.

A glance at their expression forms reveals some patterns:
\begin{enumerate}
    \item \textbf{In all Hinge loss models, }$\mathbf{z_{17}}$ \textbf{\&} $\mathbf{z_{21}}$ \textbf{are represented only in the term ($\mathbf{z^2_{17} + z^2_{17}}$) or ($\mathbf{|z_{17}| + z^2_{17}}$)}. This displays a tendency to use the two eccentricity variables in the same way. Moreover, what is relevant is the sum of their absolute value; in other words, what is relevant is the \emph{degree} of eccentricity, not the precise axis of elongation. This adheres to the domain constraint that spatial orientation should not matter.
    \item \textbf{In most models, output score is monotonically increased by increasing }$\mathbf{z_{29}}$ \textbf{and decreased by increasing }$\mathbf{z_{3}}$. The models have learnt that small size is a strong indicator of metaphase. Curiously, the regression models found that it was possible to obtain 79\% accuracy simply using the expression "$z_{29}$",
    and that it was possible to obtain 91\% accuracy using expressions that only contain $z_{29}$.
    \item \textbf{MSE loss models show a greater tendency toward additive structures}. In 7 out of 10 MSE loss expressions, terms involving each input variable can be separated. This is not true of any of the Hinge loss models.
\end{enumerate}

We further analyze two hinge loss models and two MSE loss models, in order to get a sense of the striking diversity of forms these expressions can take while attaining similar accuracy. We chose to investigate the first and third hinge loss expressions from our collection (hereafter, "Exp. H1" \& "Exp. H3") and also the first and third MSE loss expressions ("Exp. M1" \& "Exp. M3").

What is immediately apparent about Exp. H1 (Fig. \ref{fig:exp_hinge}a) is that the combined eccentricity term $(z^2_{17} + z^2_{21})$ can never be negative, regardless of the values of $z_{17}$ and $z_{21}$. Therefore, it appears to function as a weighting term, modulating the balance between the two size components $z_{29}$ and $e^{e^{z_3}}$. The double-exponent $e^{e^{z_3}}$ rises so sharply for $z_3 > 1$ so as to impose a virtual veto on any classifications of metaphase (Fig. \ref{fig:exph1_detailed}). Meanwhile, for $z_3 < -1$, this term changes little, and metaphase classifications are allowed for $z_{29} > 0$, depending on the value of $(z^2_{17} + z^2_{21})$. In the range $-1< z_3 < 1$, the role of this eccentricity weighting term becomes more decisive.

\renewcommand{\expwidth}{0.43}
\begin{figure}
 \centering
    \begin{subfigure}[b]{\expwidth\textwidth}
      \includegraphics[width=\textwidth]{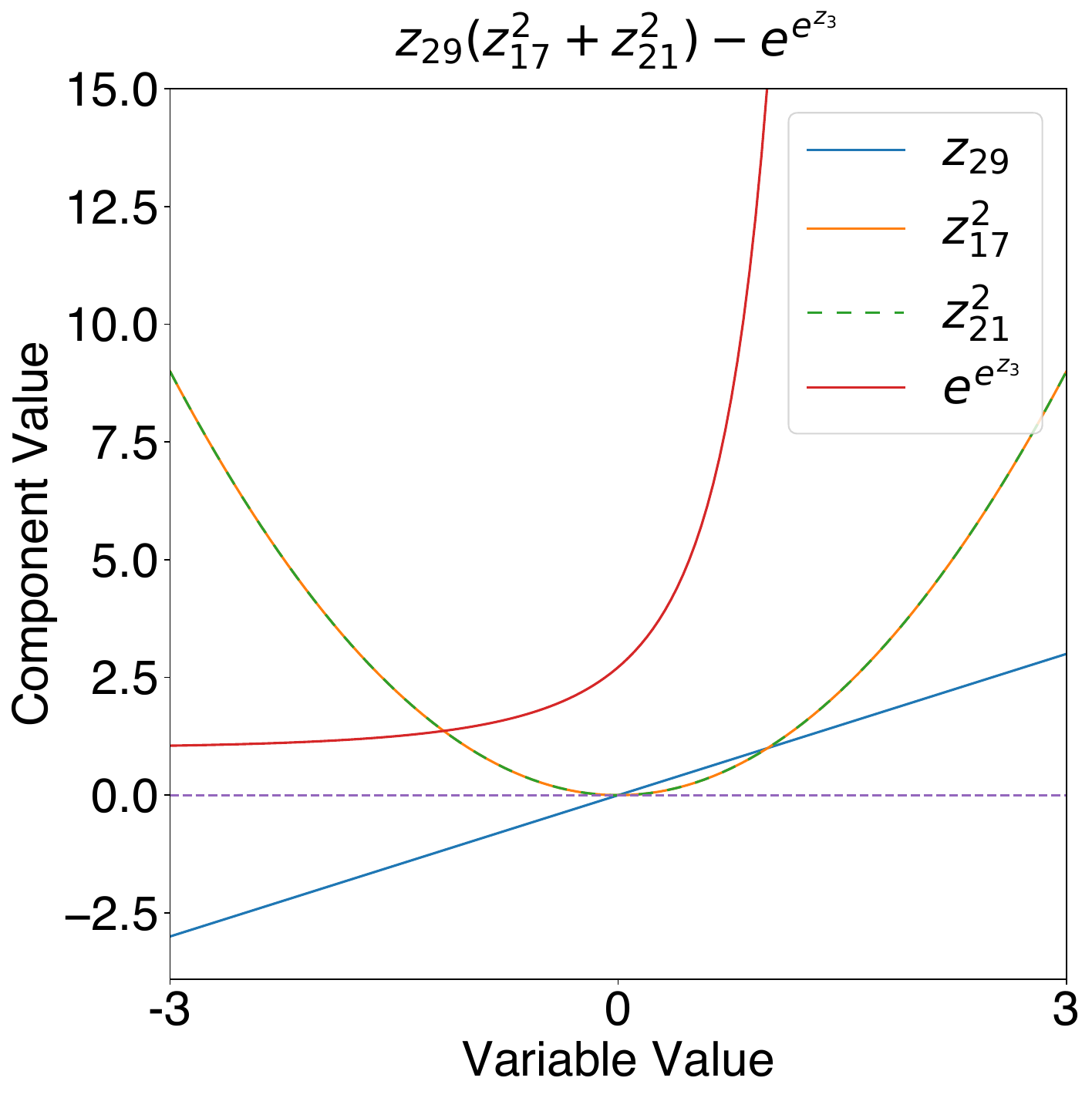}
      \caption{Exp. H1.}
    \end{subfigure}
    \begin{subfigure}[b]{\expwidth\textwidth}
      \includegraphics[width=\textwidth]{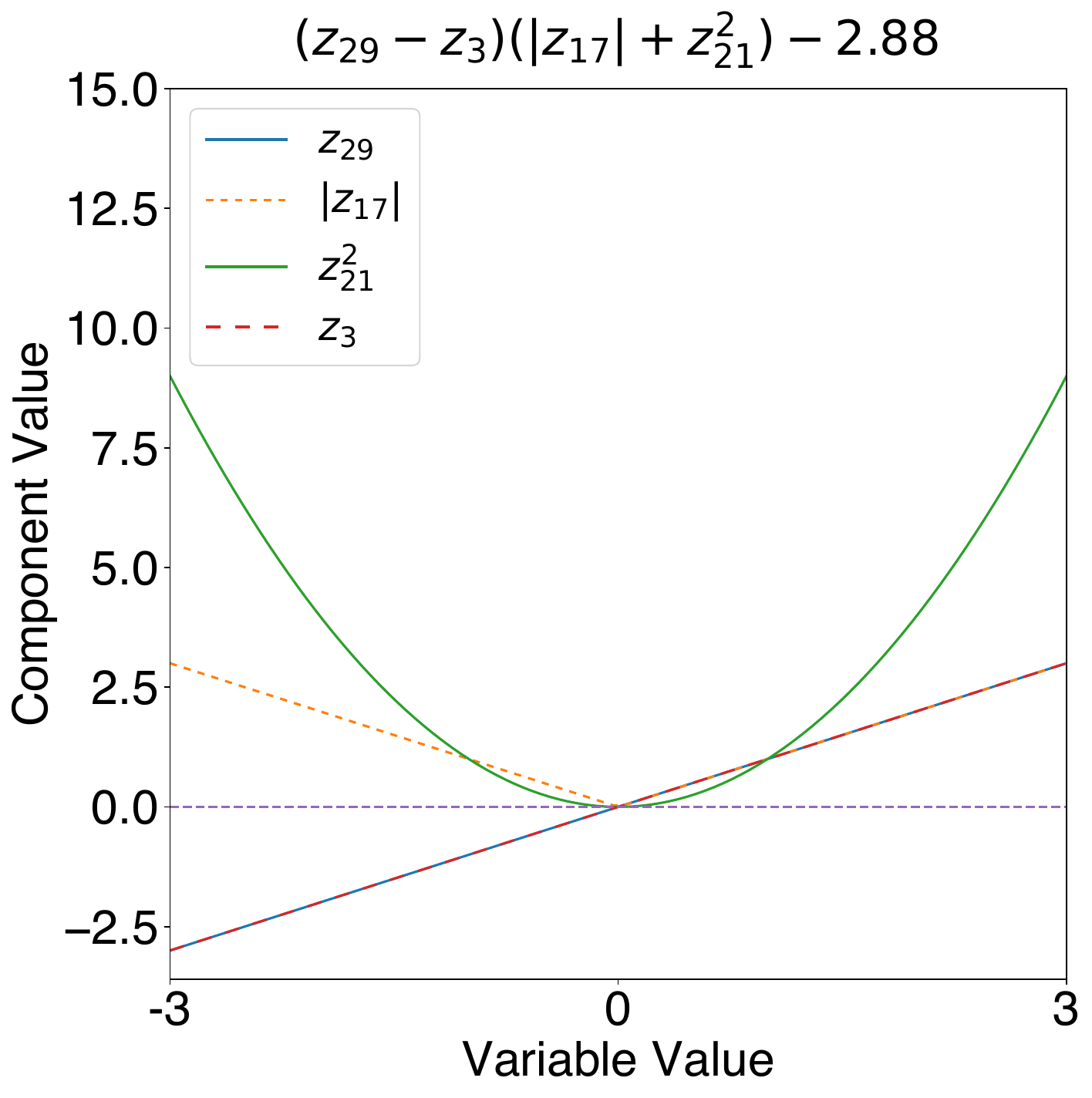}
      \caption{Exp. H3.}
    \end{subfigure}
    \caption{Hinge loss expressions: Component value mapped against the value of their relevant input variables. The purple dashed line represents a value of zero.}
    \label{fig:exp_hinge}
\end{figure}

\renewcommand{\figwidth}{1
}
\begin{figure}
 \centering
    \includegraphics[width=\figwidth\textwidth]{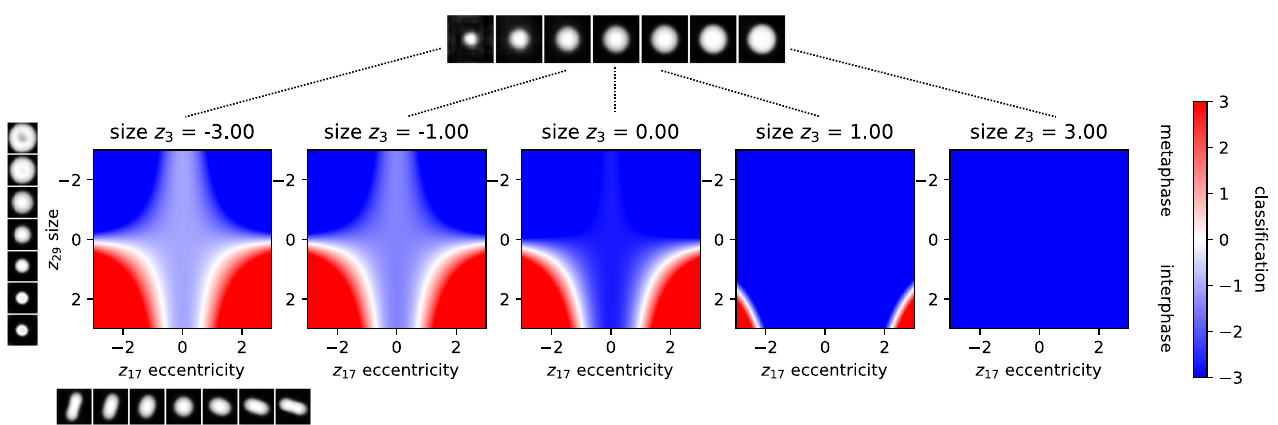}
       \caption{Further analysis of Exp. H1. Here, the decision boundary is plotted for varying values of $z_{3}$, $z_{17}$ \& $z_{29}$ within the typical range of values [-3, 3] and the value of $z_{21}$ is held constant at 0 for the purposes of clarity.}
    \label{fig:exph1_detailed}
\end{figure}

In Exp. H3, the eccentricity component reprises its role as a weighting factor, with the difference that $|z_{17}|$ is used in place of $z^2_{17}$. This time, it weights the combined term $(z_{29} - z_3)$ against the constant $2.88$. To obtain a classification of metaphase, it is necessary that $(z_{29} - z_3)(|z_{17}| + z^2_{21}) \geq 2.88$. If $z_3 \geq z_{29}$, signifying that the cell is at least a certain size, then metaphase classifications are excluded. On the other hand, if $z_3 < z_{29}$, then the classification depends on the value of $(|z_{17}| + z^2_{21})$. In other words, high cell size is sufficient to exclude metaphase classifications, but low cell size is insufficient to exclude any classification; rather, it is in this low cell size regime that the eccentricity component assumes importance.

\begin{figure}
 \centering
    \begin{subfigure}[b]{\expwidth\textwidth}
      \includegraphics[width=\textwidth]{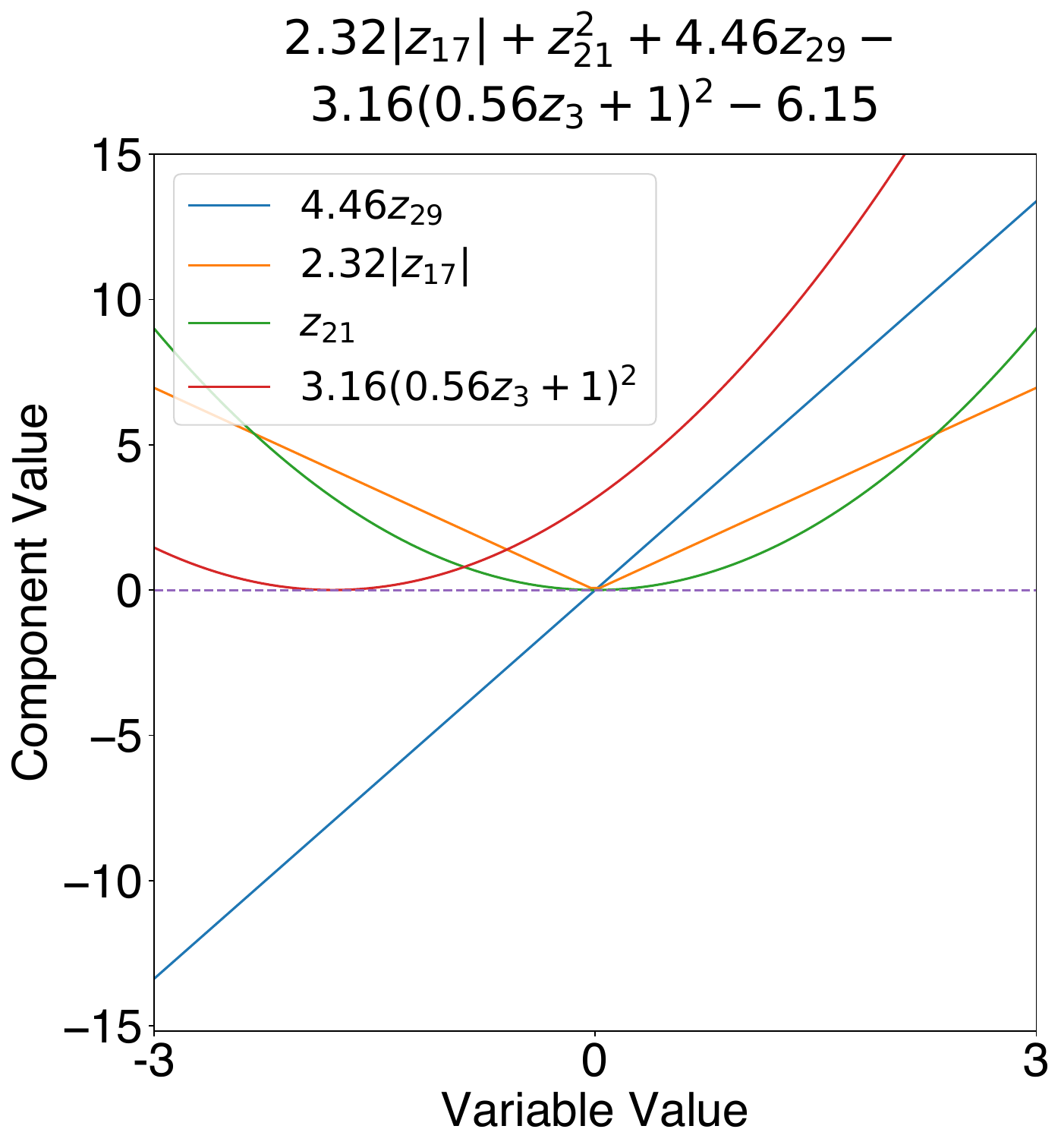}
      \caption{Exp. M1.}
    \end{subfigure}
    \begin{subfigure}[b]{\expwidth\textwidth}
      \includegraphics[width=\textwidth]{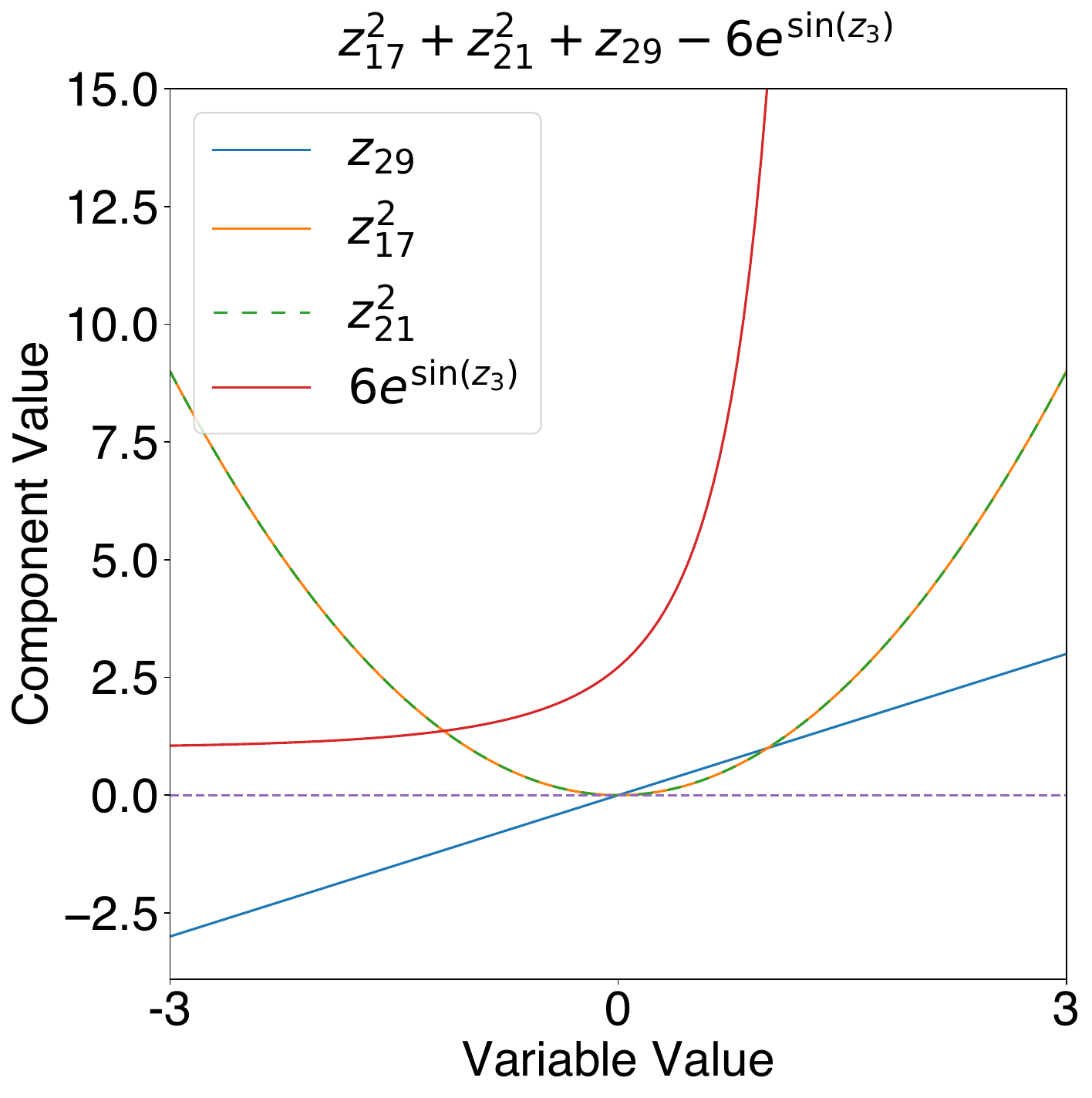}
      \caption{Exp. M3.}
    \end{subfigure}
    \caption{MSE loss expressions: Component value mapped against the value of their relevant input variables.}
    \label{fig:exp_mse}
\end{figure}

In Exp. M1 (Fig. \ref{fig:exp_mse}a), the eccentricity terms do not serve as a weighting factor, but contribute additively to the output score. Nevertheless, cell size still appears to be the sole factor capable of excluding classifications, due to the value range of the relevant expression components. Situations with high $z_3$ and low $z_{29}$ prescribe interphase classifications; even if the nucleus is highly eccentric, the relevant terms would be unable to push the output score over zero. On the other hand, situations in which $4.46z_{29} - 3.16(0.56z_3+1)^2 \geq 0$ prescribe metaphase classifications, given that the eccentricity terms can never be negative. Hence, unlike the Hinge loss models, the MSE models are capable of excluding \emph{both} interphase and metaphase on the basis of cell size alone.

Exp. M3 (Fig. \ref{fig:exp_mse}b) operates similarly. What is interesting is that the terms are all the same as those for Exp. H1, with a slight variation between $e^{e^{z_3}}$ and $6e^{\text{sin}(z_3)}$; however, in Exp. M3, the terms are all used additively, unlike in Exp. H1. Nevertheless, one basic rule remains the same: if $z_3$ is greater than $\sim1$, then classifications of metaphase are forbidden. The primary difference, again, is that cell size alone can exclude classifications of interphase as well, if $z_{29} - 6e^{\text{sin}(z_3)} \geq 0$.

In summary, there is a considerable diversity of form that Scheme 4 models can adopt to achieve high accuracy. These include minor differences, such as the variation between $|z_{17}|$ and $z^2_{17}$, as well as more substantial differences of mathematical form, such as that between multiplicative and additive structures, and finally, broad qualitative differences, such as the fact that Hinge loss models can exclude metaphase on the basis of zero eccentricity, while MSE models could not. They therefore give us multiple answers to the original goal question, "what distinguishes a cell in interphase from one in metaphase?". Nevertheless, all of these expressions basically capture the same general principle: interphase cells tend to be larger and rounder than metaphase cells.

\subsection{Adversarial attacks}

To further test the domain-appropriateness of our models, we applied adversarial attacks on each model to find counterfactual examples. The particular attack we used, the Fast Gradient Sign Method (\S\ref{section:methods_fgsm}), modifies the input by taking a jump within input space in the direction of the loss gradient, thereby worsening the performance of the model. The idea here is to find the minimal changes to the input required to flip the classification from interphase to metaphase, and vice versa. By inspecting the nature of these modifications, we can reach conclusions regarding the sensitivity of our models to various aspects of the input.

We conducted two types of adversarial attack, within image space and within latent space. Image-based attacks were implemented on Scheme 1-4 models while latent-based attacks were implemented on Scheme 2-4 models (given that Scheme 1 models do not possess an interpretable latent space).

\subsubsection{Image-based attacks}

\begin{figure}
 \centering
       \includegraphics[width=0.8\textwidth]{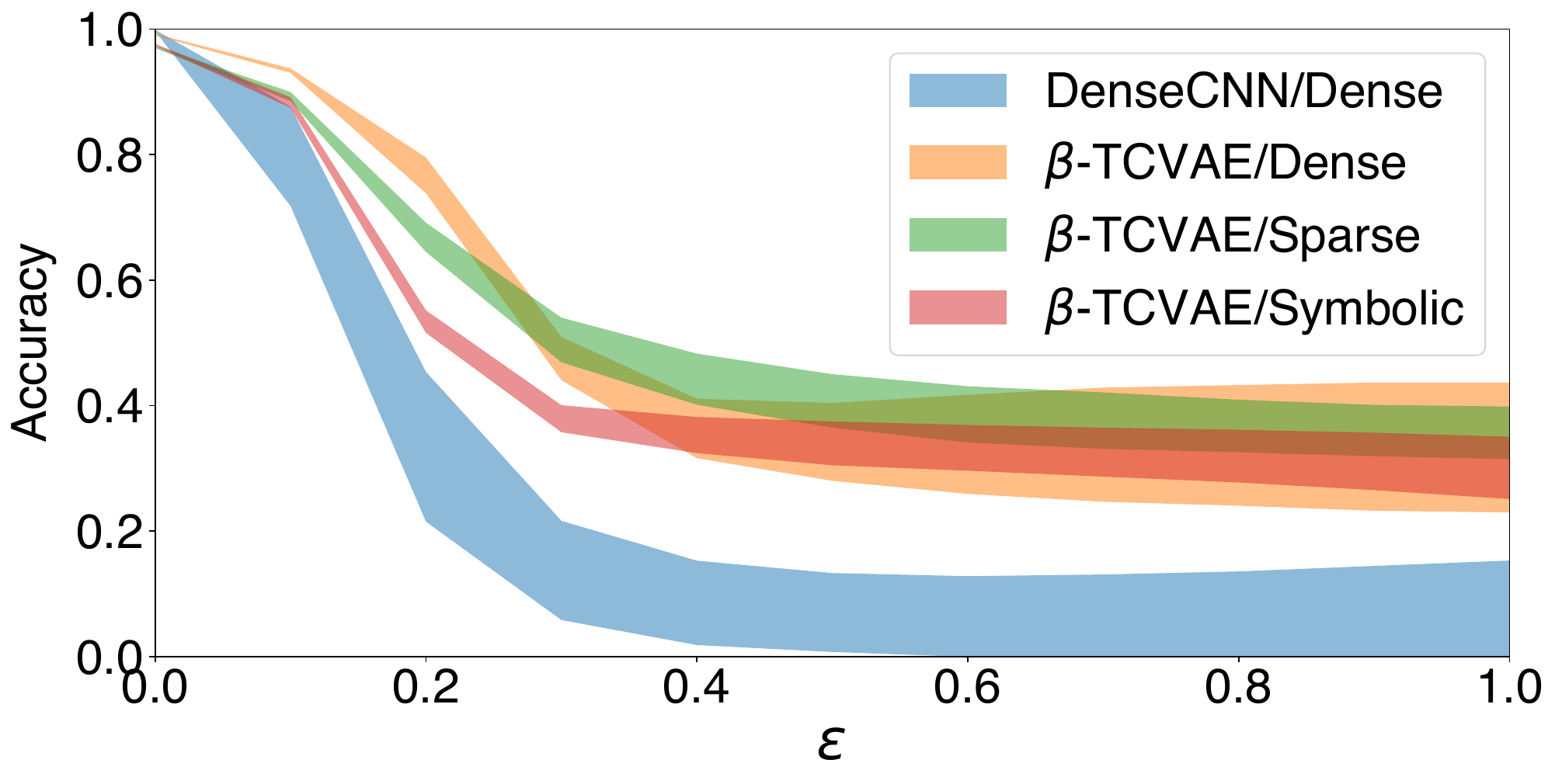}
       \caption{Adversarial attack curve for image-based perturbations. Plotted is testing accuracy against the perturbation magnitude $\epsilon$. The shaded region represents one standard deviation from the mean, with these being calculated over ten models for each scheme.}
    \label{fig:image_based_attack_curve}
\end{figure}

\begin{figure}
 \centering
       \includegraphics[width=0.75\textwidth]{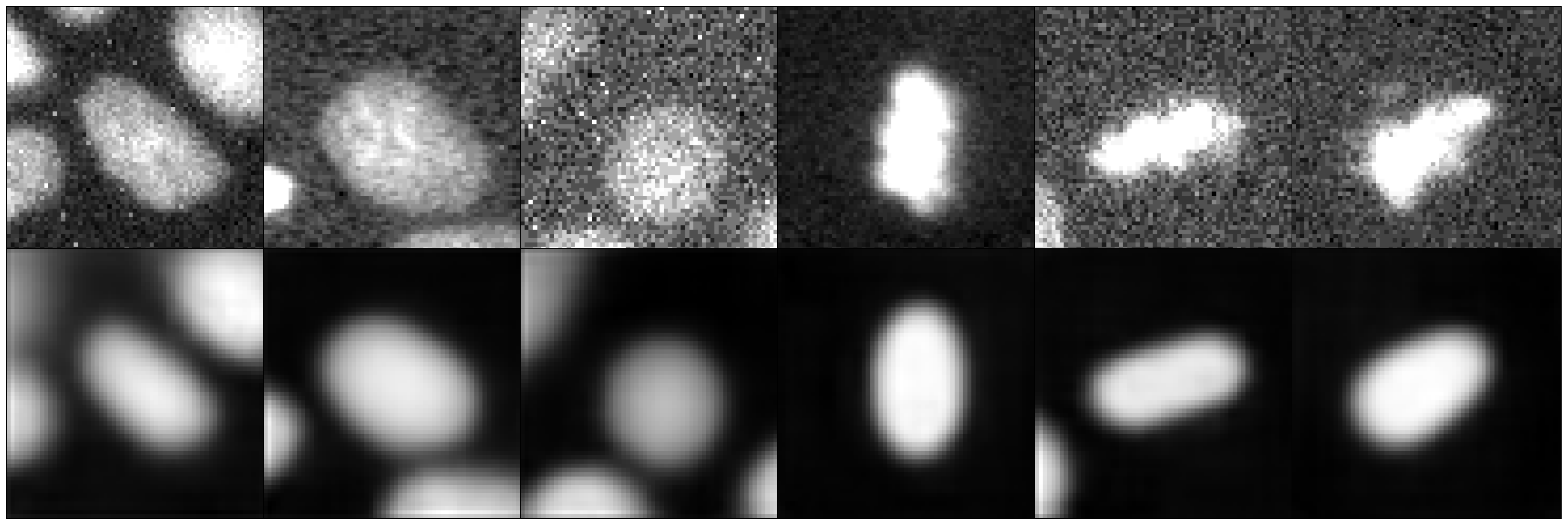}
       \caption{Un-perturbed image examples. \textbf{Top:} Image. \textbf{Bottom:} $\beta$-TCVAE reconstruction.}
    \label{fig:image_attack_original}
\end{figure}

Overall, Scheme 1 models were much less robust to image-based attacks than were Scheme 2-4 models (Fig. \ref{fig:image_based_attack_curve}). However, perhaps more interesting than the relative robustness to attack is the difference in the \emph{nature} of perturbation applied on the original images (Fig. \ref{fig:image_attack_original}) that could flip the classification. As displayed in Fig. \ref{fig:image_based_attacks}, successful attacks on Scheme 1 models appear more widely dispersed around the image, with only a slight bias on the central cell, and they do not lead to any differences in morphology significant enough to be captured by the $\beta$-TCVAE encoding. On the other hand, successful attacks on Scheme 2-4 models alter the images in ways consistent with both domain knowledge, and the results shown in \S\ref{section:decision_boundary}. In other words, Interphase $\rightarrow$ Metaphase perturbations will form cells that are smaller and/or more elongated, while Metaphase $\rightarrow$ Interphase perturbations form cells that are more rounded and diffuse. The latter transition typically occurs by depressing pixel values associated with the metaphase chromatin signature in order to simulate a more diffuse signature. On the other hand, Interphase $\rightarrow$ Metaphase perturbations typically "carve out" a pill-shaped region at the center of the cell by depressing all other regions where chromatin is present.

\renewcommand{\figwidth}{0.49}
\begin{figure}
 \centering
    \begin{subfigure}[b]{\figwidth\textwidth}
      \includegraphics[width=\textwidth]{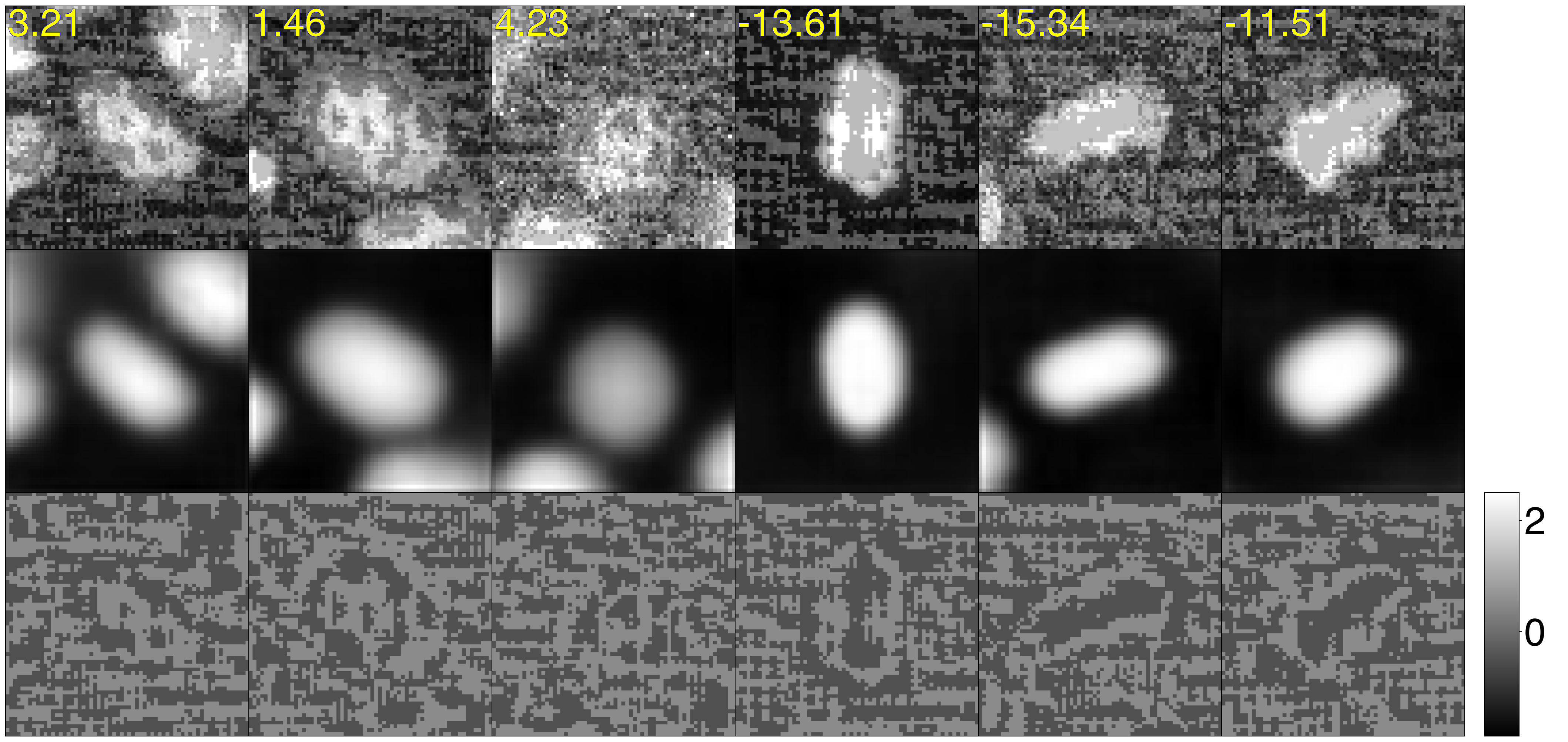}
      \caption{Attacks by Scheme 1 models.}
    \end{subfigure}
    \begin{subfigure}[b]{\figwidth\textwidth}
      \includegraphics[width=\textwidth]{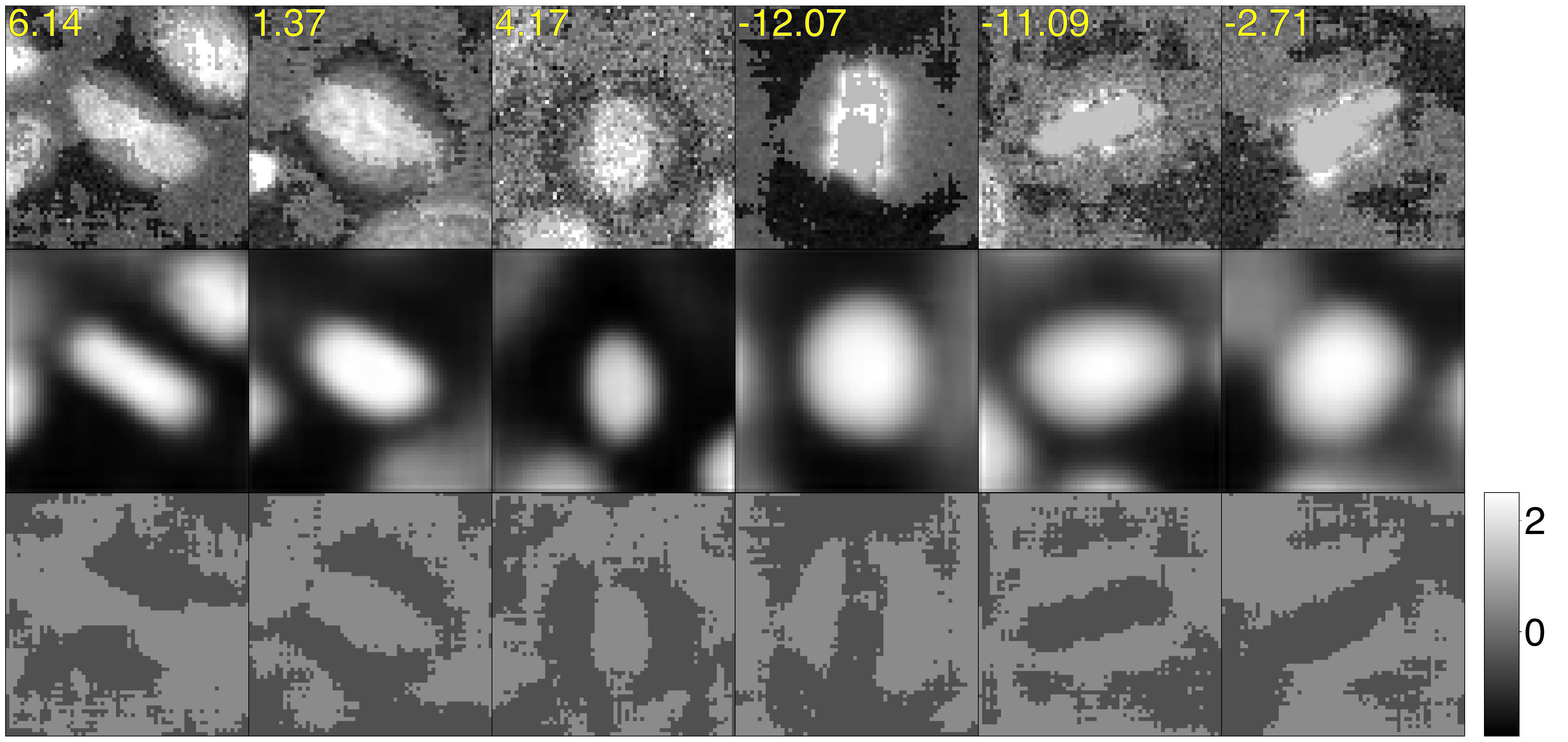}
      \caption{Attacks by Scheme 2 models.}
    \end{subfigure}
    \begin{subfigure}[b]{\figwidth\textwidth}
      \includegraphics[width=\textwidth]{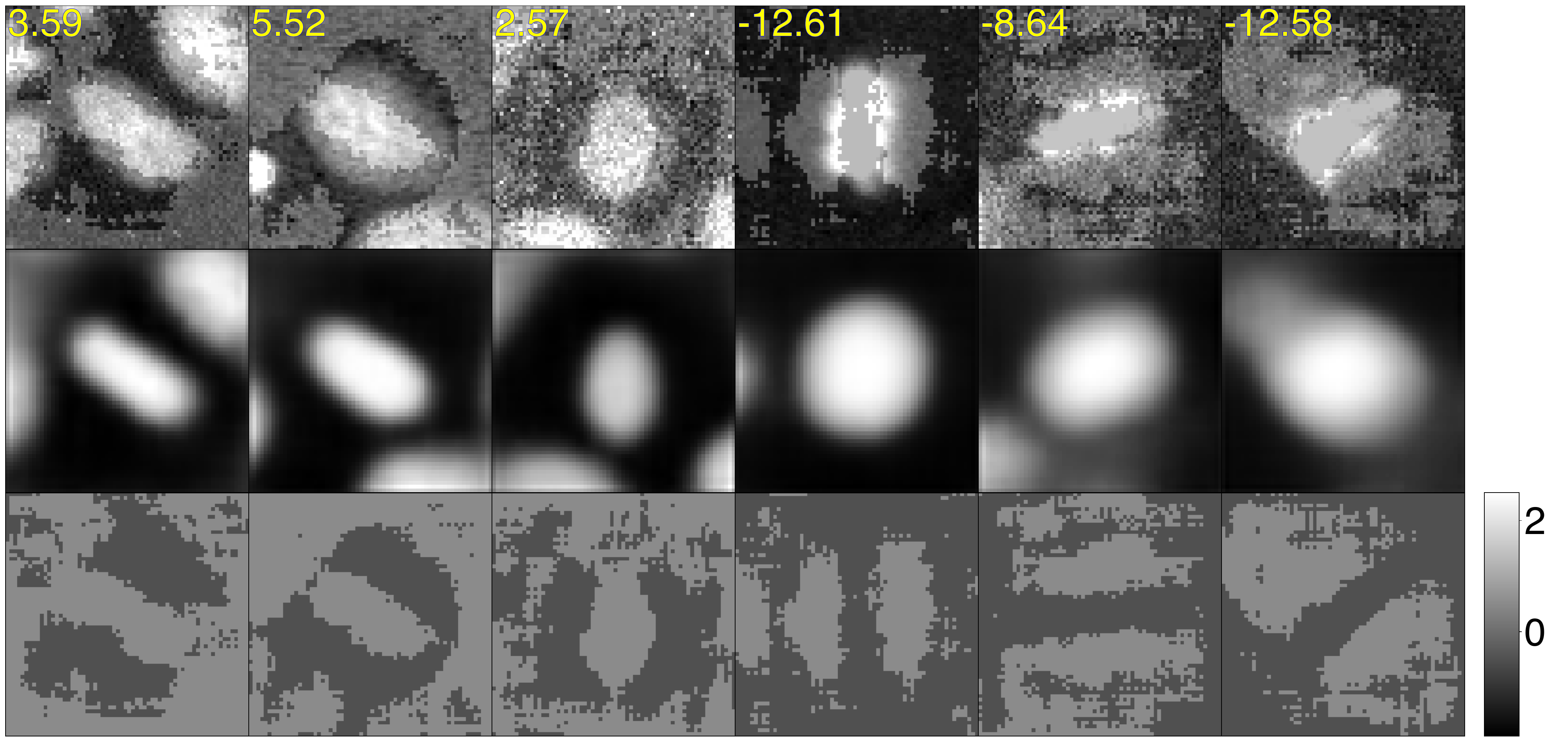}
      \caption{Attacks by Scheme 3 models.}
    \end{subfigure}
    \begin{subfigure}[b]{\figwidth\textwidth}
      \includegraphics[width=\textwidth]{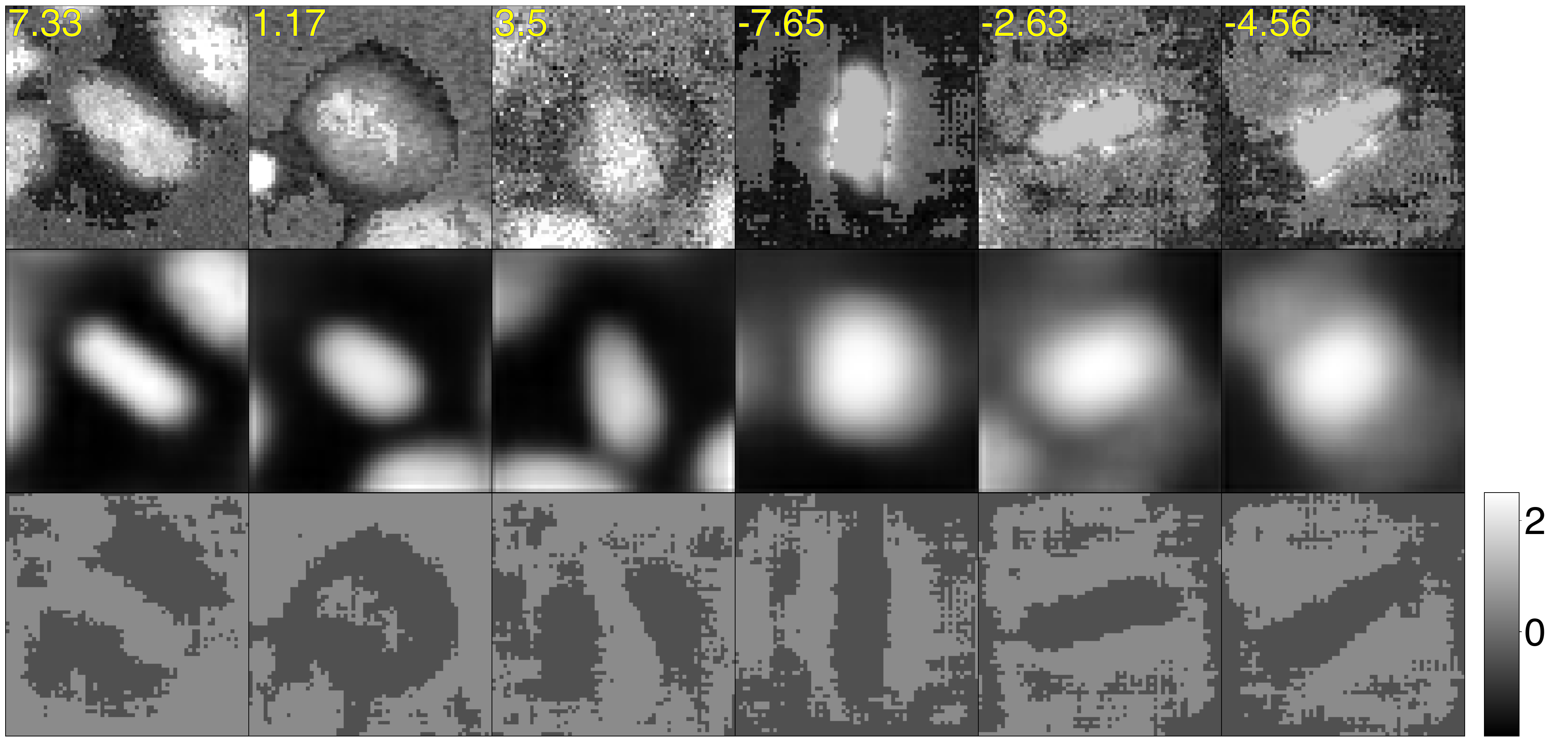}
      \caption{Attacks by Scheme 4 models.}
    \end{subfigure}
    \caption{Image-based attacks at $\epsilon = 0.5$. Exp. H1 was used for \textbf{(d)}. Shown are three interphase images (left) and three metaphase images (right), with attacks by various models. In each sub-figure are shown the perturbed image (top), associated $\beta$-TCVAE reconstruction (middle) and the pixel-wise difference between the perturbed and un-perturbed images (bottom). Numbers shown are the post-attack output scores.}
    \label{fig:image_based_attacks}
\end{figure}

\subsubsection{Latent-based attacks}

\begin{figure}
 \centering
       \includegraphics[width=0.8\textwidth]{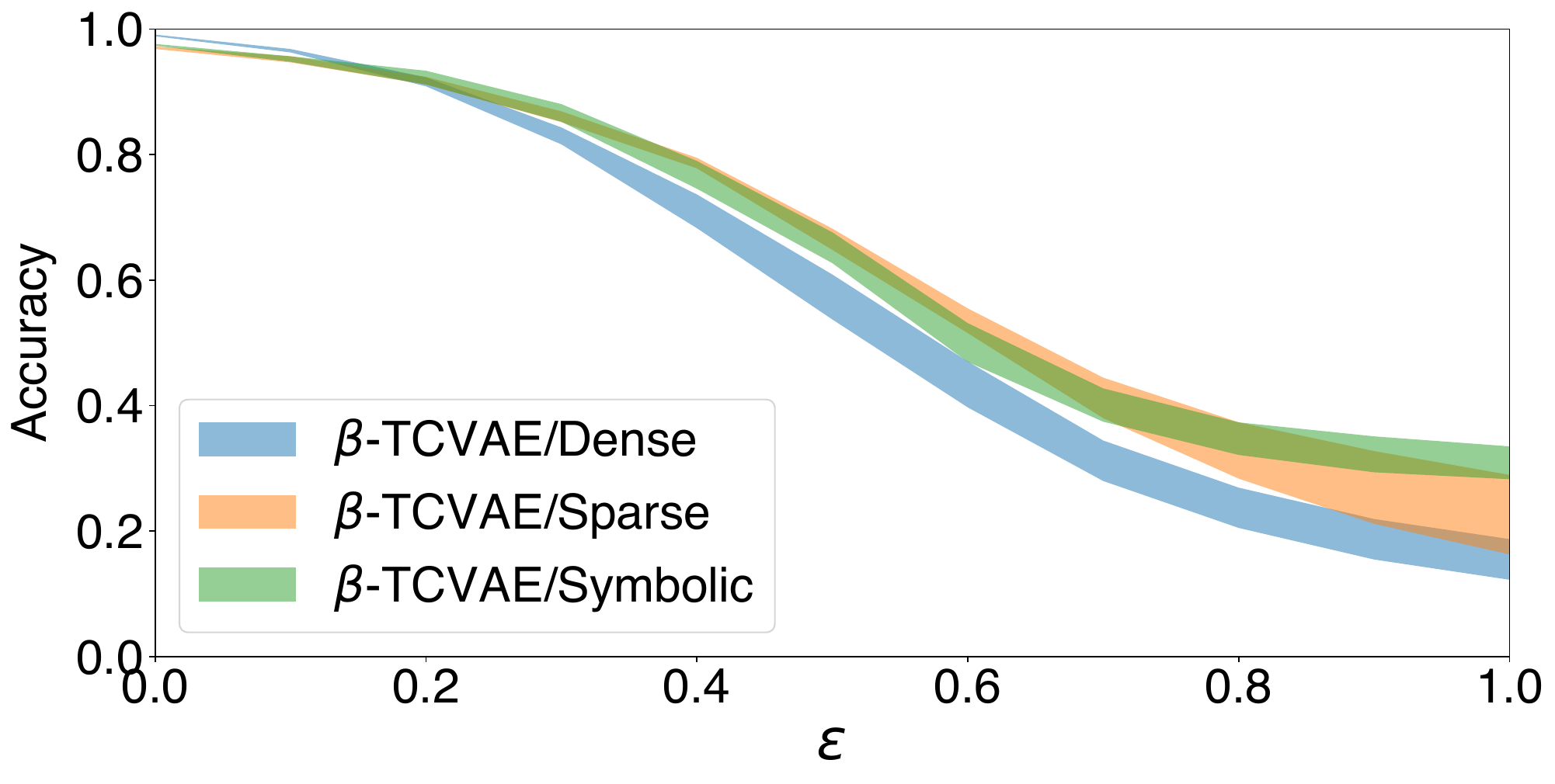}
       \caption{Adversarial attack curve for latent-based perturbations.}
    \label{fig:latent_based_attack_curve}
\end{figure}

\begin{figure}
 \centering
    \begin{subfigure}[b]{\figwidth\textwidth}
      \includegraphics[width=\textwidth]{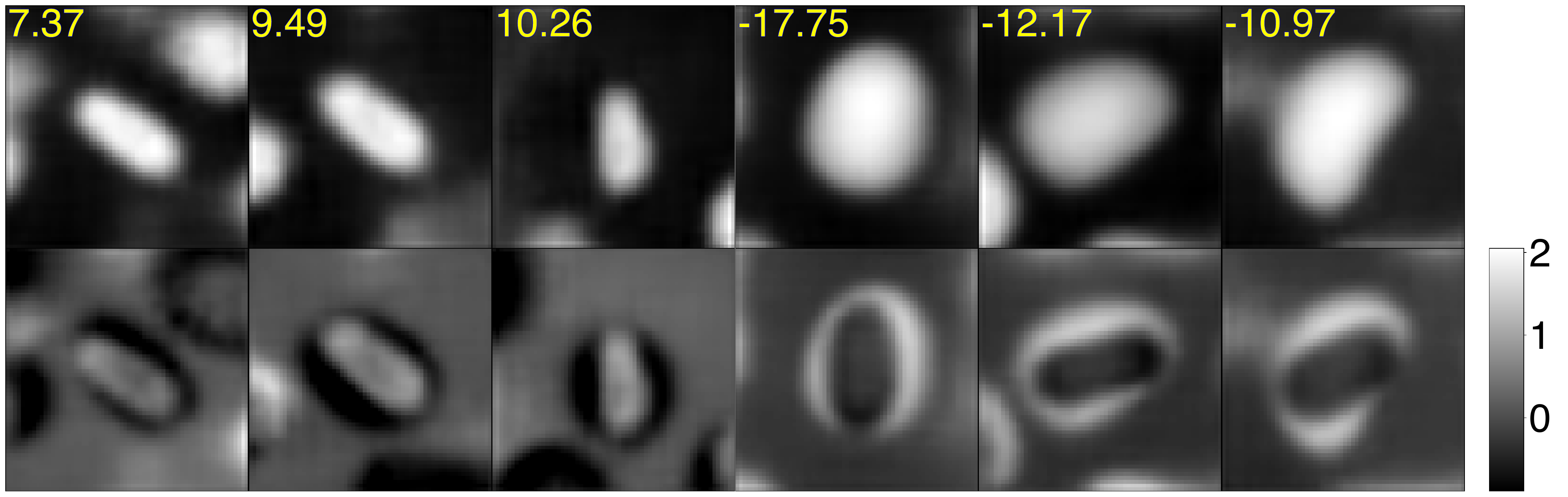}
      \caption{Attacks by Scheme 2 models.}
    \end{subfigure}
    \begin{subfigure}[b]{\figwidth\textwidth}
      \includegraphics[width=\textwidth]{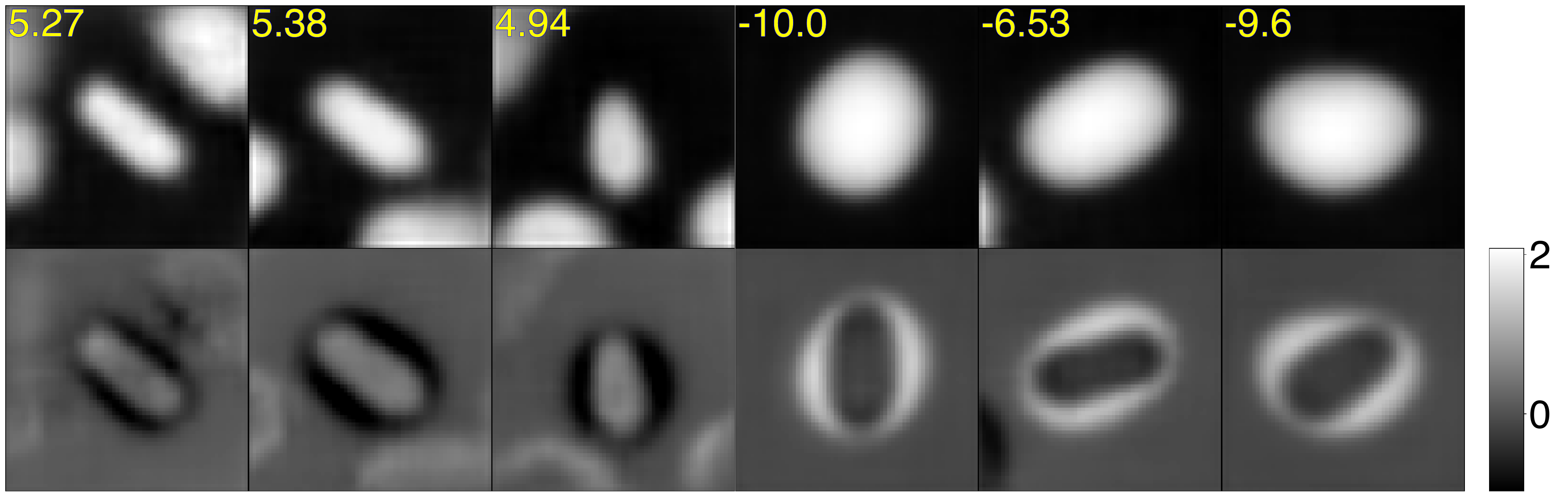}
      \caption{Attacks by Scheme 3 models.}
    \end{subfigure}
    \begin{subfigure}[b]{\figwidth\textwidth}
      \includegraphics[width=\textwidth]{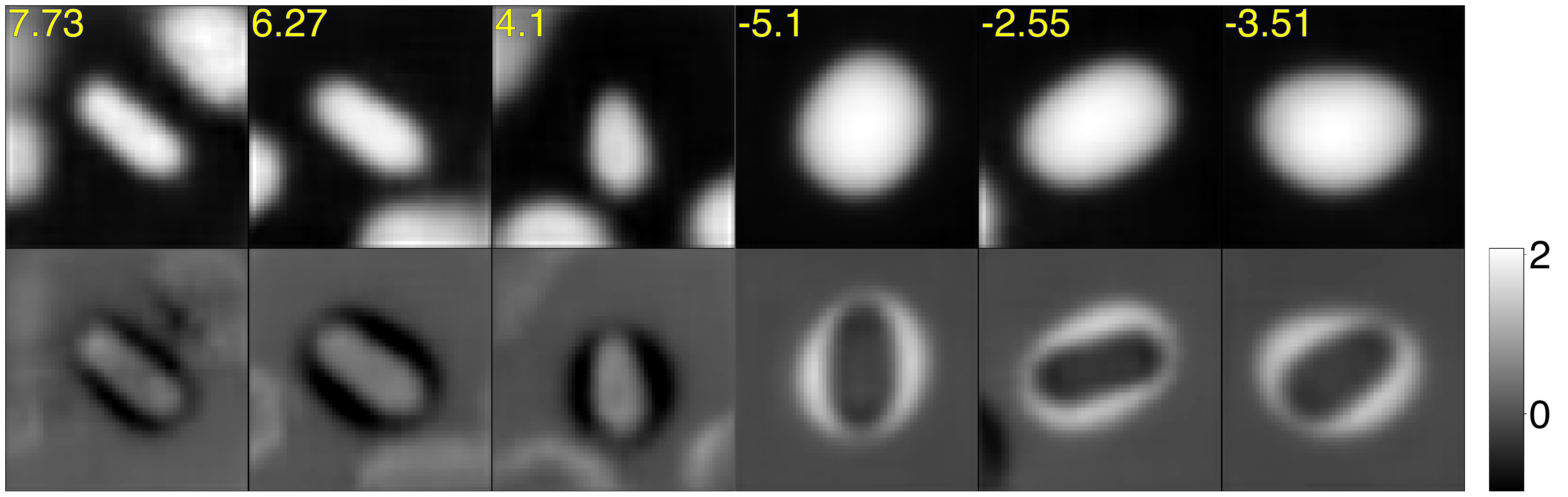}
      \caption{Attacks by Scheme 4 models.}
    \end{subfigure}
    \caption{Latent-based attacks at $\epsilon = 1.0$. Exp. H1 was used for \textbf{(c)}.}
    \label{fig:latent_based_attacks}
\end{figure}

Scheme 2-4 models were roughly equal in their level of robustness to latent-based attacks (Fig \ref{fig:latent_based_attack_curve}). Furthermore, attacks on these models produce sensible transformations of nuclear morphology; in interphase images, the region around the nucleus is depressed to form a pill-shape in the center, and in metaphase images, the reverse occurs, forming a rounded shape (Fig. \ref{fig:latent_based_attacks}).

However, as an extra test, we sought to assess the robustness of Scheme 2 models to perturbations of latent features that \emph{do not} encode the central cell morphology. This consists of all latent variables all except $z_3$, $z_{17}$, $z_{21}$ \& $z_{29}$, (hereafter dubbed "neighborhood features" despite the fact that two of them encode the central cell position, shown in Fig. \ref{fig:latent_variables_neighborhood}). In other words, we sought to assess whether Scheme 2 models adhere to the domain constraint that the classification of cell state in the current context should depend only on nuclear morphology.

\begin{figure}
 \centering
       \includegraphics[width=0.8\textwidth]{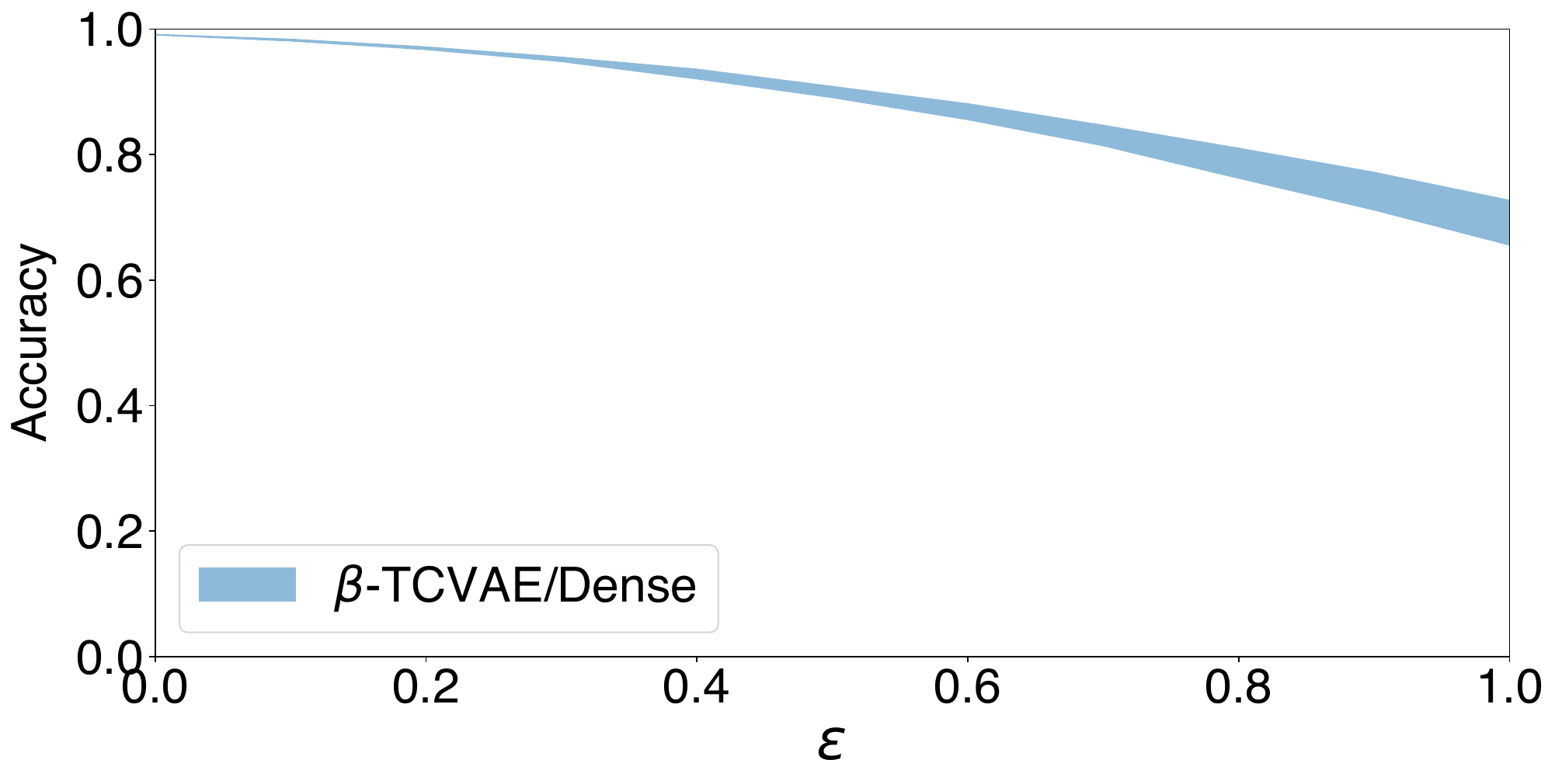}
       \caption{Adversarial attack curve for latent-based perturbations that leave unchanged those latent variables relevant to central cell morphology ($z_3$, $z_{17}$, $z_{21}$ \& $z_{29}$).}
    \label{fig:attack_curve_excluded}
\end{figure}

In general, the sensitivity of Scheme 2 performance to variation in neighborhood features is non-zero (Fig. \ref{fig:attack_curve_excluded}). The effects are not as severe as transformations that include the central cell (Fig. \ref{fig:latent_based_attack_curve}), however, attacks at $\epsilon = 1.0$ were still able to reduce accuracy to $69 \pm 3\%$ from $99.1 \pm 0.1\%$. When these neighborhood features are varied, the prominent effects in image space are changes in the area external to the central cell. There are some minor variations in the morphology of the central cell itself, but these do not significantly alter this morphology.

\begin{figure}
 \centering
       \includegraphics[width=0.8\textwidth]{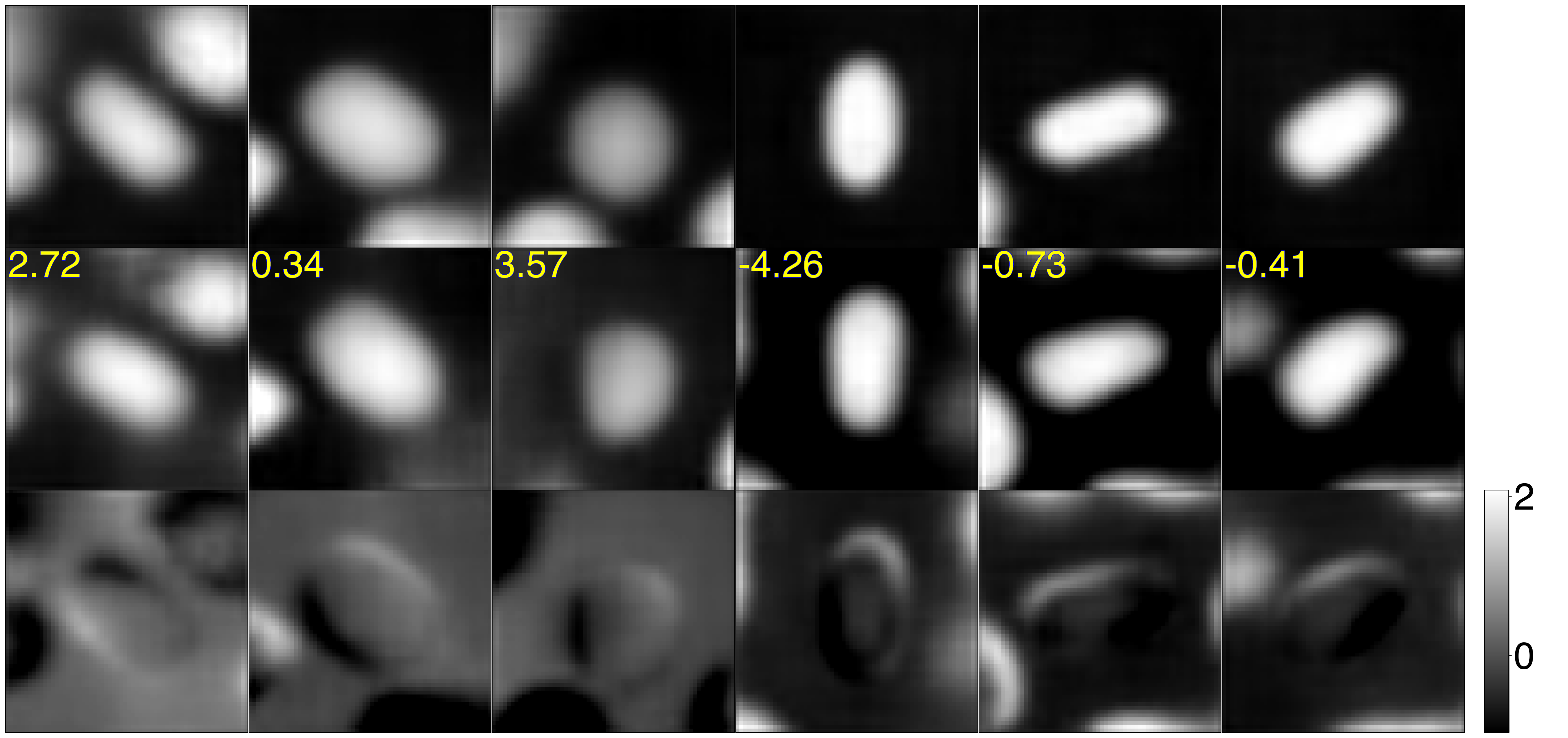}
       \caption{Latent-based attacks at $\epsilon = 1.0$, excluding those latent variables relevant to central cell morphology ($z_3$, $z_{17}$, $z_{21}$ \& $z_{29}$). \textbf{Top}: $I_{orig}$, $\beta$-TCVAE decoding of the original latent vector. \textbf{Middle}: $I_{perturbed}$, decoding of the perturbed latent vector. \textbf{Bottom}: $I_{orig} - I_{perturbed}$. The number represents the post-perturbation output score.}
    \label{fig:attack_examples_excluded}
\end{figure}

Interestingly, a consistent effect in Interphase $\rightarrow$ Metaphase perturbations appears to be a reduction of neighborhood density, primarily through the contraction or outright removal of neighboring cells (Fig. \ref{fig:attack_examples_excluded}). The reverse is true in Metaphase $\rightarrow$ Interphase perturbations; new cells are placed in the vicinity of the central cell. We can hypothesize from this finding that Scheme 2 models use information about neighborhood density to make their classifications; sparse neighborhoods promote metaphase classifications, and dense neighborhoods promote interphase classifications. We suggest that this is due to the relationship between cell state and cell size throughout the dataset. MDCK cells follow an "adder" model of size control, meaning that they divide after they have added a certain volume to their initial (birth) size \cite{cadart_size_2018}. Therefore, cells that undergo mitosis, and therefore transition to metaphase, tend to be large and therefore, their neighborhoods will be sparse. It is therefore likely that Scheme 2 models use this neighborhood density information \emph{in addition to} central cell morphology information to gain the extra $2$-$3\%$ of accuracy over Scheme 3 \& 4 models. The bulk of their classification capability is derived via attention to central cell morphology (as indicated by the relatively gentle decrease in Fig. \ref{fig:attack_curve_excluded}); however, there is some dependence on the neighborhood. 

While this approach may increase model performance, it represents a clear "shortcut" strategy in the sense described by \citet{geirhos_shortcut_2020}. The model generalizes in a way different to the way in which human scientists would, and in a way that contradicts domain knowledge. As a result, the model could classify completely differently two images with the same central cell morphology, but different neighborhoods - a failure that clearly constitutes domain-inappropriate behavior.

\section{Conclusions}

From our results we draw six broad observations:

\begin{enumerate}
    \item \textbf{Semantic latent spaces are feasible to develop}: Using our $\beta$-TCVAE, we were able to produce an interpretable latent space from raw experimental data, and we were able to attach semantic labels to the latent variables with reasonable confidence.
    \item \textbf{Scheme 1-4 models all achieve very similar performance}: Whilst Scheme 1 models remain in the lead in terms of testing accuracy, the gap between Scheme 1 and Scheme 3 (the lowest performing) is only about 2-3\% (Table. \ref{table:test_performance}).
    \item \textbf{Scheme 3 \& 4 classification heads are far more parsimonious}: Scheme 3 models are able, on average, to capture 98\% of the performance of Scheme 2 models with only 2.2\% of the parameter count and 2.1\% of the expression size (Table. \ref{table:test_performance}). The case with Scheme 4 models is even more striking; they are able to capture 98\% of Scheme 2 performance with only 0.2\% of the expression size.
    \item \textbf{Scheme 3 classification heads are able to identify the relevant input features}: They are able to specify both the appropriate number and selection of input features, as deemed relevant by prior domain knowledge.
    \item \textbf{Scheme 4 symbolic expressions are interpretable and analyzable}: It is possible to draw broad conclusions about the behavior of the model from inspection of the classification expression. In particular, we discovered that cell size was the key determinant of classification in all of these models, and that eccentricity played slightly different roles in Hinge loss and MSE loss models. In the latter, eccentricity alone can decide the classification; a feature not present in the former.
    \item \textbf{Scheme 3 sparse networks are interpretable but the process is more difficult}: Interpretation and analysis of sparse networks is considerably more taxing than analysis of symbolic expressions, especially in cases where multiple input variables influence one hidden layer neuron to form "compound" internal features whose meaning and function can be difficult to assess.
    \item \textbf{Scheme 3 \& 4 adversarial attacks consist of domain-appropriate perturbations}: Scheme 1 models are sensitive to small, unstructured image perturbations. Meanwhile, image-based attacks on Scheme 2-4 models consist of structured perturbations that affect central cell morphology in domain-appropriate ways. Scheme 2 models are, however, sensitive to perturbations of neighborhood features, while, due to the nature of their restricted input space, Scheme 3 \& 4 models are not. This strategy is a clear example of "shortcut learning", where the model generalizes in a way inappropriate to the domain.
\end{enumerate}

\begin{figure}
 \centering
       \includegraphics[width=0.8\textwidth]{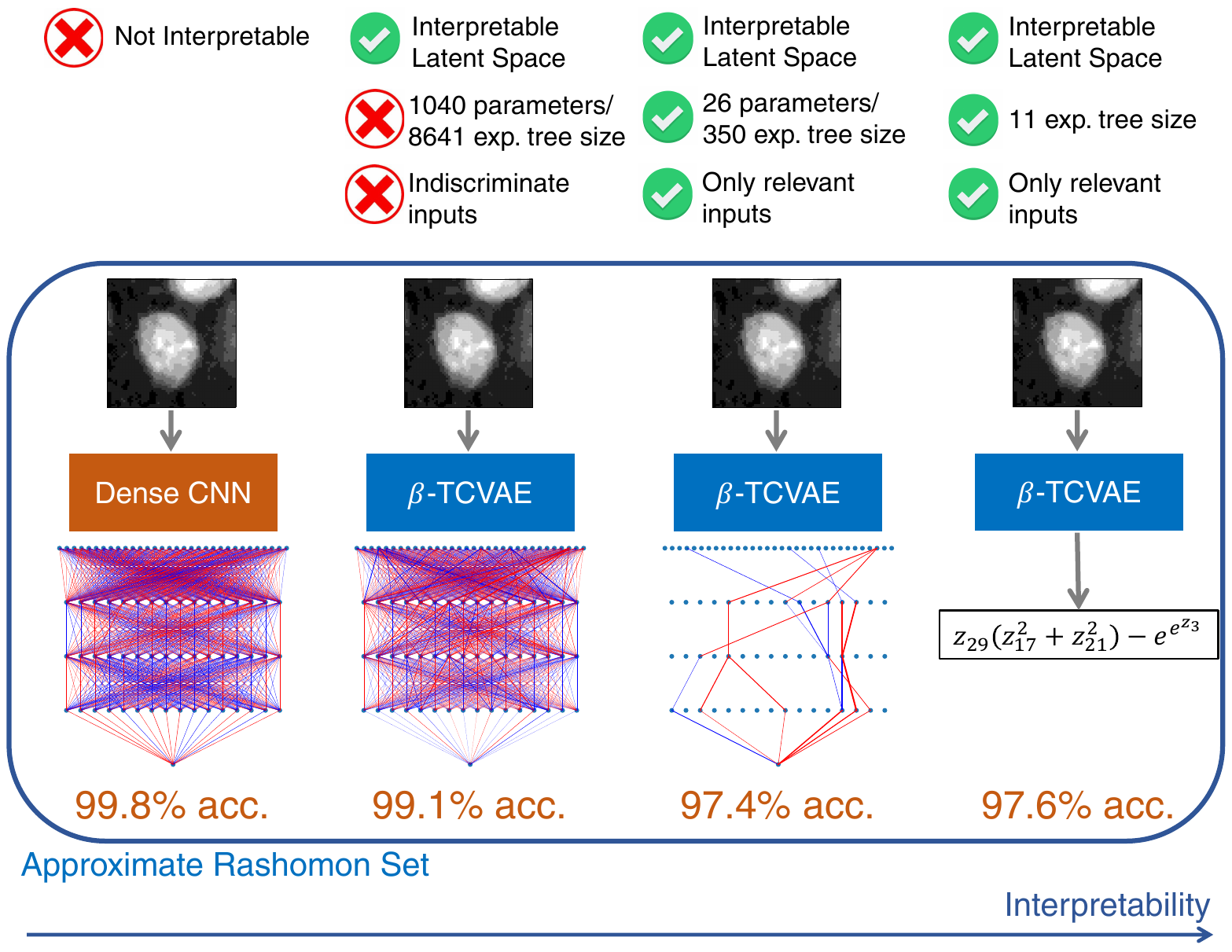}
       \caption{Approximate Rashomon set featuring the top-performing models from Schemes 1, 2, 3 \& 4 (in order, from left to right). For Schemes 1-3, performance was measured by accuracy. For Scheme 4, the chosen expression was the one judged to strike the optimal balance between accuracy and simplicity.}
    \label{fig:model_comparison}
\end{figure}

From these observations we conclude that it is indeed possible to produce models at the intersection of the performant, interpretable and domain-appropriate subspaces (Fig. \ref{fig:model_concept}b). While Scheme 2-4 models experienced some drop in performance with increasing interpretability, we observed that this drop is minuscule compared to the significant gains in model analyzability and domain-appropriateness. In particular, our Scheme 3 models were able to take a dataset of raw images and identify the relevant subspace of inputs (pertaining to central cell morphology), and Scheme 4 models were able to find functions on these inputs that are consistent with our prior knowledge of the problem. Furthermore, the interpretability of these models allowed us to identify cell size, rather than eccentricity, as the predominant input feature. Hence, we were able to gain insight into the operation of the model. In summary, we were able to find an approximate Rashomon set, with an accuracy range of $<3\%$, that included un-interpretable and domain-inappropriate models as well as interpretable and domain-appropriate ones (Fig. \ref{fig:model_comparison}).

\section{Discussion}

In \S\ref{section:dangers_deep_learning}, we introduced the notion of "usefulness" in scientific models and decomposed this concept into three criteria: performance, interpretability and domain-appropriateness. Deep neural networks typically excel at performance, but are generally un-interpretable, and are prone to shortcut learning, which undermines domain-appropriateness. We proposed three strategies to rectify this issue: disentangled representation learning to transform raw data into semantic data, sparse network training to reduce the complexity of DNN models, and symbolic regression to supplement our DNN-derived representations with an interpretable free-form learning technique. Our end-goal was the creation of an ideal "discovery system", that could learn regularities from data that human scientists would consider as valid scientific laws.

Broadly, these techniques were successful, with Scheme 3-4 models showing greater interpretability and domain-appropriateness compared to Scheme 1 \& 2 models, while suffering only minuscule reductions in performance. We have therefore demonstrated the potential of our techniques in our test system. However, there remain significant challenges should we attempt to generalize them to other, more complex domains.

Firstly, extracting disentangled semantic representations from raw data remains a considerable challenge. Our test problem featured a dataset in which the images show a great deal of consistency and symmetry; there is always a cell (blob) in the middle, surrounded by some cells (other blobs) in the vicinity. Generalizing our results to datasets with far more numerous factors of variation may require further development in the field of representation learning.

A second, and related, issue, is that in the absence of prior knowledge, it may be difficult to assess whether our latent representation has learnt the \emph{right} features of the input data. In this study, we inspected the reconstructions of our $\beta$-TCVAE to verify that it had developed a suitable representation of nuclear shape. With our prior knowledge, we judged this to be sufficient for interphase/metaphase classification. However, it might be \emph{insufficient} for interphase/\emph{prometaphase} classification, given that the difference between the two is primarily textural, and our $\beta$-TCVAE did not develop any representations of nuclear texture. To address this issue, we may wish to implement strategies such as the joint optimization of the latent space between task-specific and VAE-related objectives \cite{van_der_valk_joint_2023}.

Thirdly, in the absence of prior knowledge, it may be non-trivial to determine whether our models are "unnecessarily complex". We observed several instances of such behaviors in both Scheme 3 \& 4 models using assumptions of monotonicity derived from prior knowledge. However, for unfamiliar systems, it may be difficult to determine whether complex behavior of this kind is simply an accident of model architecture, or is a "legitimate" pattern drawn from the data.

Finally, strategies may have to be designed to handle tasks other than binary classification or regression. For example, symbolic regression is readily applicable to binary classification - the regression model would then be trained to produce a score whose value indicates the class - however, this approach might be invalid for multi-class classification. One potential fix would be to have one output (logit) neuron for each class. Then, we could train a symbolic regression model on the output of each neuron and use inequalities to discover regions of input space that correspond to each class. Needless to say, this approach is far more complex than the binary classification case that we have presented here.

Overall, challenges are to be expected when generalizing our strategies to systems of greater complexity, however, these are mainly technical and conceptual. There are no obstacles in \emph{principle} to the application of our methods to other systems.

As a final note, it is interesting to observe the vast diversity of models that could all achieve very similar performance. Even within Scheme 3 alone, there exists a large variation of topologies (Appendix \ref{appendix:sparse_model_topologies}) and within Scheme 4 models there exists a large variation of expression forms (Tables \ref{table:symbolic_accuracy_hinge} \& \ref{table:symbolic_accuracy_mse}) that all fall within the Rashomon set. These models are capable of approximating each other in their outputs, while operating through a variety of internal mechanisms. What's clear is that training models with high complexity is often not necessary nor even sufficient to achieve desirable characteristics such as performance and robustness \cite{wang_larger_2023}.

These observations also serve to reinforce George Box's claim that while no model can "truly" describe the world in its entirety, some are nevertheless "useful" \cite{box_robustness_1979, box_science_1976}. It would be difficult to pick a "true" model out of the ones we have assessed, however - following the criteria of interpretability and domain-appropriateness - we \emph{can} say which models are more "useful". If we were to answer the original goal question, "what distinguishes a cell in interphase from one in metaphase?" we would probably consult our Scheme 4 models rather than our Scheme 1 models, even though the latter achieve marginally greater performance.

In summary, this work has demonstrated the possibility of free-form scientific induction under constraint, using deep neural networks and associated interpretability techniques. There is much room for further exploration, and it is exciting to ponder the extent to which we can imbue machines with whatever it is that underlies our capacity for science.

\section{Acknowledgements}

This work was funded by The Alan Turing Institute. This work was also supported in part by Baskerville: a national accelerated compute resource under the EPSRC Grant EP/T022221/1, \& The Alan Turing Institute under EPSRC grant EP/N510129/1.

\section{Author contributions}

CJS and ARL designed and conceived the research. CJS performed the computational work. CJS and ARL evaluated the results and wrote the paper.

\bibliography{references}

\appendix

\section{Datasets}
Image datasets were from our previously published research \cite{bove_local_2017,ulicna_automated_2021,soelistyo_learning_2022}.

\section{Extended methods}
\subsection{Total Correlation VAE}
\label{section:extended_methods_tcvae}

The VAE is a latent variable model that consists of an encoder network parameterized by $\theta$ and a decoder network parameterized by $\phi$. The encoder network transforms an input $\mathbf{x}$ into its latent space representation $\mathbf{z}$ via a two-step process: first, the computation of $\mathbf{\mu}$ and $\mathbf{\sigma}$ vectors, then the sampling of $\mathbf{z}$ from a multivariate Gaussian distribution: $q_{\phi}(\mathbf{z}\vert\mathbf{x}) = \mathcal{N}(\mathbf{\mu},\mathbf{\sigma})$. Then, the decoder forms a reconstruction based on the latent space representation $\mathbf{z}$.

In the original formulation, the model is trained using the following loss function \cite{kingma_auto-encoding_2022}.
\begin{equation}
    \mathcal{L}(\theta, \phi; \mathbf{x}, \mathbf{z}) =
    \text{D}_{\text{KL}}( q_\phi(\mathbf{z} \vert \mathbf{x}) \:\vert\vert\: p(\mathbf{z})) - \; \mathbb{E}_{q_\phi(\mathbf{z} \vert \mathbf{x})}[\log p_\theta(\mathbf{x} \vert \mathbf{z})] \\
\end{equation}
This loss function is composed of two terms. The first penalizes the Kullback-Leibler divergence ($\text{D}_{\text{KL}}(\cdot\:\vert\vert\:\cdot)$) between the latent space distribution $q_{\phi}(\mathbf{z}\vert\mathbf{x})$ and a prior, chosen as the isotropic unit Gaussian $\mathcal{N}(0,\mathbf{I})$. The second maximizes the marginal likelihood of the original data given this latent space representation. This latter term acts as a reconstruction loss, which penalizes differences between the reconstructed output, and the original input. This is typically calculated in practice using the mean squared difference. Meanwhile, the KL divergence term regularizes the latent space to promote a continuous representation of the underlying input data.

One key contribution of the VAE is its ability to construct disentangled latent spaces, where different latent variables encode semantically distinct features of the data. This is particularly true in the $\beta$-VAE variant, where the two loss terms are differently weighted, with more prominence given to the KL divergence term \cite{higgins_beta-vae_2016, burgess_understanding_2018}.

The $\beta$-TCVAE operates in a similar vein \cite{chen_isolating_2019}. Here, each data sample is indexed by a uniform random variable $n$. Hence, the expected value of the KL divergence term over the dataset can be separated into three components\footnote{A detailed proof is provided by \citet{chen_isolating_2019}.}:
\begin{equation}
    \mathbb{E}_{p(n)}[\text{D}_{\text{KL}}( q(\mathbf{z} \vert \mathbf{x}) \:\vert\vert\: p(\mathbf{z}))] = \text{D}_{\text{KL}}(q(z,n) \vert\vert q(z)p(n)) + \text{D}_{\text{KL}}(q(z) \vert\vert \prod_{j}q(z_j)) + \sum_{j}\text{D}_{\text{KL}}(q(z_j) \vert\vert (z_j)),
\end{equation}

where $j$ represents the latent dimension. Following \citet{chen_isolating_2019}, we can interpret the first two terms as mutual information (MI) terms, and weight the three terms differently using the coefficients $\alpha$, $\beta$ and $\gamma$:

\begin{equation}
    \mathbb{E}_{p(n)}[\text{D}_{\text{KL}}( q(\mathbf{z} \vert \mathbf{x}) \:\vert\vert\: p(\mathbf{z}))] = \alpha I_q(z;n) + \beta C_q(z_1,z_2,...,z_L) + \gamma\sum_{j}\text{D}_{\text{KL}}(q(z_j) \vert\vert (z_j)),
\end{equation}
where $L$ is the dimension of the latent space. The first term $I_q(z;n)$ measures the mutual information between the choice of dataset index $n$ and the corresponding latent vector $\mathbf{z}$, under the empirical data distribution $q(z,n)$. Intuitively, a lower value of this "index-code MI" might represent a higher clustering concentration within latent space of samples throughout the dataset.

The second term $C_q(z_1,z_2,...,z_L)$ represents a multivariate generalization of mutual information called the total correlation, applied over the latent dimensions. This essentially indicates the shared information between latent dimensions. \citet{chen_isolating_2019} hypothesize that heavily penalizing this term (i.e., high $\beta$) would encourage disentangled representations, as information about the value of one latent variable would provide little information about the values of the rest.

Lastly, the third term measures the dimension-wise KL divergence of the latent space distribution; this fulfils the regularization objective.

For feature extraction, we used a convolutional $\beta$-TCVAE. The encoder consisted of four convolutional layers, with 8, 16, 32 \& 64 filters respectively, followed by a fully-connected layer with 256 units, then the sampling layers, each with 32 units. The 32-dimensional latent space representation is then projected back to image space using a decoder network with the reverse architecture, using upsampling layers. The $\beta$-TCVAE was trained for 5 epochs, using a batch size of 256 and a learning rate of 0.0005, with the Adam optimizer. Flips and 90{\textdegree} rotations were used as augmentation.

\subsection{Cell state classification}
\label{section:extended_methods_classification}

For the Dense CNN we used four convolutional layers with 8, 16, 32 \& 64 filters respectively. For the Dense Head, we used four fully-connected layers with 32, 16, 16, 16 and 2 units respectively, with the latter layer providing the final outputs. Scheme 1 \& 2 models were trained using a dropout rate of 0.3, while the Sparse Head was trained with no dropout. The Sparse Head contained the same architecture and initial connectivity as the Dense Head. These configurations were chosen such that Scheme 1, 2 \& 3 models all contain a similar number of parameters ($\sim$380,000 for Scheme 1 and $\sim$316,000 for Schemes 2 \& 3). 

Models from Schemes 1, 2 \& 3 were trained for 100 epochs, using a batch size of 64, with flips and 90{\textdegree} rotations as augmentation. For Scheme 3, we used a dense-training warm up period of $T_{warmup}=20$ epochs. 

\subsection{Sparsity: RigL}
\label{section:extended_methods_rigl}

As with other similar algorithms, \emph{RigL} aims to find the optimal topology of non-zero weights, given some fixed, pre-determined sparsity value $S$. It achieves this with iterative cycles of pruning and re-growth, implemented every $\Delta T$ training iterations (where $\Delta T$ is a hyperparameter). Between each cycle, the network uses gradient descent to find the optimal weight values, as usual. At the beginning of training, $(1-S)\times N^l$ weights of each layer are randomly deactivated, where $N^l$ represents the total number of connection weights at layer $l$. When inactive, the connection weights are set to zero and they do not update between training iterations. At every subsequent pruning step, the smallest $(1-S)\times N^l\times \alpha$ weights (in terms of weight magnitude) are deactivated. Then, at every re-growth step, the same number of weights is chosen for re-growth, this time based on the magnitude of their associated loss gradient. Hence, the top $(1-S)\times N^l\times \alpha$ inactive weights, based on gradient magnitude, are re-grown, with their initial weight value set to zero.

This way, the sparsity value of $S$ is maintained throughout training, with each layer possessing $S\times N^l$ active weights at all times. Hence, \emph{RigL} belongs to the class of pruning techniques that does \emph{not} redistribute sparsities between layers during training. $S$, $\Delta T$, and the update fraction $\alpha$ are all hyper-parameters to be chosen prior to training.

Following \citet{evci_rigging_2019}, we specify a network-wide value for $S$ but allocate to different layers $l$ a separate sparsity value $s^l$. We use the Erdős-Rényi scheme \cite{mocanu_scalable_2018}, in which the layer-specific sparsities scale according to $1-\frac{n^{l}+n^{l+1}}{n^{l}\times n^{l+1}}$, where $n^l$ is the number of input neurons at layer $l$. The actual values for $s^l$ are chosen such that $S=\frac{\sum_{l}s^lN^l}{N}$, where $N$ is the total number of connections in the network. Therefore, the number of inactive connections in each layer is not $S\times N^l$ but $s^l\times N^l$.

We also attenuate the update fraction $\alpha$ as training progresses, using a cosine annealing schedule as the decay function:
\begin{equation}
    f_{decay}(t; \alpha, T_{end}) = \frac{\alpha}{2} \left( 1+\text{cos}\left(\frac{t\pi}{T_{end}}\right) \right) \text{\cite{dettmers_sparse_2019}}.
\end{equation}
With \emph{RigL}, it has been reported that the annealled schedule slightly outperforms a schedule of constant $\alpha$ \cite{evci_rigging_2019}. The use of this schedule adds an additional hyper-parameter $T_{end}$, after which the weight topology does not update.

In our study, we adapt \emph{RigL} by adding two modifications. Firstly, rather than randomly pruning $(1-S)\times N^l$ weights at the beginning of training, we train the model as a dense network for a pre-specified "warm up" period $T_{warmup}$. We observed that this method delivered superior performance results and identified a consistent set of relevant inputs between training runs, benefits we did not observe when randomly pruning at the beginning. We hypothesize that prior dense training allows for the development of useful connection paths, which are then reinforced and retained when the first pruning step is implemented after $T_{warmup}$.

Secondly, we implement two post-training pruning steps that eliminating the following types of connection weight:
\begin{enumerate}
    \item Leaf weight: Connections that feed into a non-output-layer neuron that has no output connections. These connections do not contribute to the output layer at all.
    \item Bias weight: Connections that emerge from a non-input-layer neuron that has no input connections. These are called "bias weights" because at each forward pass, they simply add a fixed value to their target neuron, which is the product of the source neuron's bias and the connection weight value. These "bias weights" can be removed and compensated by adjusting the bias of the target neuron\footnote{We did not use layer or batch normalization to train our sparse networks. If this had been used, the extra parameters would have to have been considered when adjusting the bias of the target neuron.}.
\end{enumerate}
The criteria are implemented in this order. This post-training procedure does not affect the operation of the model at all, but it removes superfluous connection weights, allowing for a more useful assessment of model sparsity. To allow for the innocuous removal of bias weights, we do not use batch normalization when training sparse networks.

\begin{algorithm}
\caption{Modified RigL}
\label{algorithm:rigl}
\begin{algorithmic}
    \State \textbf{Input:} Network $f_{\Theta}$, dataset $D$, learning rate $\eta$
    \comment{$\Theta$ are the initial (un-pruned) weights.}
    
    \hspace{5mm} Sparsity distribution: $\mathbb{S}=\{s^1,...,s^L \}$

    \hspace{5mm} Update schedule: $\Delta T$, $T_{end}$, $\alpha$, $f_{decay}$


    \For {each training step $t$}
        \State Sample a batch $B_t \sim D$ 
        \State $L_t = \sum_{i \in B_t} \mathcal{L}(f_\theta(x_i), y_i)$
        \comment{\textit{$\mathcal{L}$ is the loss function.}}

        \If{$t<T_{warmup}$}
            \State $\Theta = \Theta - \eta\nabla_{\Theta}L_t$
                \comment{Update dense weights (replace with chosen optimizer)}
        \Else
            \If{$(t-T_{warmup})($mod $\Delta T)==0$ and $(t-T_{warmup})<T_{end}$}
                \For {each layer $l$}
                    \State $k = f_{decay}(t-T_{warmup}; \alpha, T_{end})(1-s^l)N^l$
                    \comment{No. of connections to update}
                    \State $\mathbb{I}_{active}=ArgTopK(|\theta^l|, (1-s^l)N^l-k)$
                    \comment{Active connections to keep}
                    \State $\mathbb{I}_{grow}=ArgTopK_{i\notin \mathbb{I}_{active}}(|\nabla_{\Theta^l}L_t|, k)$
                    \comment{Inactive connections to grow}
                    \State $\theta \gets \Theta$ connections included in  $\mathbb{I}_{active} \cup \mathbb{I}_{grow}$
                    \comment{Update topology}
                \EndFor
            \Else
                \State $\theta = \theta - \eta\nabla_{\theta}L_t$
                \comment{Update sparse weights}
            \EndIf
        \EndIf
    \EndFor

    \State $\theta \gets$ Prune leaf and bias weights
    
\end{algorithmic}
\end{algorithm}

\subsubsection{Hyper-parameter tuning}
\label{section:sparsity_tuning}

As described in \S\ref{section:methods_rigl}, \emph{RigL} operates using a small set of hyper-parameters: the sparsity value $S$ (which is the fraction of weights set to zero), update fraction $\alpha$ and update interval $\Delta T$, as well as the end-step $T_{end}$ for the update fraction decay schedule. Hyper-parameter search was conducted on three of these four values - all except $T_{end}$, which was set to $T_{end}=0.75$.

For this, we used the \textsc{Optuna} \cite{akiba_optuna_2019} package, a Python interface that access several optimization algorithms, including grid search and random search, as well as Bayesian methods such as the tree parzen estimator \cite{bergstra_algorithms_2011}. We ran the \textsc{Optuna} search for 200 trials, with each trial querying a unique hyper-parameter configuration. Each trial consisted of five runs, each being run for a reduced duration of 40 epochs. The value ranges we provided reflect our experience from prior experiments: 0.95-0.97 for $S$, 100-200 iterations for $\Delta T$ and 0.7-0.9 for $\alpha$. However, in principle, optimization algorithms such as \textsc{Optuna} can output sensible hyper-parameter values in the absence of any prior knowledge (e.g., by setting the range of $S$ to 0.0-1.0). For our objective function, we used the sum of the validation accuracy and the final sparsity (after the post-training pruning steps), with each of these being averaged over the five runs associated with each trial. Hence, the goal is both to maximize classification performance and minimize the number of active weights in the final model.

While this hyper-parameter search was done in order to generally optimize classification performance, the search for an optimal $S$ has particular significance, for it represents the maximum sparsity appropriate to a particular task. This reflects both the complexity of the inputs required and the complexity of the target function itself. Hence, our objective function - the sum of the accuracy and the final sparsity - reflects the principle of Occam's razor.

The values yielded were $S=0.951$, $\Delta T=115$ and $\alpha=0.758$.

\subsection{PySR implementation}
\label{section:extended_methods_pysr}

We configured our regression models such that they could choose between four binary operators ($+, \times, -$ and $\div$) and six unary operators ($x^2$, $e^{x}, \text{log}(x), \text{sin}(x), \sqrt{x}$ and $|x|$) in the expression trees. We penalized all operators, constants and variables in the tree with a complexity score of 1, except for the $\text{sin}(x)$ (score of 3), $e^{x}$ (score of 2) and $\text{log}(x)$ (score of 2) functions. We used a parsimony value of 0.001 to scale the complexity score in the overall loss function, and we set 20 as a hard limit on the complexity of our expressions.


\section{Calculation of head expression size}
\label{appendix:head_size_calculation}

\begin{figure}[ht!]
     \centering
     \begin{subfigure}[b]{0.22\textwidth}
         \centering
         \includegraphics[width=\textwidth]{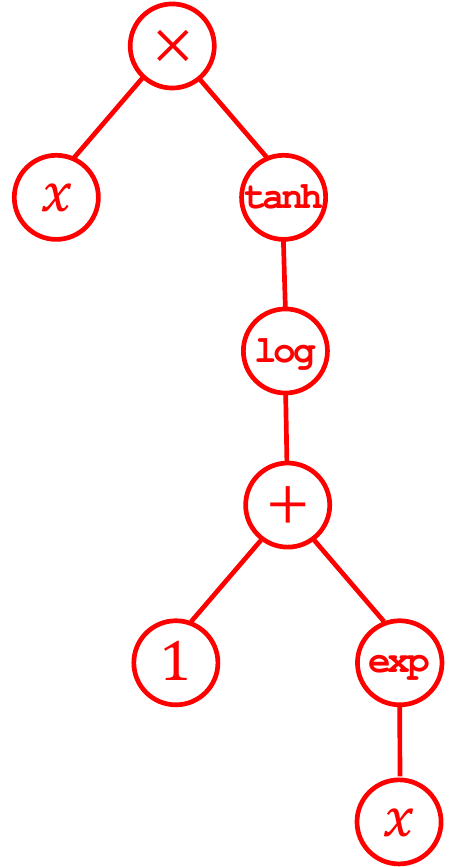}
         \caption{Mish activation function.}
     \end{subfigure}
     \hfill
     \begin{subfigure}[b]{0.42\textwidth}
         \centering
         \includegraphics[width=\textwidth]{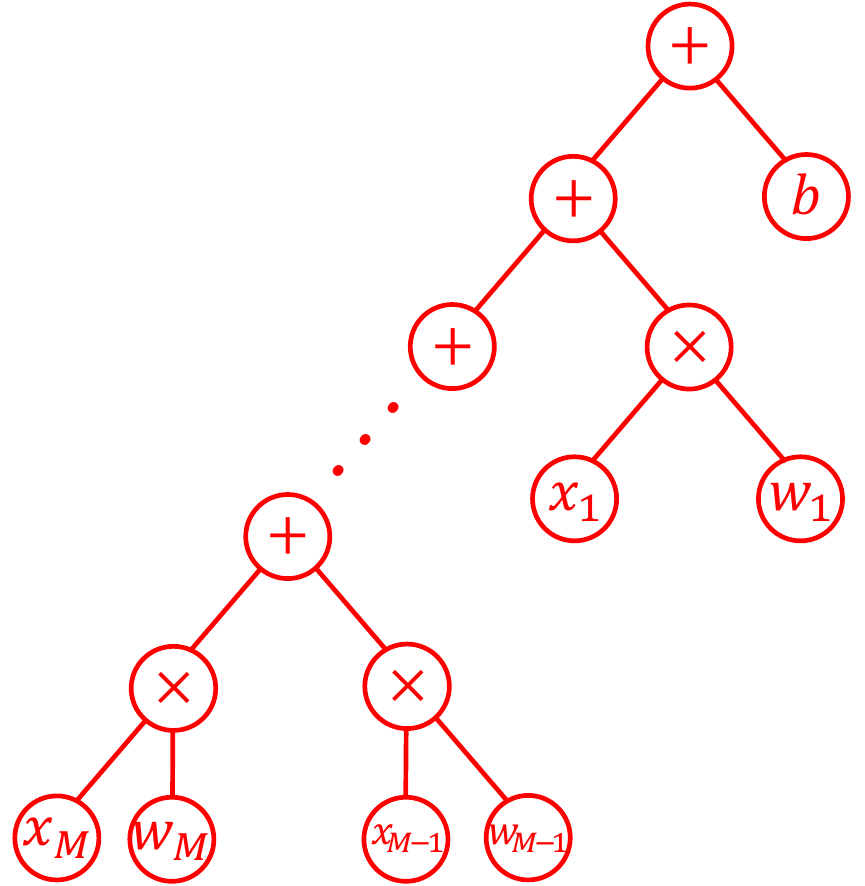}
         \caption{Addition of neuron bias and input terms.}
     \end{subfigure}
        \caption{Expression trees. $M$ is the total number of source neurons.}
        \label{fig:mish_trees}
\end{figure}

The goal here is to transform a neural network into a symbolic expression. When a neuron at layer $i$ accepts inputs $x$ from connected neurons $m$ at layer $i-1$, the output $y$ of the neuron will be:
\begin{equation}
\label{eq:neuron_output}
    y = f \left(b + \sum_m x_m w_{mn} \right),
\end{equation}
where $f$ is the activation function and $w_{m}$ is the weight connecting neuron $m$ at layer $i-1$ to the present neuron at layer $i$. We use the activation function \emph{Mish} \cite{misra_mish_2020}, which can be written as
\begin{equation}
\label{eq:mish}
    f(x) = x\cdot\text{tanh}(\text{log}(1+e^x)).
\end{equation}
The expression tree representation of this function is shown in Fig. \ref{fig:mish_trees}a. The expression inside $f(\cdot)$ in Eq. \ref{eq:neuron_output} can likewise be represented by the tree shown in Fig. \ref{fig:mish_trees}b.

\renewcommand{\figwidth}{0.05}

It is possible to see that for any given number of source neurons $M$, the expression size of the tree in Fig. \ref{fig:mish_trees}b is $4M+1$. Therefore, when one substitutes the tree in Fig. \ref{fig:mish_trees}b for each $x$ term in the \emph{Mish} tree in Fig. \ref{fig:mish_trees}a, one can calculate the total expression size as $8(M+1)$. Therefore, a pair of consecutive layers fully connected to each other, with $M$ and $N$ numbers of neurons respectively, can be represented by an expression tree with size $8N(M+1)$. This is in the absence of normalization.

For all our classification heads, no activation was used in the final layer; therefore, for this layer, the expression size is simply $N(4M + 1)$. For Scheme 2 models with $I$ layers and layer-wise neuron counts $n_1,n_2,...,n_I$, the total expression size $E$ is therefore
\begin{equation}
\label{eq:scheme2_exp_size}
    E = n_I(4n_{I-1}+1) + \sum^{I-2}_{i=1} 8n_{i+1}(n_i+1).
\end{equation}

For Scheme 1 models, this calculation is complicated by the fact that batch normalization \cite{ioffe_batch_2015} is used. Batch normalization reduces internal covariate shift by normalizing layer-wise inputs according to the batch-wide mean and variance. The output $y$ of this normalization is
\begin{equation}
    y = \frac{x-\text{E}[x]}{\sqrt{\text{Var}[x]+\epsilon}}\cdot\gamma + \beta,
\end{equation}
where $\text{E}[x]$ and $\text{Var}[x]$ represent the batch-wise mean and variance respectively, $\epsilon$ is a small numerical stabilizer and $\gamma$ and $\beta$ are trainable parameters to scale and shift the final output. This is done on a per-neuron basis, and before any activation. Furthermore, the mean and variance terms are calculated during training, but for our models, they are frozen during evaluation, so they do not involve any calculation during the forward pass.

\begin{figure}
 \centering
       \includegraphics[width=0.37\textwidth]{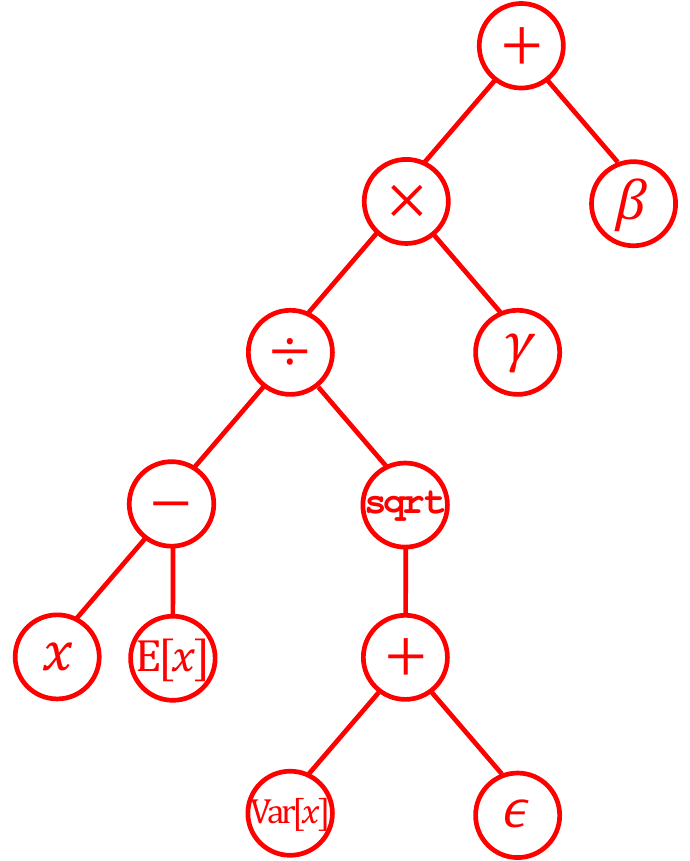}
       \caption{Expression tree for batch normalization.}
    \label{fig:batch_norm_tree}
\end{figure}

The expression tree for batch normalization is shown in Fig. \ref{fig:batch_norm_tree}. The size of this tree is 12, therefore the size of the input to activation is $4M+12$, so the total expression size of the neuron is $8M+30$. The total expression size for the feed-forward network, again omitting normalization and activation from the final layer, is then
\begin{equation}
\label{eq:scheme1_exp_size}
    E = n_I(4n_{I-1}+1) + \sum^{I-2}_{i=1} n_{i+1}(8n_i+30).
\end{equation}

Finally, for Scheme 3 models, the expression size can be calculated by summing $8M_j+1$ (or $4M_j+1$ for the final layer) over the target neurons with active connections, where $M_j$ is the number of source neurons connected to each target neuron $j$. More precisely, for layers $i$ with layer-wise active neuron count $J_i$, the whole network expression size is
\begin{equation}
\label{eq:scheme3_exp_size}
    E = \left( \sum_{j=1}^{J_I}4M_j+1 \right) + \left(\sum_{i=1}^{I-1}\sum_{j=1}^{J_i}8M_j+1 \right).
\end{equation}

\newpage

\section{Symbolic expressions obtained}
\label{section:extended_symbolic}

\renewcommand{\arraystretch}{1.5}
\begin{table}[h!]
\begin{center}
\centering
\begin{tabularx}{\textwidth} { 
  | >{\centering\arraybackslash\hsize=0.05\hsize}X 
  | >{\centering\arraybackslash\hsize=0.56\hsize}X 
  | >{\centering\arraybackslash\hsize=0.22\hsize}X 
  | >{\centering\arraybackslash\hsize=0.17\hsize}X 
  | }
 \hline
 \textbf{No.} & \textbf{Expression} & \textbf{Expression size} & \textbf{Accuracy} \\
 \hline
 1 & $z_{29}(z^2_{17}+z^2_{21})-e^{e^{z_3}}$  &  11  &  97.6\%  \\
 \hline
 2 & $0.83(z_{29} - 1.37z_3 - |z_3|)(z^2_{17} + z^2_{21} - 0.19) - 1.74$  &  20  &  97.8\%  \\
 \hline
 3 & $(z_{29}-z_3)(|z_{17}|+z^2_{21})-2.88$  &  11  &  97.4\%  \\
 \hline
 4 & $0.74(z_{29} - 0.62z_3)(z^2_{17} + z^2_{21}) - 2.11$  &  15  &  97.5\%  \\
 \hline
 5 & $(z^2_{17} + z^2_{21})(z_{29}-\text{sin}(z_3)) - 2.86$  &  12  &  97.3\%  \\
 \hline
 6 & $0.74(z^2_{17} + z^2_{21})(z_{29}-\text{sin}(z_3+0.18))-1.99$  &  16  &  97.4\%  \\
 \hline
 7 & $0.71(z^2_{17} + z^2_{21})(z_{29} - \frac{z_3}{1.4\sqrt{|z_3|}})-2.11$  &  19  &  97.6\%  \\
 \hline
 8 & $(z^2_{17} + z^2_{21})(z_{29} - e^{z_3} + 0.71) - 2.03$  &  14  &  97.4\%  \\
 \hline
 9 & $0.70(z^2_{17} + z^2_{21})(z_{29}-\frac{z_3}{z^2_3+0.66}) - 2.11$  &  19  &  97.5\%  \\
 \hline
 10 & $(z_{29} - 0.44z_3)(z^2_{17} + z^2_{21}) - 2.56$  &  13  &  97.4\%  \\
 \hline
\end{tabularx}
\caption{Hinge loss models. Symbolic expression, testing accuracy and complexity across ten Scheme 4 models trained using hinge loss. "Expression size" is the number of nodes in the expression tree. It differs from "complexity" since the latter is calculated in accordance with the additional penalty placed on $\text{sin}(x)$, $e^{x}$ \& $\text{log}(x)$.}
\label{table:symbolic_accuracy_hinge}
\end{center}
\end{table}
\renewcommand{\arraystretch}{1}

\renewcommand{\arraystretch}{1.5}
\begin{table}[h!]
\begin{center}
\centering
\begin{tabularx}{\textwidth} { 
  | >{\centering\arraybackslash\hsize=0.05\hsize}X 
  | >{\centering\arraybackslash\hsize=0.56\hsize}X 
  | >{\centering\arraybackslash\hsize=0.22\hsize}X 
  | >{\centering\arraybackslash\hsize=0.17\hsize}X 
  | }
 \hline
 \textbf{No.} & \textbf{Expression} & \textbf{Expression size} & \textbf{Accuracy} \\
 \hline
 1 & $2.32|z_{17}| + z^2_{21} + 4.46z_{29} - 3.16(0.56z_3 + 1)^2 - 6.15$  &  18  &  97.4\%  \\
 \hline
 2 & $z^2_{21} + 4.22(\sqrt{|z_{17}|} + \text{sin}(z_{29}) - \text{sin}z_3)) - 10.54$  &  16  &  96.9\%  \\
 \hline
 3 & $z^2_{17} + z^2_{21} + z_{29} - 6.00e^{\text{sin}(z_3)}$  &  13  &  97.0\%  \\
 \hline
  4 & $z^2_{17} + z^2_{21} + z_{29} - z^2_3 - 8.20(0.35z_3 + 1)^2 + 2.13$  &  17  &  96.9\%  \\
 \hline
 5 & $2|z_{17}| + z^2_{21} + 3.72z_{29} - 3.72(0.52z_3 + 1)^2 - 4.94$  &  18  &  97.3\%  \\
 \hline
 6 & $3.79z_{29} - |z^2_{17} + z^2_{21} - 3.93(0.50x_3 + 1)^2 - 5.49| + 1.45$  &  19  &  97.0\%  \\
 \hline
 7 & $z^2_{17} + z^2_{21} + (z_3 + 3.99)(z_{29} - z_3) - 6.85$  &  15  &  97.4\%  \\
 \hline
 8 & $z^2_{17} + (|z_{21}| - z_3 + 2.55)(e^{\text{sin}(z_{29})}) - 10.00$  &  15  &  97.0\%  \\
 \hline
 9 & $z^2_{17} + z^2_{21} + 4.32(\text{sin}(z_{29}) - \text{sin}(z_{3})) - 8.77$  &  18  &  97.1\%  \\
 \hline
 10 & $(z_{17} - 0.25)^2 + z^2_{21} + 3.76z_{29} - 3.16e^{z_3} - 3.53$  &  18  &  97.1\%  \\
 \hline
\end{tabularx}
\caption{MSE loss models. Symbolic expression, testing accuracy and complexity across ten Scheme 4 models trained using MSE loss.}
\label{table:symbolic_accuracy_mse}
\end{center}
\end{table}
\renewcommand{\arraystretch}{1}

\newpage

\section{Neighborhood features}
\label{section:extended_traversals}
\renewcommand{\figwidth}{0.87}
\begin{figure}[!b]
 \centering
    \begin{subfigure}[b]{\figwidth\textwidth}
      \includegraphics[width=\textwidth]{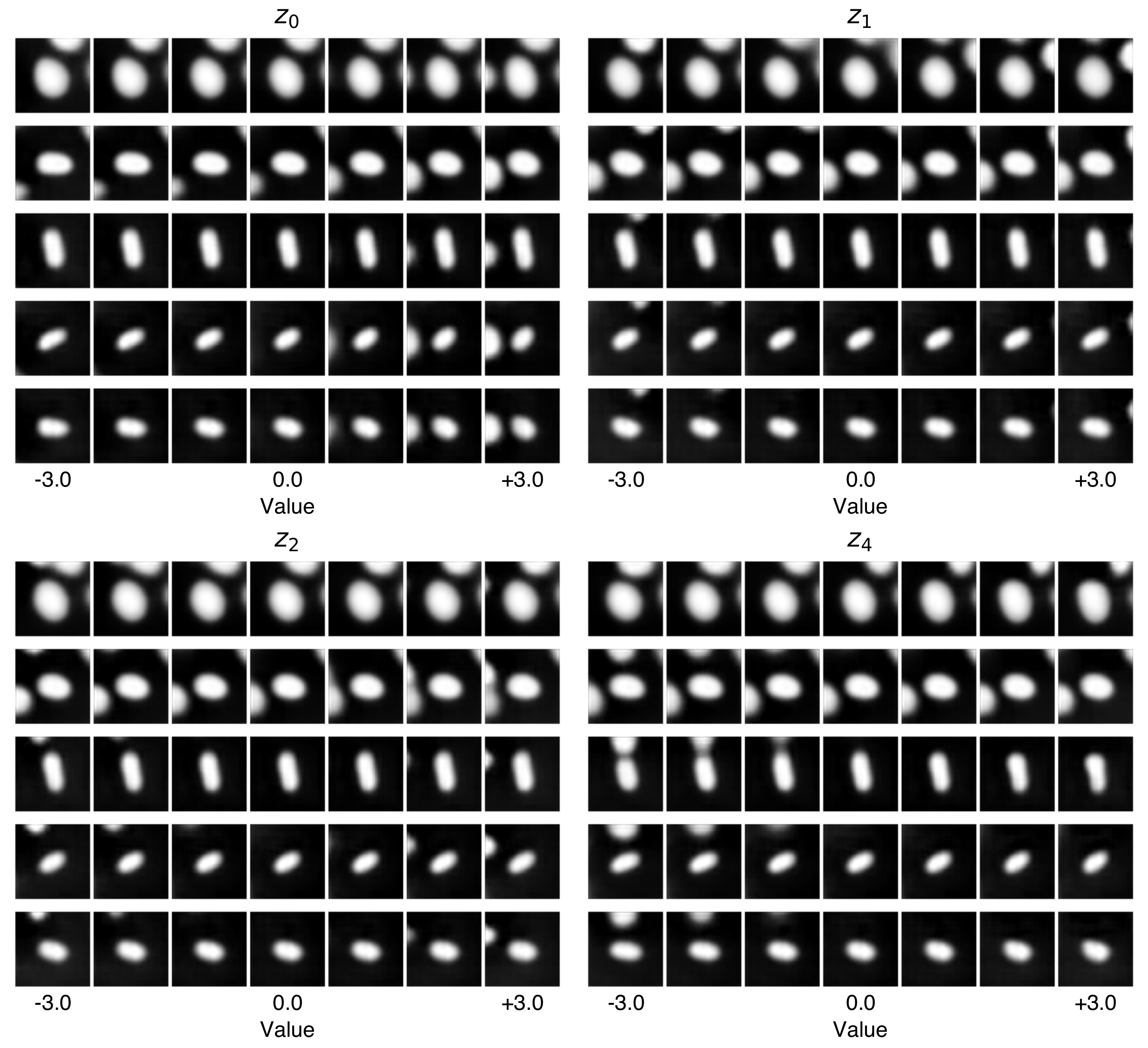}
      \caption{Some latent variables that encode neighborhood features.}
    \end{subfigure}
    \begin{subfigure}[b]{\figwidth\textwidth}
      \includegraphics[width=\textwidth]{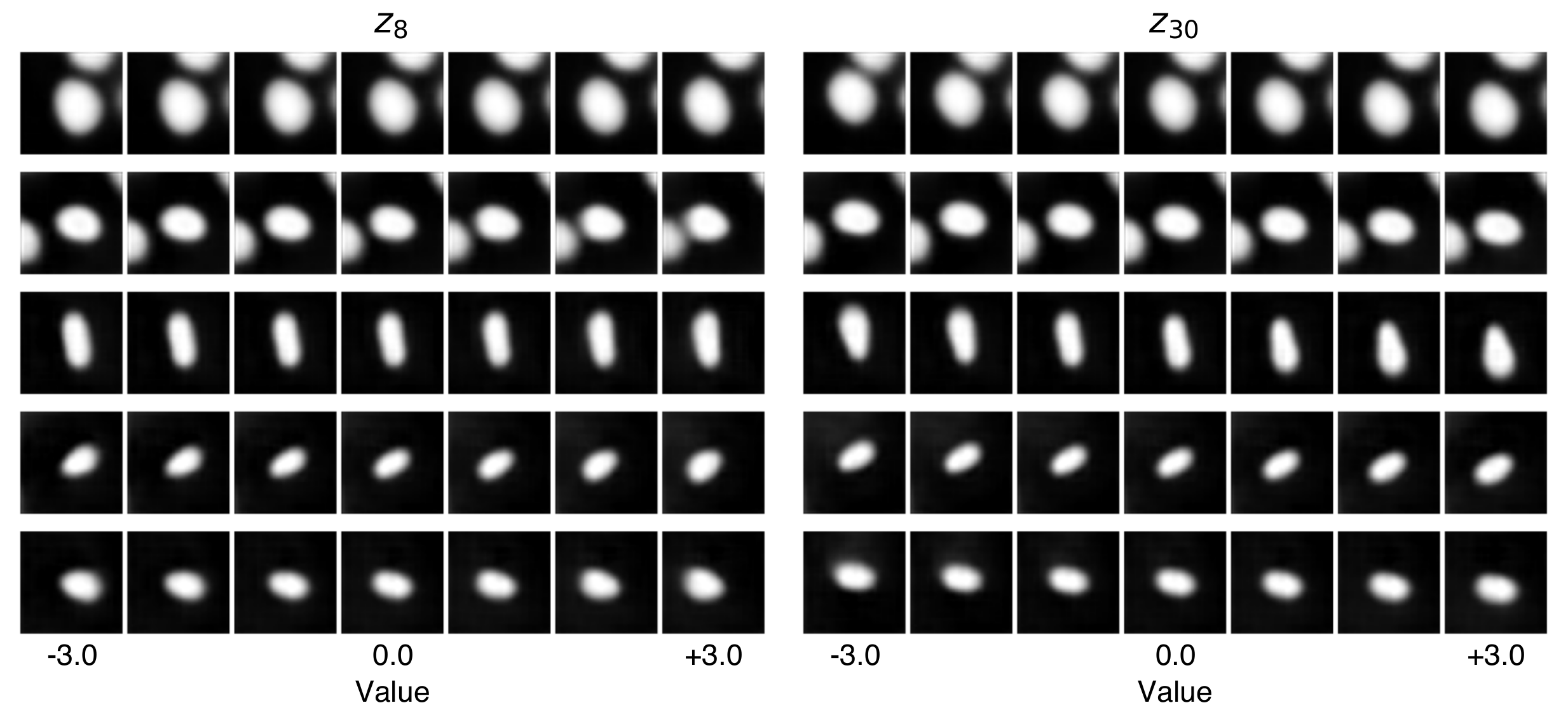}
      \caption{Latent variables that encode central cell position: $z_8$ ($x$-position) \& $z_{30}$ ($y$-position)}
    \end{subfigure}
    \caption{Latent variables that encode neighborhood features and central cell position.}
    \label{fig:latent_variables_neighborhood}
\end{figure}

\newpage
\section{Sparse model topologies}
\label{appendix:sparse_model_topologies}

\def\width{0.3}
\begin{figure}[!b]
     \centering
     \begin{subfigure}[b]{\width\textwidth}
         \centering
         \includegraphics[width=\textwidth]{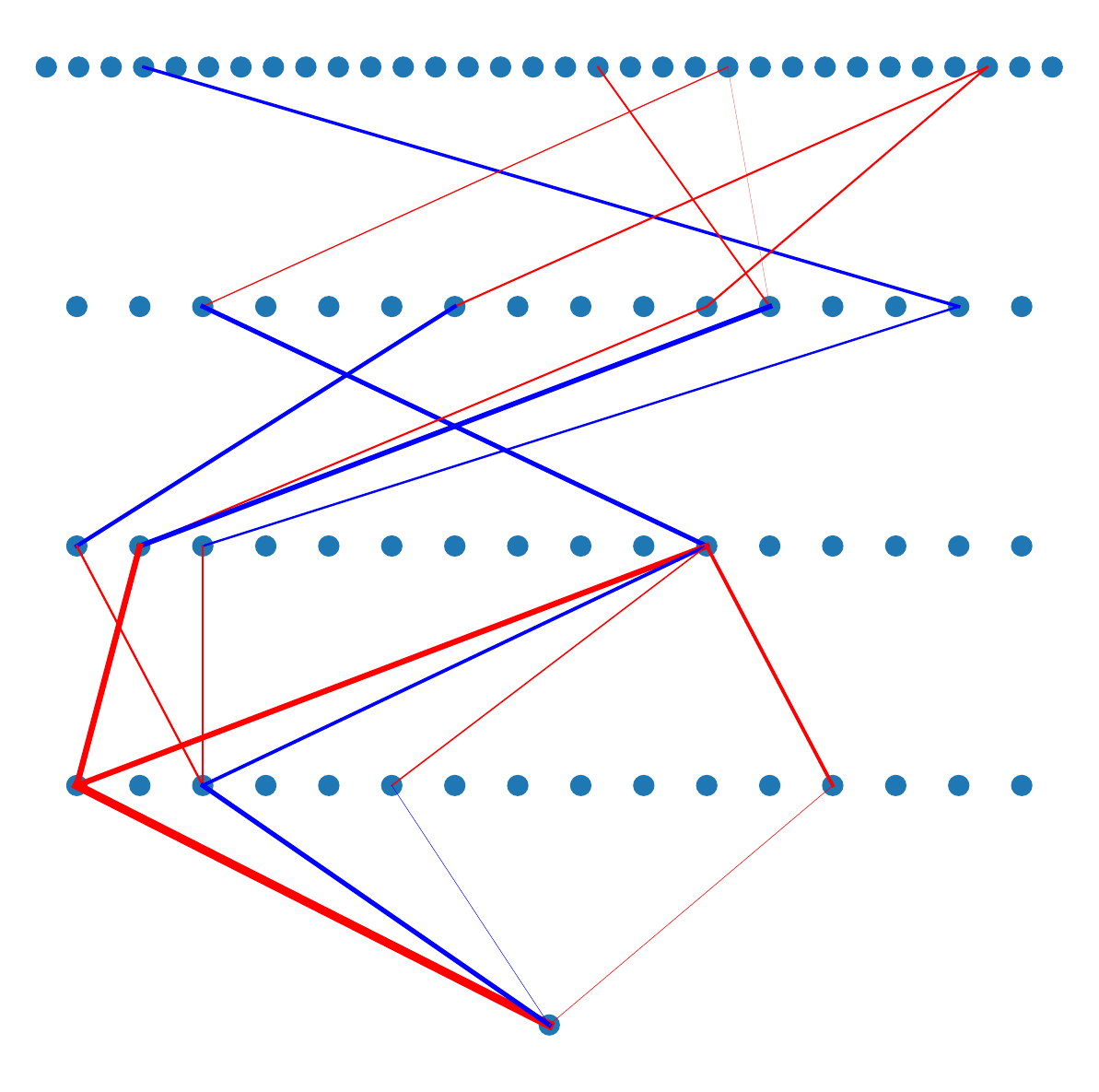}
     \end{subfigure}
     \hfill
     \begin{subfigure}[b]{\width\textwidth}
         \centering
         \includegraphics[width=\textwidth]{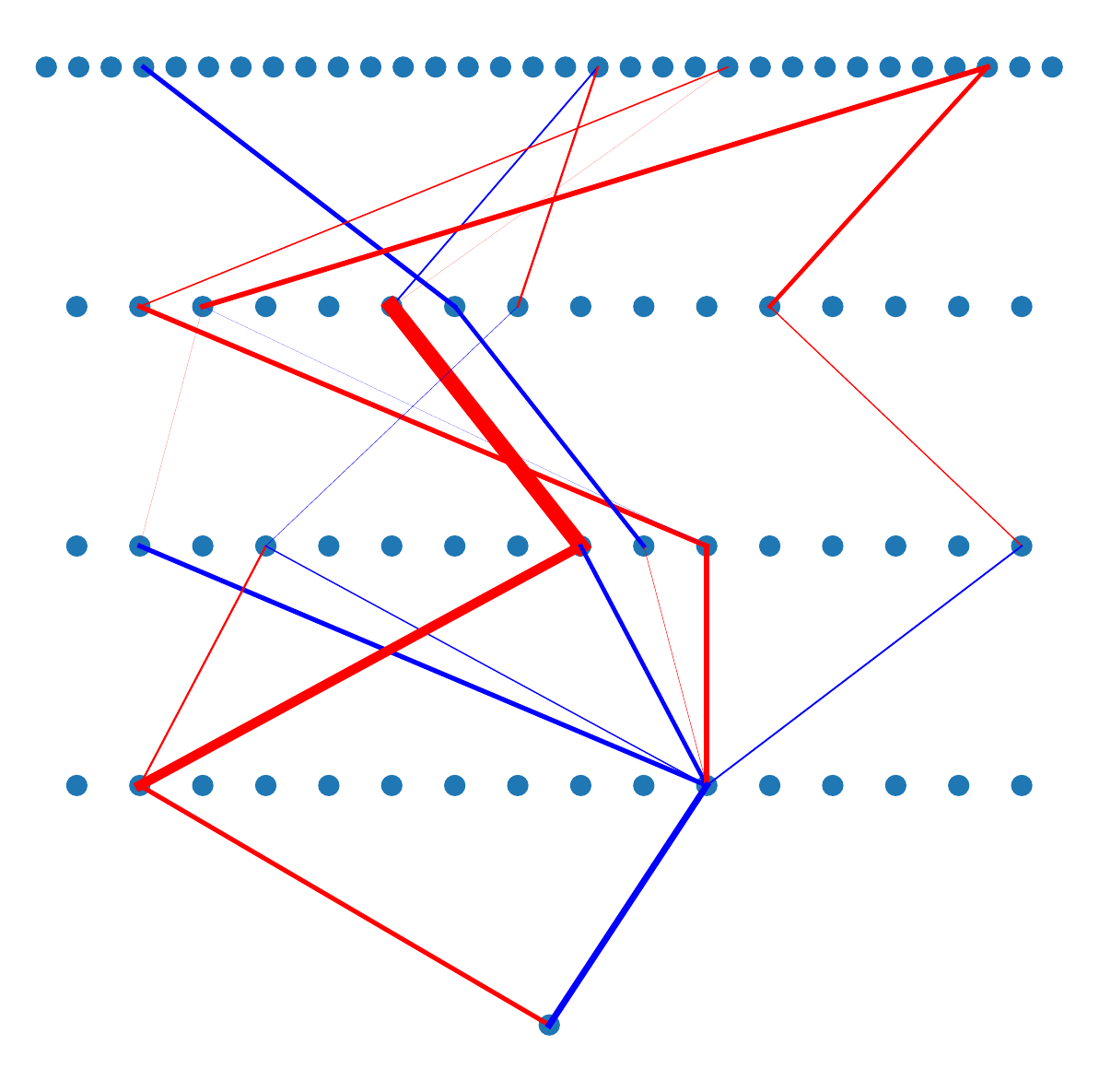}
     \end{subfigure}
     \hfill
     \begin{subfigure}[b]{\width\textwidth}
         \centering
         \includegraphics[width=\textwidth]{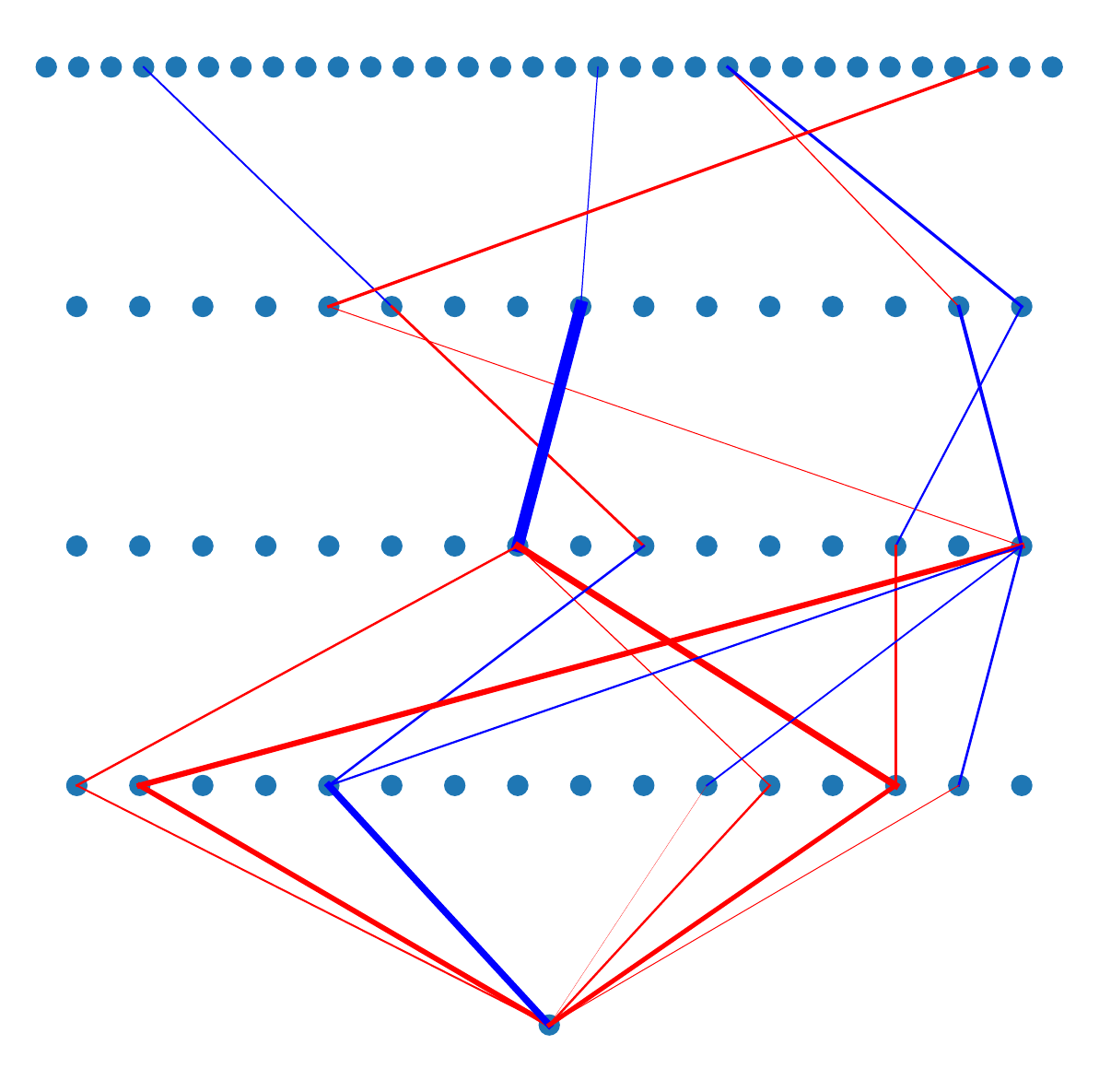}
     \end{subfigure}
     \hfill
     \begin{subfigure}[b]{\width\textwidth}
         \centering
         \includegraphics[width=\textwidth]{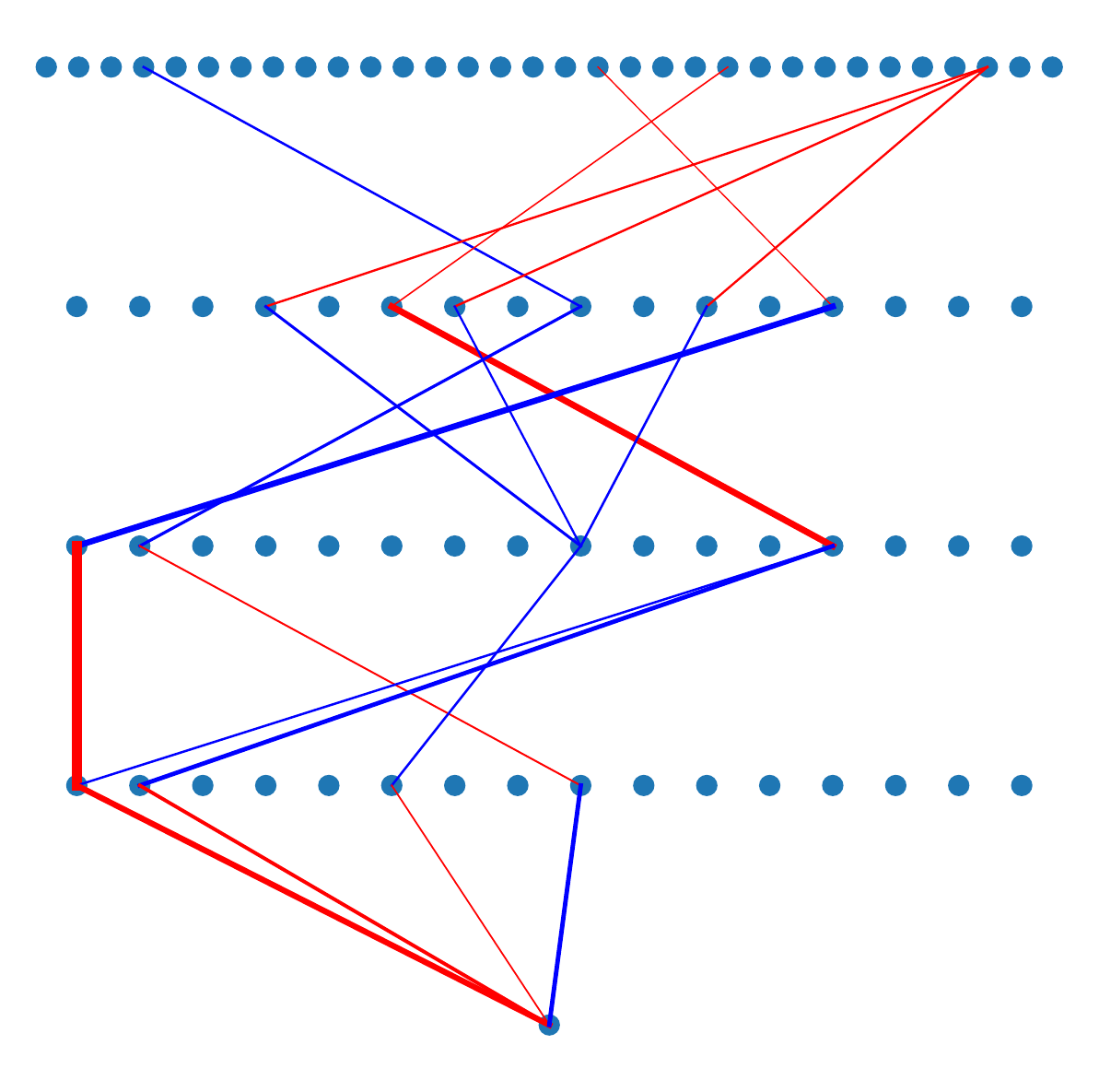}
     \end{subfigure}
     \hfill
     \begin{subfigure}[b]{\width\textwidth}
         \centering
         \includegraphics[width=\textwidth]{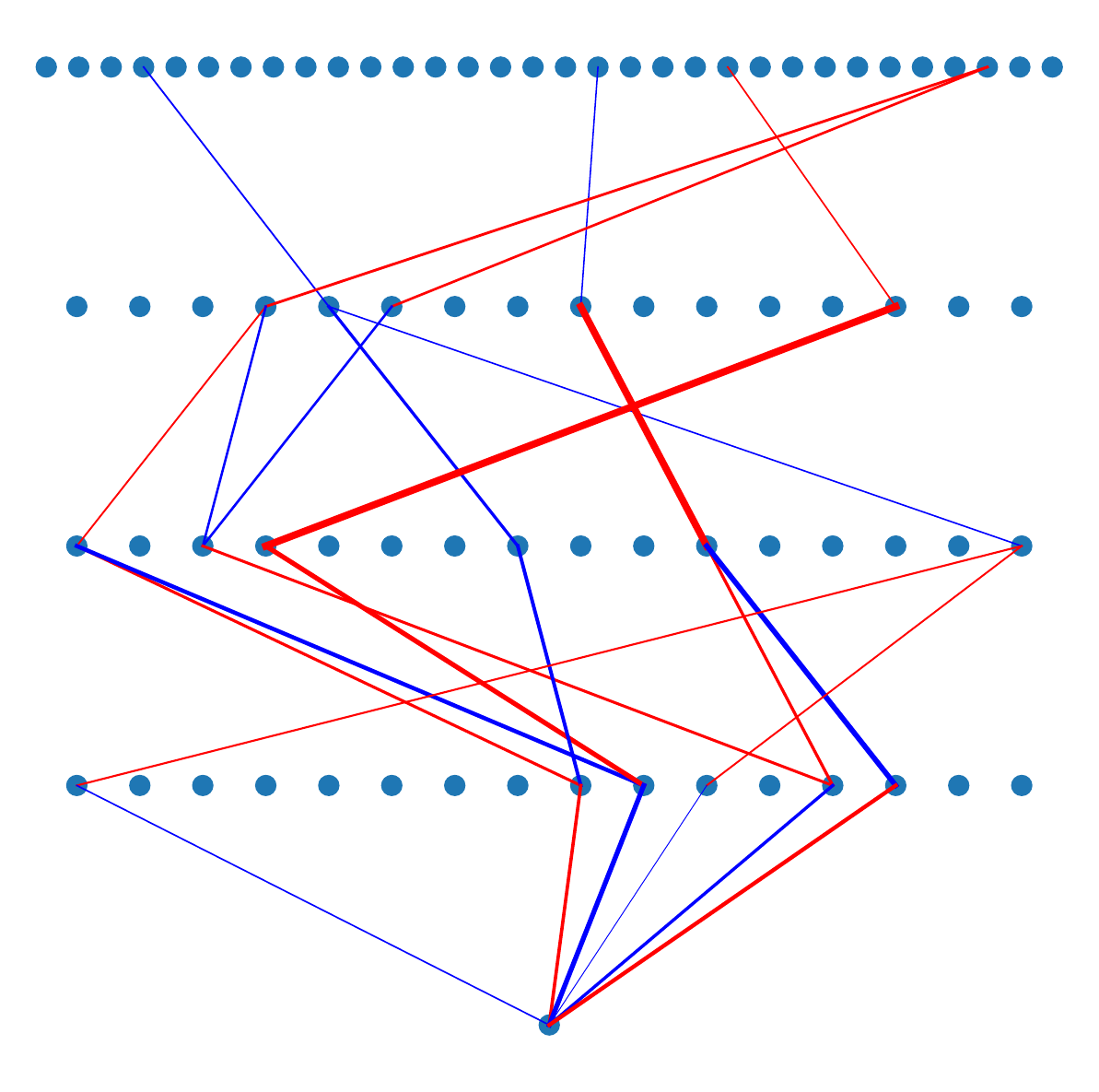}
     \end{subfigure}
     \hfill
     \begin{subfigure}[b]{\width\textwidth}
         \centering
         \includegraphics[width=\textwidth]{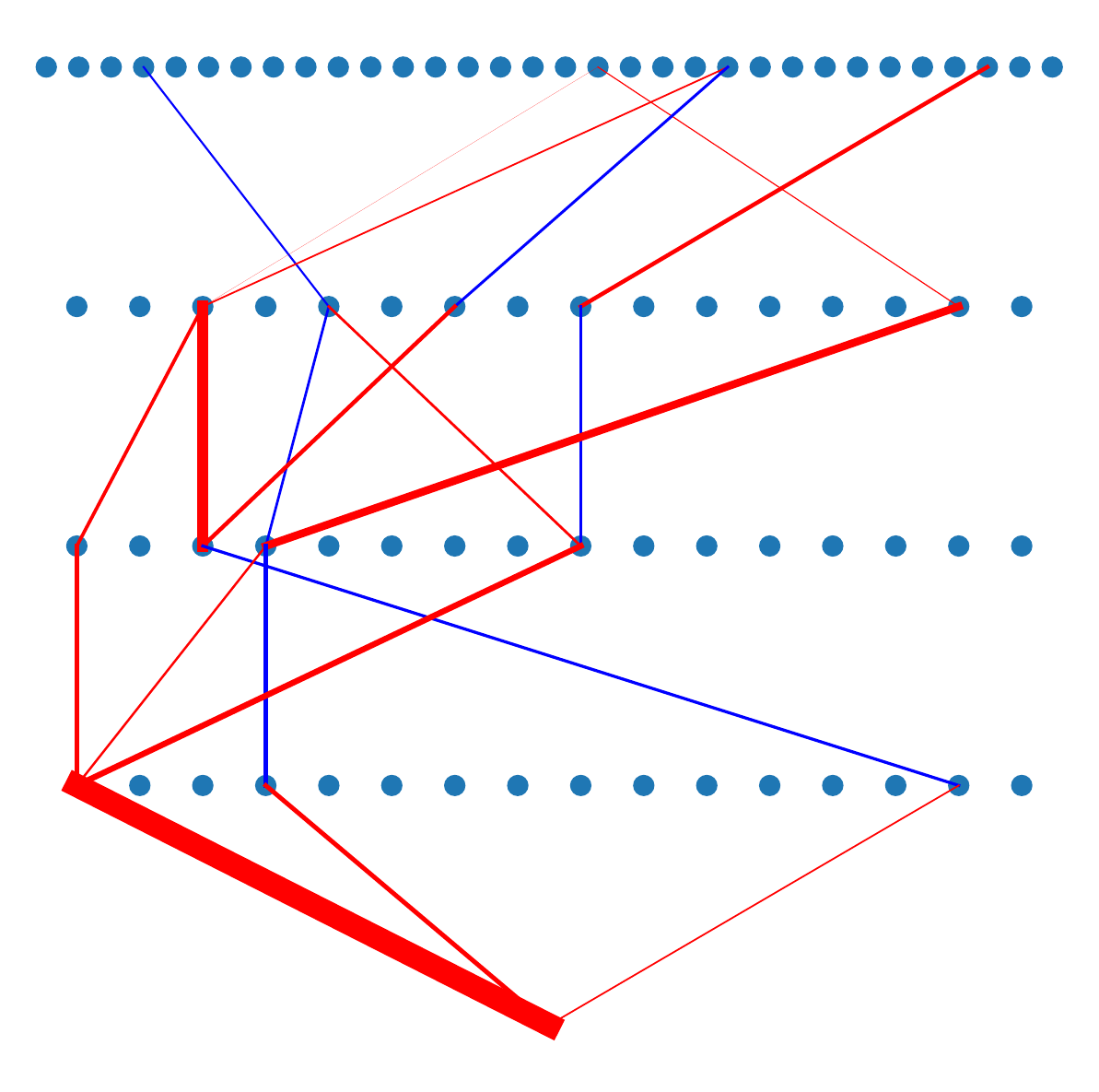}
     \end{subfigure}
     \hfill
     \begin{subfigure}[b]{\width\textwidth}
         \centering
         \includegraphics[width=\textwidth]{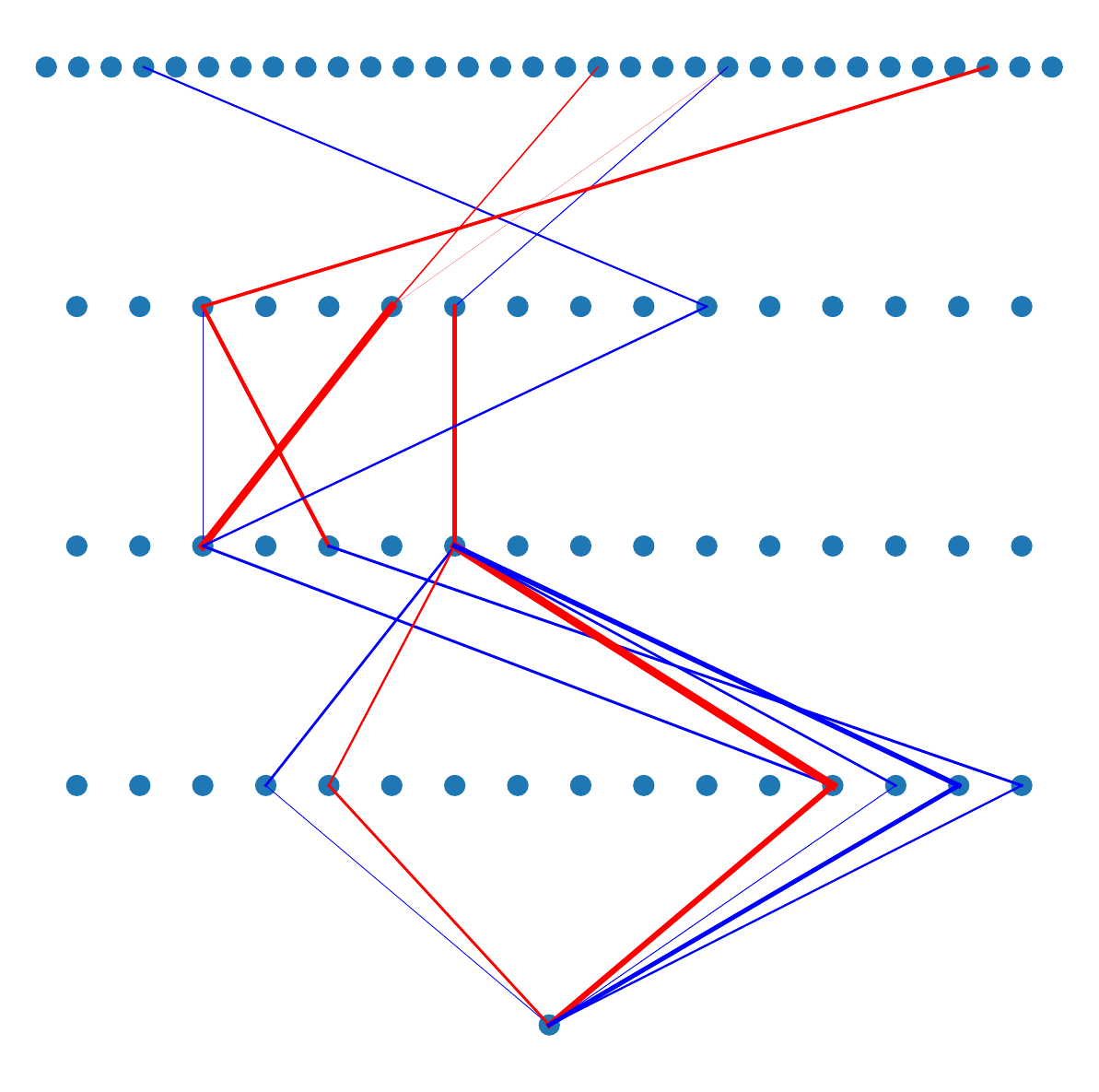}
     \end{subfigure}
     \hfill
     \begin{subfigure}[b]{\width\textwidth}
         \centering
         \includegraphics[width=\textwidth]{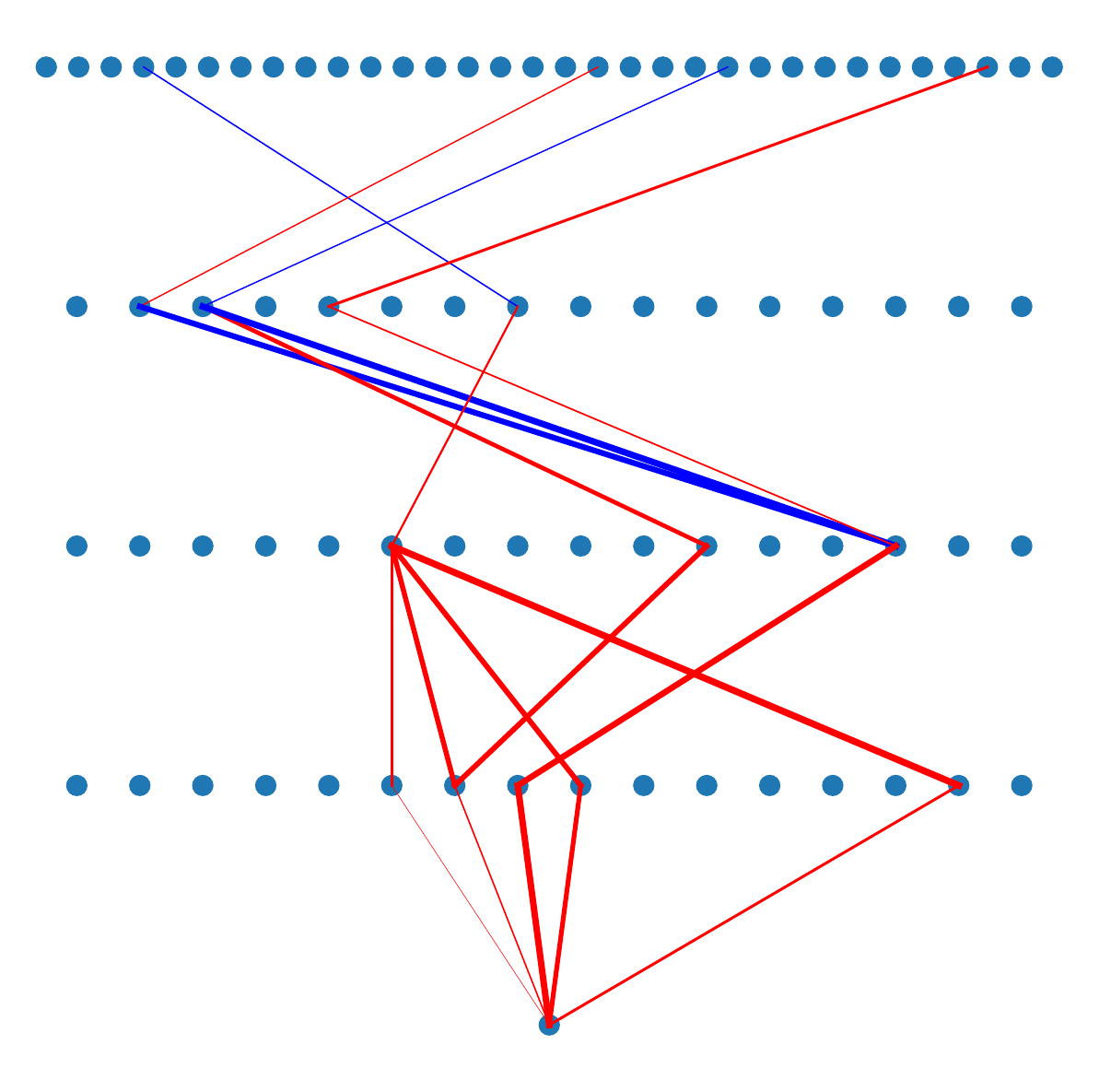}
     \end{subfigure}
     \hfill
     \begin{subfigure}[b]{\width\textwidth}
         \centering
         \includegraphics[width=\textwidth]{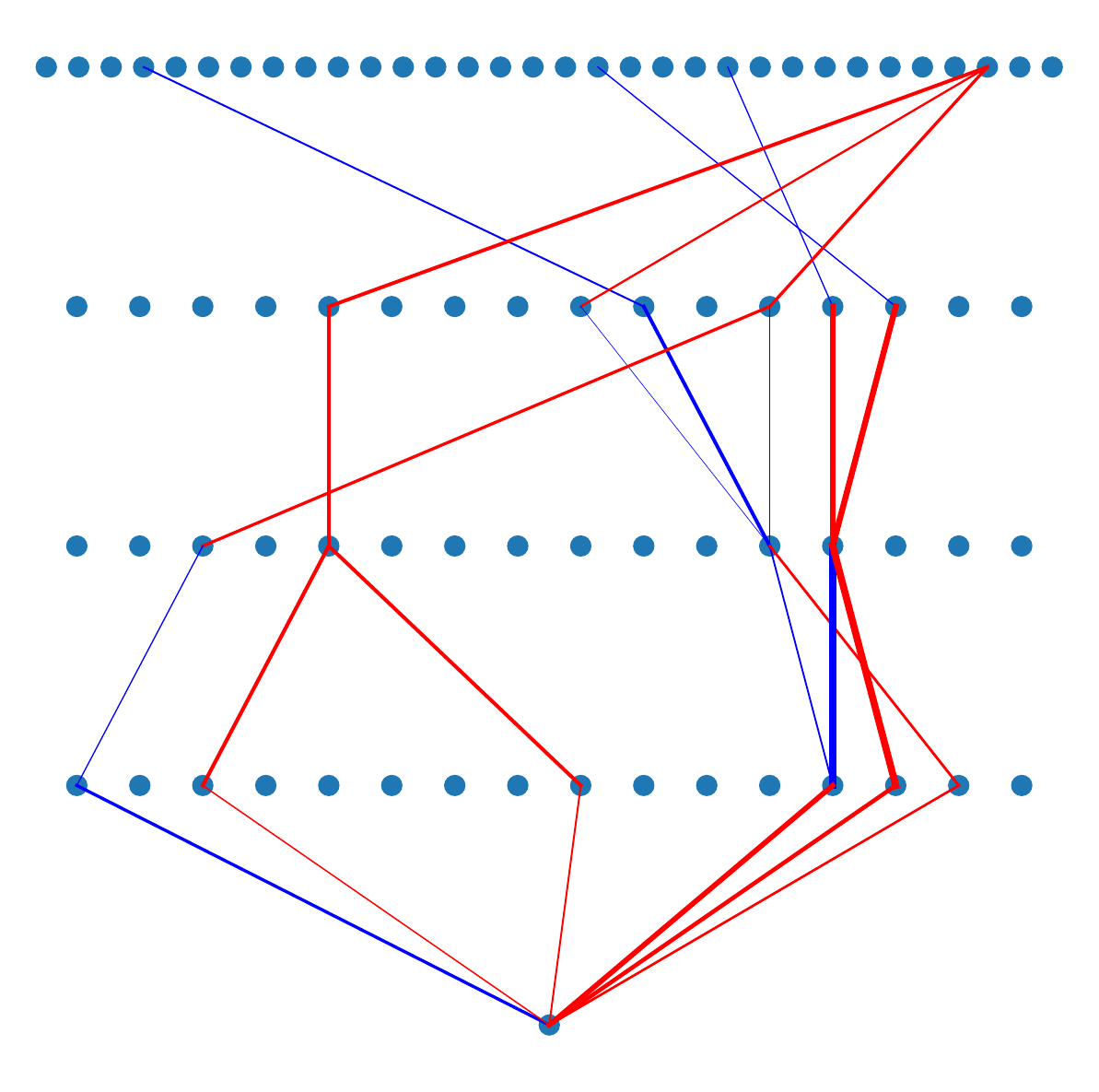}
     \end{subfigure}
     \hfill
     \begin{subfigure}[b]{\width\textwidth}
         \centering
         \includegraphics[width=\textwidth]{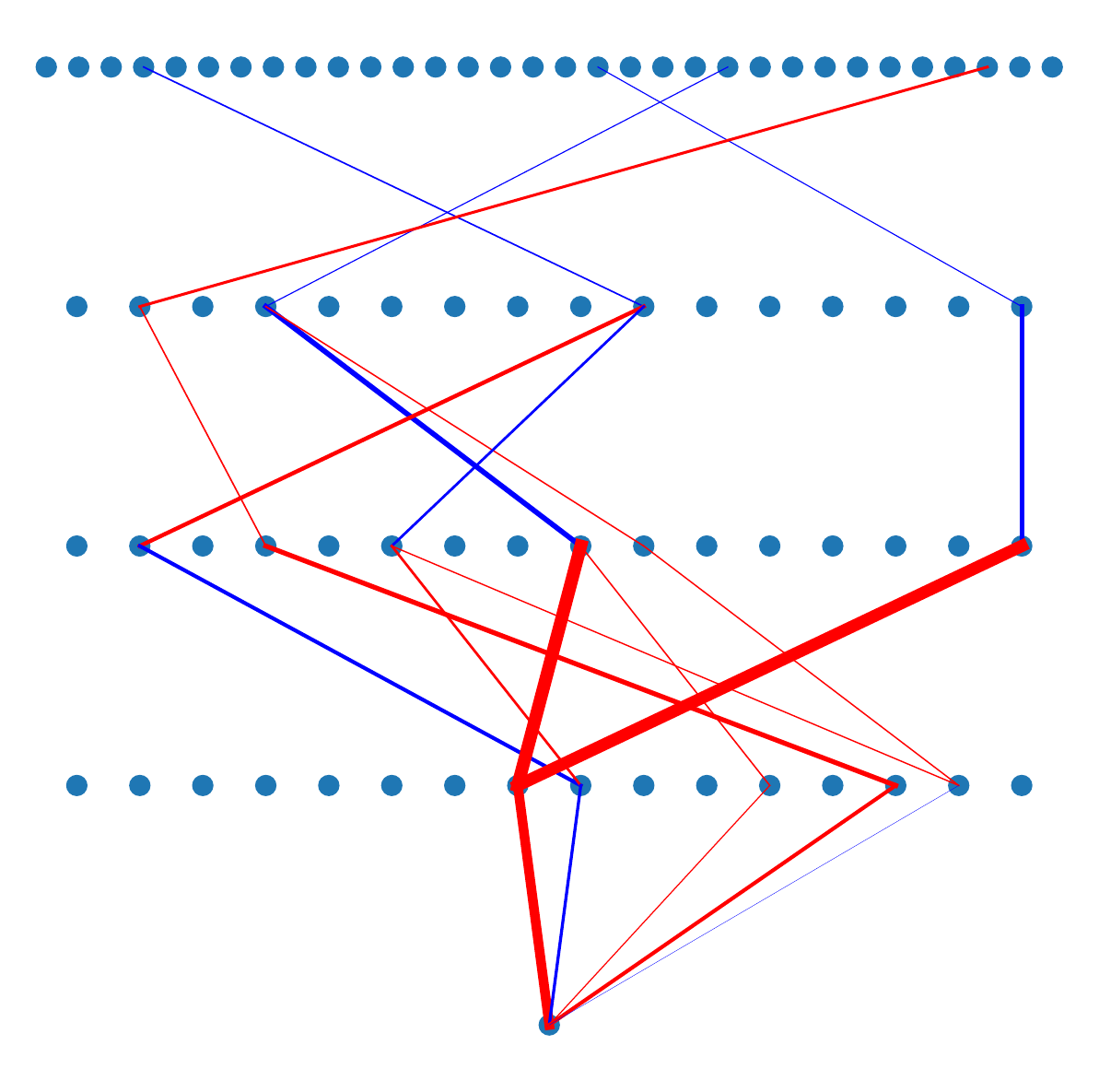}
     \end{subfigure}
        \caption{Scheme 3 model topologies. Blue connections are positively weighted, red connections are negatively weighted. The thickness of the connection line is proportional to its weight magnitude. The network flow is up to down, so the top layer is the input layer and the bottom layer is the output layer.}
        \label{fig:topologies}
\end{figure}

\newpage
\section{Failure modes}

One significant benefit of interpretability is that it allows us to understand how and why our models fail when faced with out-of-distribution (OOD) data. Such situations can occur when the data input system produces erroneous outputs, or when the training data insufficiently represents the full range of possible inputs. One common example in automated microscopy is the collection of blank images, where all the pixels are zero-valued. This occurrence can arise from asynchrony between LED illumination and camera capture during experimental data collection.

Blank images are not part of the training dataset of the $\beta$-TCVAE, therefore we would expect their reconstructions to be poor (Fig. \ref{fig:blank_image_reconstruction}). However, we are also interested in studying the output of our classification models when applied on them. All ten of our sparse models (Appendix \ref{appendix:sparse_model_topologies}) and all four of the symbolic expressions we analyzed (\S\ref{section:decision_boundary}) classified these blank images as interphase. A cursory glance at the latent space encoding of the blank image reveals why (Fig. \ref{fig:blank_image_encoding}). In the eccentricity sub-space (consisting of variables $z_{17}$ \& $z_{21}$). In the size sub-space ($z_{3}$ \& $z_{29}$), the value of $z_{29}$ is negative; however, the value of $z_{3}$ is even lower. With these statements, it is possible to rationalize how we would obtain a negatively-valued output from our symbolic expressions, as well as the one sparse network we analyzed (\S\ref{section:sparse_network_analysis}).

\renewcommand{\expwidth}{0.3}
\begin{figure}[h!]
 \centering
    \begin{subfigure}[b]{\expwidth\textwidth}
    \includegraphics[width=\textwidth]{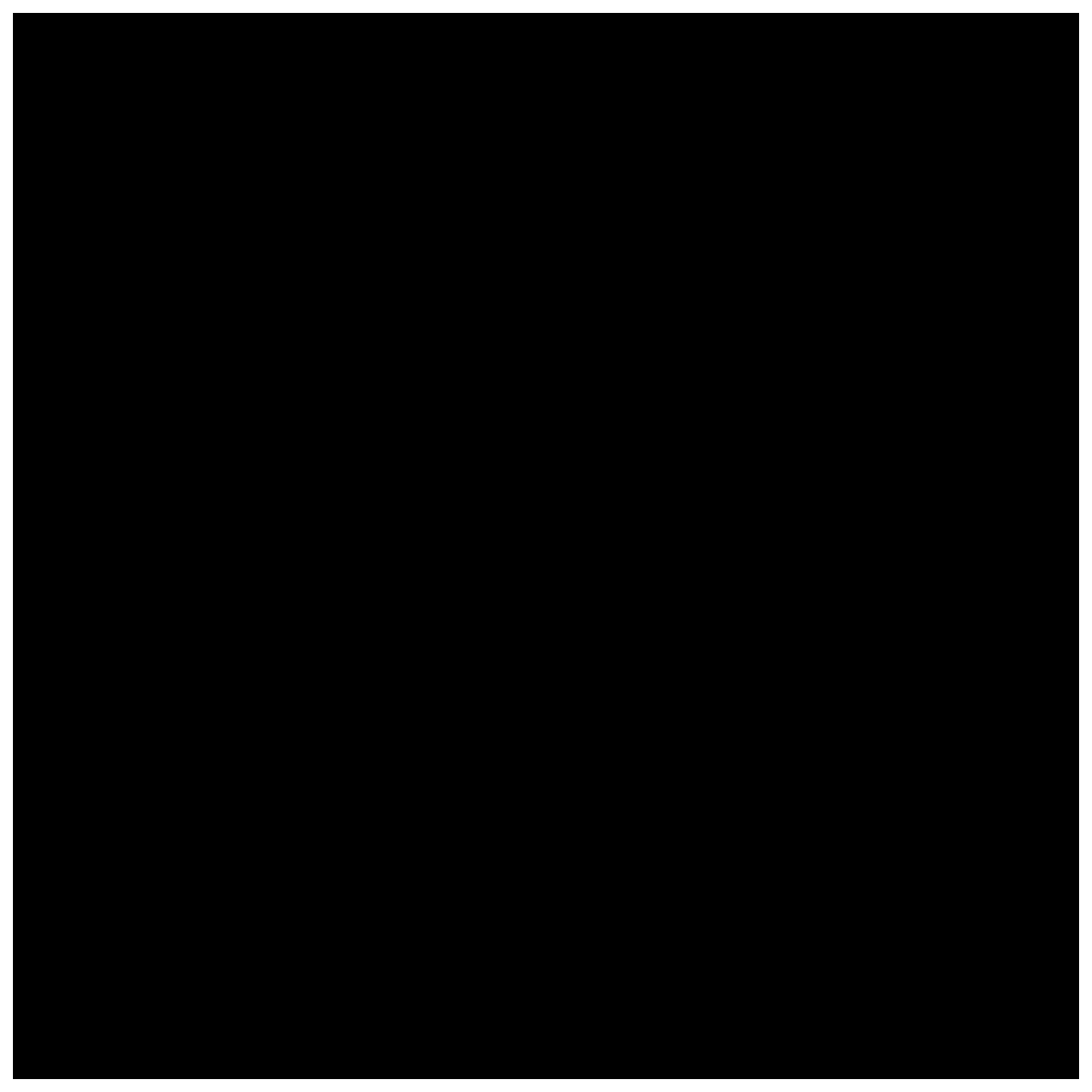}
      \caption{Original.}
    \end{subfigure}
    \hspace{15mm}
    \begin{subfigure}[b]{\expwidth\textwidth}
      \includegraphics[width=\textwidth]{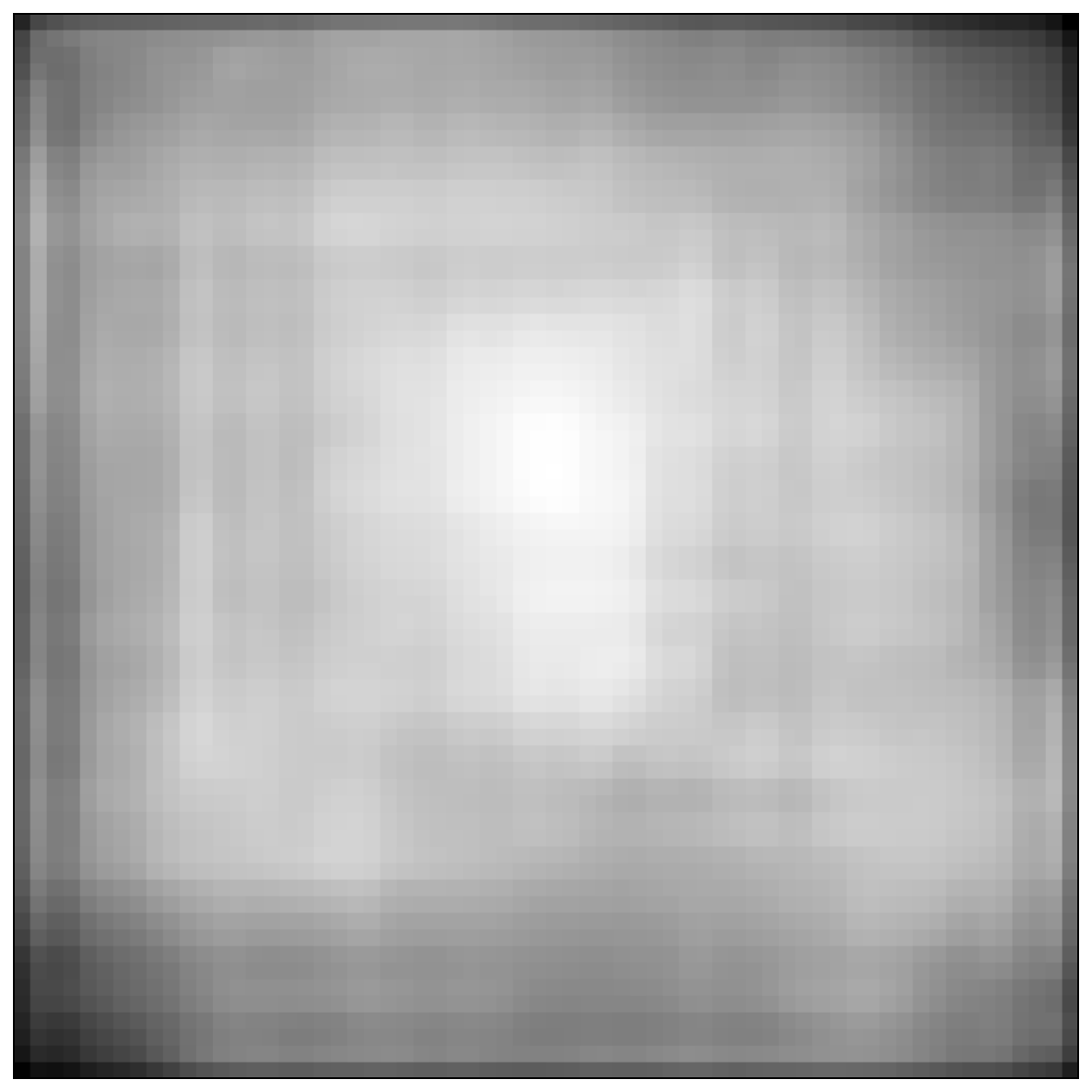}
      \caption{$\beta$-TCVAE reconstruction.}
    \end{subfigure}
    \caption{Blank image.}
    \label{fig:blank_image_reconstruction}
\end{figure}

\renewcommand{\expwidth}{0.4}
\begin{figure}[h!]
 \centering
 \begin{subfigure}[b]{\expwidth\textwidth}
    \includegraphics[width=\textwidth]{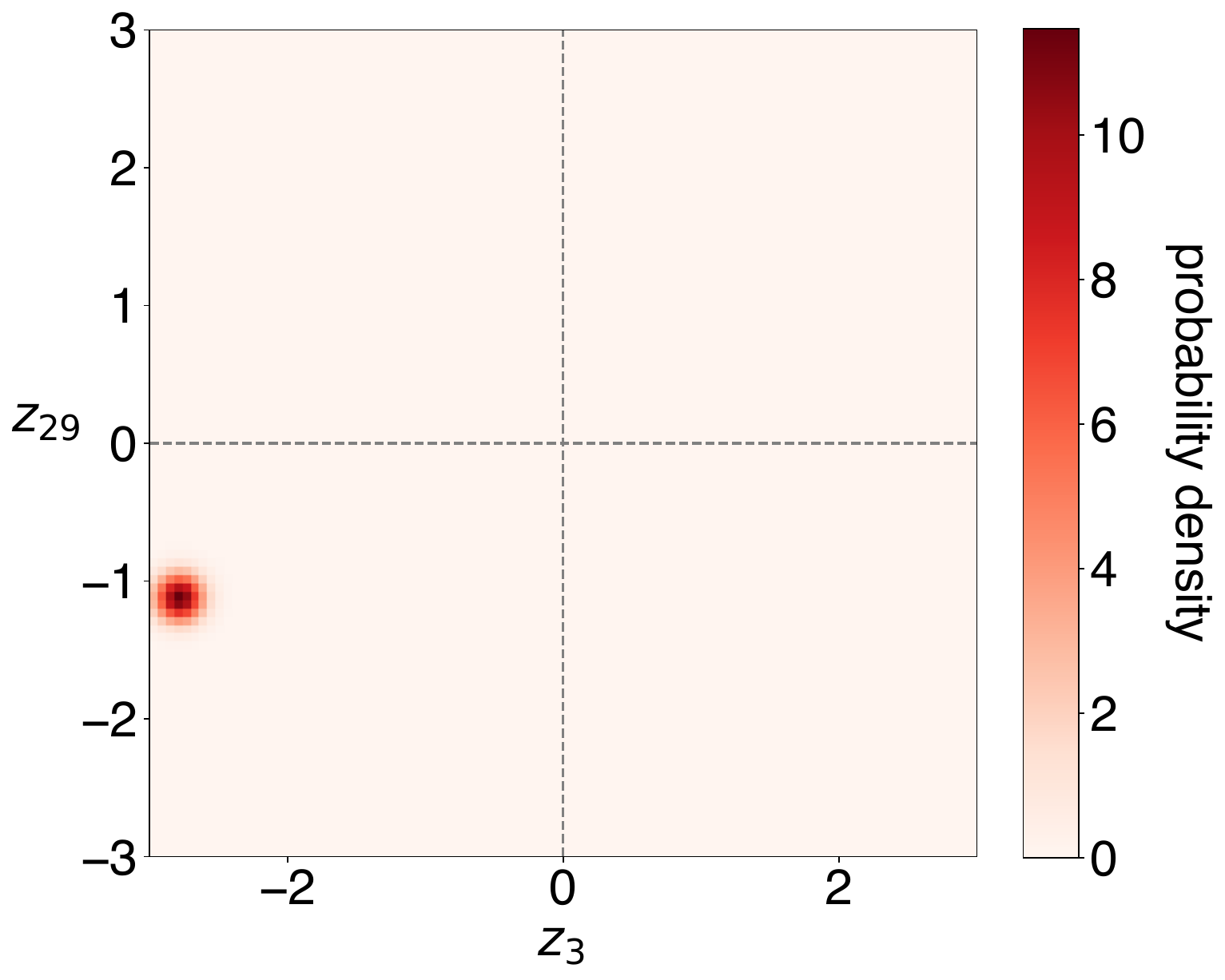}
      \caption{Size sub-space}
    \end{subfigure}
    \hspace{5mm}
    \begin{subfigure}[b]{\expwidth\textwidth}
    \includegraphics[width=\textwidth]{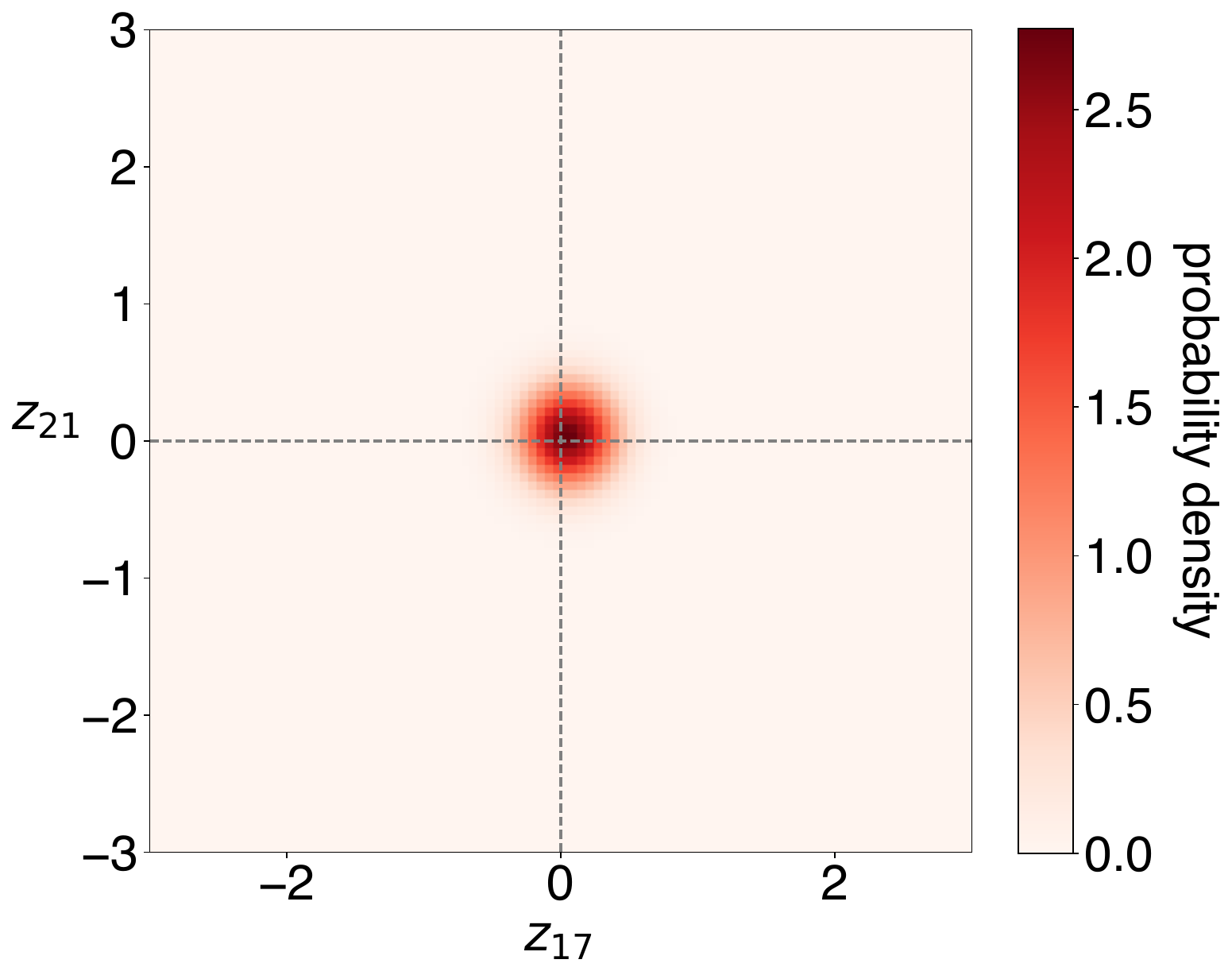}
      \caption{Eccentricity sub-space}
    \end{subfigure}
    \caption{Blank image encoding. Gaussian distribution shown reflects the mean and variance of the encoding.}
    \label{fig:blank_image_encoding}
\end{figure}

Therefore, even though this point in latent space does not encode any sensible cell image, it is possible to identify characteristics that explain the interphase classifications produced by our models. The only un-interpretable aspect of this classification is in the latent space encoding itself; since our $\beta$-TCVAE is un-interpretable with respect to OOD inputs, we can only speculate why the blank image was encoded to this specific point in the latent space. Nevertheless, the interpretability of our downstream models allow us to explain why they produce interphase classifications given this particular failure mode. We suggest that similar analyses could be used to study other failure modes in a wide range of contexts.

\end{document}